%% file: main.tex
\documentclass[lettersize,journal]{IEEEtran}
\usepackage{amsmath,amsfonts}
\usepackage{listings}
\usepackage{algorithmic}
\usepackage{bm}
\usepackage[ruled, vlined, linesnumbered]{algorithm2e}
\usepackage{array}
\usepackage{textcomp}
\usepackage{stfloats}
\usepackage{url}
\usepackage{verbatim}
\usepackage{graphicx}
\usepackage{cite}
\usepackage{multirow}
\usepackage{booktabs}
\usepackage{color}
\usepackage{subcaption}
\usepackage{pifont}
\usepackage[flushleft]{threeparttable}  
\usepackage[table]{xcolor}
\usepackage{wrapfig} 

\def\argmin{\mathop{\rm argmin}\limits}

\def\ourbenchmark{\texttt{EvoXBench}}
\def\firsttestsuite{C-10/MOP}
\def\secondtestsuite{IN-1K/MOP}

\begin{document}

\title{Neural Architecture Search as Multiobjective Optimization Benchmarks: Problem Formulation and Performance Assessment}

\author{Zhichao Lu,~\IEEEmembership{Member,~IEEE},
        Ran Cheng,~\IEEEmembership{Senior Member,~IEEE},
        Yaochu Jin,~\IEEEmembership{Fellow,~IEEE},\\
        Kay Chen Tan,~\IEEEmembership{Fellow,~IEEE},
        and~Kalyanmoy Deb,~\IEEEmembership{Fellow,~IEEE}
        \thanks{Z. Lu and R. Cheng are with the  
                Department of Computer Science and Engineering, 
                Southern University of Science and Technology, Shenzhen 518055, China. 
                E-mail: \{luzhichaocn, ranchengcn\}@gmail.com
                (\emph{Corresponding author: Ran Cheng}).}
        \thanks{Y. Jin is with the Chair of Nature Inspired Computing and Engineering, Faculty of Technology, Bielefeld University, 33615 Bielefeld, Germany, and also with the Department of Computer Science, University of Surrey, Guildford, Surrey GU2 7XH, U.K. Email:  yaochu.jin@uni-bielefeld.de.}
        \thanks{K. C. Tan is with the Department of Computing, The Hong Kong Polytechnic University, Hong Kong SAR. E-mail: kctan@polyu.edu.hk}
        \thanks{K. Deb is with the Department of Electrical and Computer Engineering, Michigan State University, East Lansing, MI 48824, USA. E-mail: kdeb@msu.edu}
        \thanks{This work was supported by the National Natural Science Foundation of China (No. 62106097, 61906081), China Postdoctoral Science Foundation (No. 2021M691424), the Shenzhen Peacock Plan (No. KQTD2016112514355531), the Program for Guangdong Introducing Innovative and Entrepreneurial Teams (No. 2017ZT07X386). Y. Jin is funded by an Alexander von Humboldt Professor for Artificial Intelligence endowed by the German Federal Ministry of Education and Research.}.
        
}

\markboth{IEEE Transactions on Evolutionary Computation,~Vol.~X, No.~X, July~2022}%
{Lu \MakeLowercase{\textit{et al.}}: A Sample Article Using IEEEtran.cls for IEEE Journals}


\maketitle

\input{0-abstract}

\input{1-introduction}
\input{2-background}

\input{3-formulation}

\input{4-benchmark}
\input{5-testsuite}
\input{6-experiment}
\input{7-conclusion}

\section*{Acknowledgement}
We wish to thank Beichen Huang for helping with the codes and experiments in MATLAB, Xukun Liu for helping with the database codes and monitoring the experiments, Yansong Huang for helping with the visualization, and Bowen Zheng for helping with the Transformer search space. 


\ifCLASSOPTIONcaptionsoff
  \newpage
\fi

\bibliography{egbib} 
\bibliographystyle{IEEEtran}

\appendix
\input{8-appendix}

\end{document}

%% file: 0-abstract.tex
\begin{abstract}

The ongoing advancements in network architecture design have led to remarkable achievements in deep learning across various challenging computer vision tasks.
Meanwhile, the development of neural architecture search (NAS) has provided promising approaches to automating the design of network architectures for lower prediction error.
Recently, the emerging application scenarios of deep learning (e.g., autonomous driving) have raised higher demands for network architectures considering multiple design criteria: number of parameters/weights, number of floating-point operations, inference latency, among others.
From an optimization point of view, the NAS tasks involving  multiple design criteria are intrinsically multiobjective optimization problems; hence, it is reasonable to adopt evolutionary multiobjective optimization (EMO) algorithms for tackling them.
Nonetheless, there is still a clear gap confining the related research along this pathway: on the one hand, there is a lack of a general problem formulation of NAS tasks from an optimization point of view; on the other hand, there are challenges in conducting benchmark assessments of EMO algorithms on NAS tasks.
To bridge the gap: 
(i) we formulate NAS tasks into general multi-objective optimization problems and analyze the complex characteristics from an optimization point of view; 
(ii) we present an end-to-end pipeline, dubbed \ourbenchmark{}, to generate benchmark test problems for EMO algorithms to run efficiently -- without the requirement of GPUs or Pytorch/Tensorflow; 
(iii) we instantiate two test suites  comprehensively covering two datasets, seven search spaces, and three hardware devices, involving up to eight objectives.
Based on the above, we validate the proposed test suites using six representative EMO algorithms and provide some empirical analyses.
The code of \ourbenchmark{} is available at \url{https://github.com/EMI-Group/EvoXBench}.

\end{abstract}

\begin{IEEEkeywords}
Evolutionary multi-objective optimization, neural architecture search, deep learning.
\end{IEEEkeywords}

%% file: 1-introduction.tex
\section{Introduction}

\IEEEPARstart{W}{ith} the development of deep neural networks (DNNs), deep learning has seen widespread growth in both research and applications, driven by improvements in computing power and a massive amount of data \cite{imagenet,seq2seq,silver2017mastering}. 
As the DNN models become more complex, it has come to a consensus that the manual process of designing DNN architectures is beyond human labor and thus unsustainable~\cite{vgg,resnet,dosovitskiy2021an}. 
Meanwhile, the emergence of neural architecture search (NAS) has paved a promising path towards alleviating this unsustainable process by automating the pipeline of designing DNN architectures~\cite{zoph2016}. 

In the early days, NAS was mainly dedicated to obtaining DNN models with low prediction error -- the most important target of most deep learning tasks.
More recently, however, the emerging application scenarios of deep learning (e.g. auto-driving) have raised higher demands for DNN architecture designs \cite{mosegnas}.
Consequently, apart from the prediction error, there are also other design criteria to consider, such as the number of parameters/weights and the number of floating-point operations (FLOPs).
Particularly, when deploying DNN models to edge computing devices, the hardware-related performance is equally important, such as inference latency and energy consumption.
Hence, from an optimization point of view, NAS tasks of today are intrinsically multiobjective optimization problems (MOPs) aiming to achieve optima of the multiple design criteria of the DNN architectures \cite{nsganet_tevc,nat}.

Due to the black-box nature of NAS, the complex properties from the optimization point of view (e.g. discrete decision variables, multi-modal and noisy fitness landscape, expensive and many objectives, etc.) have posed great challenges to conventional optimization methods driven by rigorous mathematical assumptions.
Accordingly, it is intuitive to resort to evolutionary multiobjective optimization (EMO) algorithms, which have well-established track records for handling complex MOPs~\cite{nsga2,moead,9723472}. 
Nonetheless, the progress of adopting EMO algorithms for NAS still lags well behind the progress of the general field of NAS research, due to the clear gap between the two fields caused by the following four main issues.

\noindent \ding{182} \textbf{Computational Cost:} Many baseline NAS methods require enormous computational cost~\cite{real2017large,nasnet2018,real2019regularized}, which is unfriendly to the researchers in academia; particularly, EMO algorithms often require a large number of function evaluations to run, such that the issue of expensive computational cost will become especially sharp.

\noindent \ding{183} \textbf{Algorithm Comparison:} Despite that the recent development has led to more efficient methods \cite{pham2018efficient,liu2018darts}, different NAS methods are not directly comparable to each other given the inherent differences in types of architectures (i.e., search spaces) and training procedures (i.e., fitness evaluations); hence, with the NAS methods in the current literature, it is very unlikely (if not impossible) to isolate the contribution of the optimization algorithm itself to the overall success of a NAS method~\cite{elsken2019neural}.

\noindent \ding{184} \textbf{Problem Formulation:} The research in EMO is often dedicated to solving MOPs with specific complex characteristics (e.g. multiple modalities, many objectives), but there is still no general formulation and analysis of NAS from the optimization point of view, thus having hindered EMO researchers to look deeper into the problems.

\noindent \ding{185} \textbf{Running Environment:} Performing NAS usually requires expertise in deep learning -- not only the experience in developing DNN models but also the knowledge in configuring the software/hardware environment; however, the conventional EMO researchers mainly focus on how to design efficient EMO algorithms, but not the environments to call/run the MOPs~\cite{platemo}. 

Among the four issues, issues \ding{182} and \ding{183} are not only for EMO but also for general NAS research.
To address these two issues, the recent NAS literature has made some attempts to propose the so-called \emph{tabular} NAS benchmarks~\cite{ying2019bench,Dong2020NAS-Bench-201,dong2021nats}. 
The basic idea is to rely on an \emph{exhaustive} evaluation of \emph{all} attainable architectures for {multiple} repetitions, resulting in confined search spaces and toy-scale DNN architectures that are unrealistic for real-world deployments or downstream tasks. 
To this end, the concept of \emph{surrogate} NAS benchmarks have been introduced \cite{zela2022surrogate}. 
It is expected that an accurate interpolation of the fitness landscape can be achieved on top of fix-budget training samples, thus extending NAS benchmarks to large search spaces with more than $10^{18}$ different DNN architectures. 
This recent progress in NAS benchmarks has led to a substantial reduction in the compute time required for NAS algorithm development and a significant improvement in the reproducibility of NAS research.

Despite the recent progress in addressing issues \ding{182} and \ding{183}, the outcome architectures (from a NAS algorithm) still need to be re-trained from scratch as the optimal weights are either not provided by or not available in the existing NAS benchmarks \cite{ying2019bench,zela2022surrogate}. Given a set of Pareto-optimal architectures returned by an EMO algorithm, such a re-training process itself can render the entire approach computationally prohibitive. 
Nonetheless, there is still little effort dedicated to addressing issues \ding{184} and \ding{185}.

Generally speaking, the challenges of issues \ding{184} and \ding{185} are two-fold:
from the research perspective, existing NAS benchmarks are merely provided as datasets, but not constructed as benchmark test problems on the basis of general formulation from the EMO point of view; 
from an engineering perspective, existing NAS benchmarks still require installations/configurations of sophisticated deep learning software/hardware environments, e.g., GPU, TensorFlow, PyTorch, among others. 
Moreover, each of the existing NAS benchmarks merely covers one or two search spaces, which is far from sufficient for benchmark comparisons between EMO algorithms; however, if one hopes to test an algorithm over different benchmarks, he/she has to make repeated modifications to have the algorithm compatible to the complicated interfaces of each benchmark, thus making the development of NAS algorithms tedious and error-prone.

This paper is dedicated to bridging the gap between EMO and NAS by addressing the four issues above. 
In summary, the main contributions are:

\vspace{2pt}
\noindent\textbf{--}
We present a general formulation of NAS problems from the multiobjective optimization perspective, involving three categories of optimization objectives: the prediction error objective, the complexity-related objectives, and the hardware-related objectives. On top of this formulation, we demonstrate a series of complex characteristics of NAS problems, including multiple modalities, many objectives, noisy objectives, degenerated Pareto fronts, among others.

\vspace{2pt}
\noindent\textbf{--}
We provide a unified and comprehensive NAS benchmark, dubbed \ourbenchmark{}, comprising seven search spaces. Specifically, \ourbenchmark{} covers two types of architectures (convolutional neural networks and vision Transformers), two widely-studied datasets (CIFAR-10 and ImageNet), six types of hardware (GPU, mobile phone, FPGA, among others.), and up to six types of optimization objectives (prediction error, FLOPs, latency, among others.).

\vspace{2pt}
\noindent\textbf{--}
We provide delicate engineering designs of \ourbenchmark{} to be user-friendly to EMO research.
Specifically, \ourbenchmark{} can be \emph{easily installed} without dependencies of deep learning software packages such as Pytorch or TensorFlow; and it can be \emph{directly called} in any programming language such as Python, MATLAB, or Java; most importantly, running without GPU, \ourbenchmark{} provides \emph{instant feedback} to MOEAs for real-time fitness evaluations.

\vspace{2pt}
\noindent\textbf{--}
On the basis of our \ourbenchmark{}, we generate two representative benchmark test suites, i.e., the \firsttestsuite{} and the \secondtestsuite{}, involving MOPs with up to six objectives.
Furthermore, we conduct a series of experiments to study the properties of the test problems using six representative MOEAs (NSGA-II, MOEA/D, IBEA, NSGA-III, HypE, and RVEA).

In the remainder of this paper, first, we provide necessary background information in Section~\ref{sec:background}; second, we provide formal NAS problem formulation and analysis in Section~\ref{sec:formulation}; third, we explain the design principles of our benchmark and test suite generation in Section~\ref{sec:design} and \ref{sec:test_suite}, respectively; fourth, we provide empirical evaluations in Section~\ref{sec:experiments}; finally, conclusions and future studies are discussed in Section~\ref{sec:conclusion}. 

%% file: 2-background.tex
\section{Background\label{sec:background}}

In this section, we provide a brief overview of the related concepts. The main notations used in this paper are summarized in Table~\ref{tab:notation}. 

\begin{table}[t]
\centering
\caption{Summary of notations.\label{tab:notation}}
\resizebox{.49\textwidth}{!}{%
    \begin{tabular}{@{\hspace{2mm}}lll@{\hspace{2mm}}}
    \toprule
    Category & Notation & Description \\ \midrule
    \multirow{3}{*}{Data} & C-10 & CIFAR-10: 10-class image classification dataset \cite{cifar10} \\
     & IN-1K & ImageNet 1K: 1,000-class image classification dataset \cite{imagenet} \\
     & $\mathcal{D}_{trn}, \mathcal{D}_{vld}, \mathcal{D}_{tst}$ & train, validation, and test splits of a dataset  \\ 
     \midrule
     \multirow{1}{*}{Hardware} & $h$ / $\mathcal{H}$ & a specific / set of hardware device \\\midrule
    \multirow{3}{*}{\begin{tabular}[c]{@{}l@{}}Decision\\ vectors\end{tabular}} & $\bm{x} \in \Omega$ & architecture decision variables and search space \\
     & $\bm{w}(\bm{x})$ & weights of an architecture \\ 
     & $D$ & number of decision variables \\ 
     \midrule
    \multirow{5}{*}{Objectives} & $f^e$ & prediction error on $\mathcal{D}_{vld}$\\
     & $\bm{f}^{c}$ & model complexity, i.e., \# of parameters and FLOPs\\
     & $\bm{f}^{\mathcal{H}}$ & hardware efficiency, e.g., latency, energy consumption, etc.\\
     & $\mathcal{L}_{trn}$ & loss on $\mathcal{D}_{trn}$ for training $\bm{w}(\bm{x})$ \\ 
     & $M$ & number of objectives \\ 
     \midrule
    \multirow{3}{*}{Others} & $N$ & population size \\
     & MAE & mean absolute error\\
     & $\tau$ & Kendall rank correlation coefficient \\
     \bottomrule
    \end{tabular}%
}
\end{table}

\subsection{Evolutionary Multiobjective Optimization}\label{sec:2.1}

Generally, a multiobjective optimization problem (MOP), involving more than one conflicting objective to be optimized simultaneously, can be formulated as:

\begin{equation}
\begin{aligned}
\min\limits_{\bm{x}} \quad & \bm{F}(\mathbf{x}) = \big(f_1(\bm{x}), f_2(\bm{x}), ..., f_M(\bm{x})\big), \\
\text{s.t.}  \quad & \bm{x} \in X, \quad \bm{F} \in Y,\\
\end{aligned}
\label{eq:problem}
\end{equation}
where $X \subset \mathbb{R}^D$ and $Y \subset \mathbb{R}^M$ are known as the \emph{decision space} and the \emph{objective space} respectively.
Since $f_1(\bm{x}), \ldots, f_M(\bm{x})$ are often conflicting with each other, there is no single solution that can achieve optima on all objectives simultaneously; instead, the optima to such an MOP is often a set of solutions trading-off between different objectives, known as the \emph{Pareto optima}. 
Specifically, the images of the Pareto optima are known as the \emph{Pareto set} (PS)  and the \emph{Pareto front} (PF) in the decision space and objective space respectively. 

In practice, the target of solving an MOP is to approximate the PS/PS with a limited number of candidate solutions trading-off between different objectives.
To this end, various EMO algorithms have been proposed during the past two decades.
Generally, the EMO algorithms can be categorized into three categories: dominance-based ones~\cite{deb2002fast,zitzler2001spea2,corne2001pesa}, decomposition-based ones~\cite{moead,liu2013decomposition}, and performance-indicator-based ones~\cite{zitzler2004indicator,emmerich2005emo}. Readers are referred to supplementary materials \S I-A for more details.

To assess the performance of various EMO algorithms in the literature, a number of challenging EMO benchmark test suites have also been designed to consider different complicated characteristics, e.g., the ZDT test suite~\cite{ZDT}, the DTLZ test suite~\cite{DTLZ}, the WFG test suite~\cite{WFG}, the MaF test suite~\cite{MaF}, the LSMOP test suite~\cite{LSMOP}, etc.
Undoubtedly, the EMO benchmark test suites play an important role in pushing the boundaries of EMO research, having promoted the ongoing delicate designs of EMO algorithms to face the new challenges of emerging benchmark test suites.


\subsection{Neural Architecture Search (NAS)}

\begin{figure}[t]
    \centering
    \begin{subfigure}[b]{0.48\textwidth}
        \centering
        \includegraphics[trim={0 0 0 0}, clip, width=.98\textwidth]{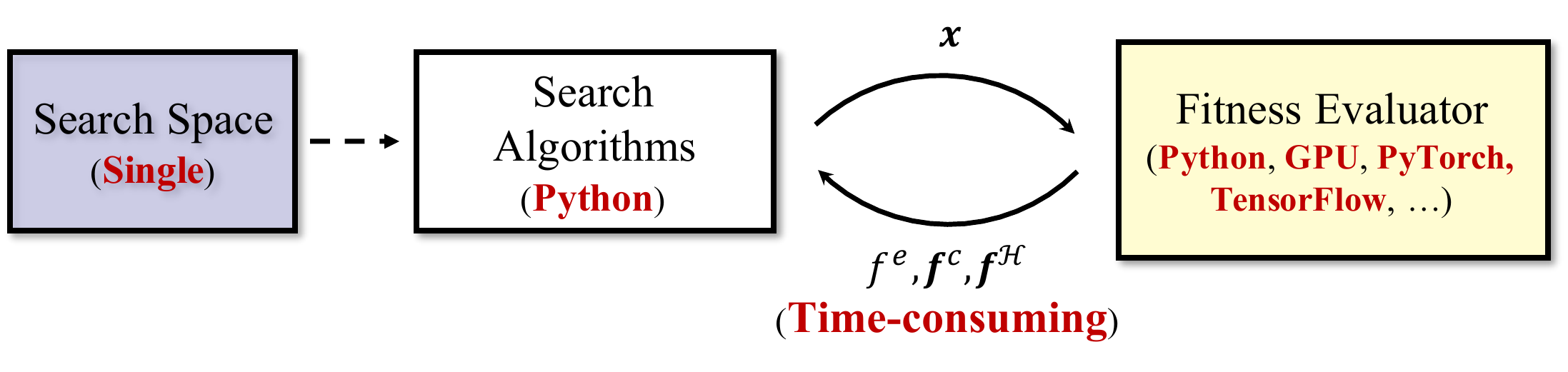}
        \vspace{-8pt}
        \caption{Standard NAS Pipeline \label{fig:nas_overview}}
    \end{subfigure} \\ \vspace{5pt}
    \centering
    \begin{subfigure}[b]{0.48\textwidth}
        \centering
        \includegraphics[trim={0 0 0 0}, clip, width=.98\textwidth]{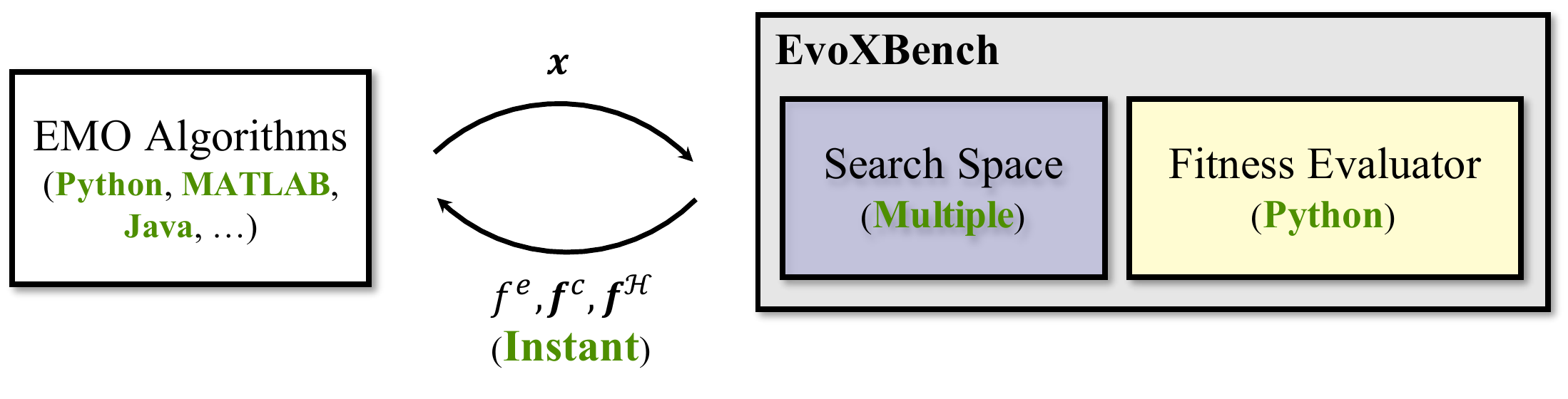}
        \vspace{-8pt}
        \caption{Our NAS Benchmark tailored for EMO\label{fig:evoxbench_overview}}
    \end{subfigure}
    \caption{Overall system-level comparison.\label{overview}}
\end{figure}

The goal of NAS is to automate the design of the architectures of DNN models by formulating it as an optimization problem \cite{zoph2016}. 
As depicted in Fig.~\ref{fig:nas_overview}, a standard NAS pipeline involves three main components: 
(i) a search space that pre-defines how a DNN architecture is represented; 
(ii) a search algorithm for generating suitable architectures according to pre-specified criteria; 
and (iii) an evaluator to estimate the fitness of an architecture \cite{elsken2019neural}. 
In the following, we provide a brief review of the literature related to NAS search spaces and refer readers to supplementary materials \S I-B for NAS search algorithms and fitness evaluators. 

\begin{figure}[t]
    \centering
    \begin{subfigure}[b]{0.48\textwidth}
        \centering
        \includegraphics[trim={0 0 0 0}, clip, width=.98\textwidth]{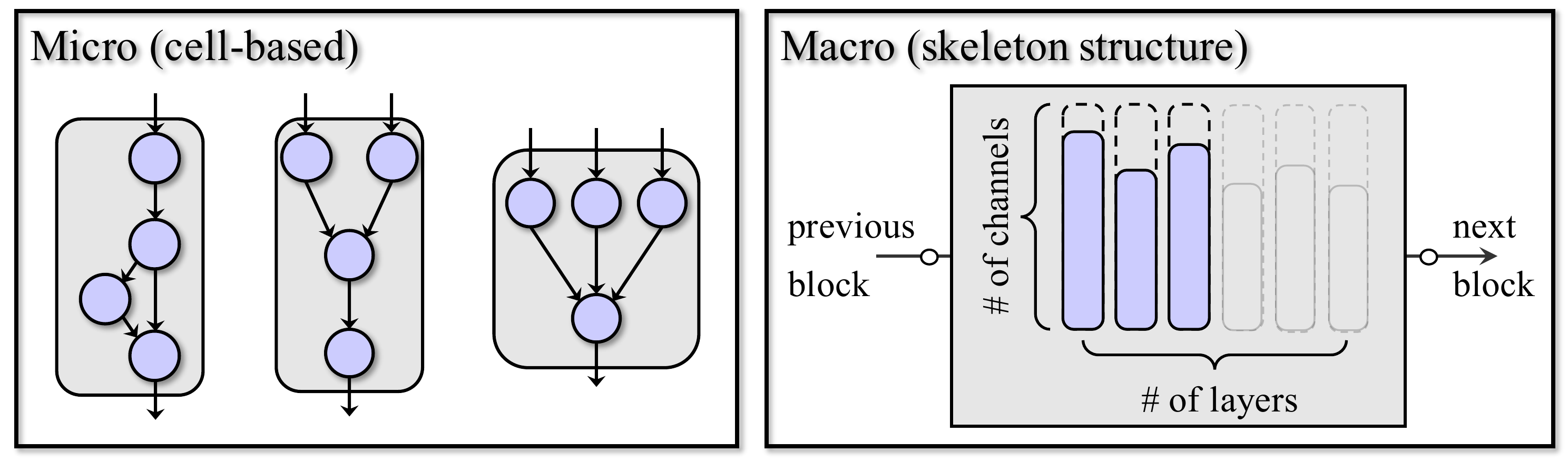}
    \end{subfigure}
    \caption{Two types of widely used search spaces.\label{fig:search_space_types}}
\end{figure}

The architectural design of a DNN model can be decomposed into the design of (i) the skeleton of the network (i.e., the depth and width of the network) and (ii) the layer (i.e., the types and arrangements of the operators such as convolution, pooling, etc.). Early work made attempts to design these two aspects simultaneously, resulting in DNN architectures with sub-optimal performance \cite{zoph2016,baker2016designing,real2017large}. Accordingly, recent efforts resort to only one aspect, and two types of search spaces have been proposed. Specifically, \emph{micro} search spaces focus on designing a modular computational block (also known as a \emph{cell}), which is repeatedly stacked to form a complete DNN architecture following a pre-specified template \cite{nasnet2018,pham2018efficient,real2019regularized,liu2018darts};
\emph{macro} search spaces focus on designing the network skeleton while leaving layers to well-established designs \cite{tan2019mnasnet,cai2018proxylessnas,Howard_2019_ICCV}. 
A pictorial illustration is provided in Fig.~\ref{fig:search_space_types}.

On one hand, micro search spaces confine the search to the inner configuration of a layer, leading to a considerable reduction in search space volume at the cost of structural diversity among layers, which has been shown to be crucial for hardware efficiency \cite{chitty2022neural}. 
On the other hand, despite compelling performance on hardware, macro search spaces are oftentimes criticized for producing architecturally similar DNN models \cite{ren2021comprehensive}.

\subsection{Existing NAS Benchmarks}
As one of the earliest tabular benchmarks\footnote{A tabular benchmark means that all attainable solutions are exhaustively evaluated and stored as a tabular dataset a priori.}, NAS-Bench-101 \cite{ying2019bench} adopts a micro search space where approximately 423K unique cell structures are exhaustively evaluated three times on the CIFAR-10 dataset \cite{cifar10} for image classification. 
All results are stored in a \emph{TFRecord} file, which is a binary file format designed specifically for TensorFlow---a dedicated software for machine learning with DNNs \cite{tensorflow2015-whitepaper}. 
A subsequent tabular benchmark of NAS-Bench-201 \cite{Dong2020NAS-Bench-201}, with a similar micro search space of 16K cell structures, extends the dataset to include the CIFAR-100 \cite{cifar10} and downsampled ImageNet 16$\times$16 \cite{chrabaszcz2017downsampled} datasets. Moving beyond tabular benchmarks, NAS-Bench-301 introduces the first surrogate-model-based benchmark on the CIFAR-10 dataset. 
With sophisticated regression models (e.g., graph isomorphism network \cite{xu2018how} and tree-based gradient boosting methods \cite{chen2016xgboost,ke2017lightgbm}) on top of a large number of training samples ($\sim$60K), it extends the volume of a micro search space to include more than $10^{21}$ cell structures. 
Both NAS-Bench-201 and -301 are implemented in PyTorch, another sophisticated deep learning software \cite{pytorch}, with additional dependencies for running sophisticated surrogate models. 

Concurrently, a steady stream of follow-up work have been reported, extending NAS benchmarks to include: (i) a macro search space for finding the optimal channels of each layer in a DNN (i.e., NATS-Bench \cite{dong2021nats}), (ii) hardware-related performance (HW-NAS-Bench \cite{li2021hwnasbench}), (iii) or other application domains such as natural language processing \cite{klyuchnikov2022bench}, speech recognition \cite{mehta2022nasbenchsuite}, transfer learning \cite{duan2021transnas}, among others.

%% file: 3-formulation.tex
\section{Problem Formulation\label{sec:formulation}}

Given a target dataset $\mathcal{D} = \{\mathcal{D}_{trn}, \mathcal{D}_{vld}, \mathcal{D}_{tst}\}$ and target hardware device set $\mathcal{H} = \{ h_{1}, \ldots, h_{|\mathcal{H}|}\}$, without loss of generality, a NAS task can be formulated as a multi-objective optimization problem:

\begin{equation}
\begin{aligned}
\min \limits_{\bm{x}} & \hspace{2mm} \bm{F}(\bm{x}) = \Big(f^{e}\big(\bm{x}; \bm{w}^*(\bm{x})\big), \bm{f}^{c} \big(\bm{x}\big), \bm{f}^{\mathcal{H}} \big(\bm{x}\big)\Big) \\
\textrm{s.t.}  & \hspace{2mm} \bm{w}^*(\bm{x}) \in \argmin~\mathcal{L}_{trn}(\bm{x}; \bm{w}), \hspace{1em}\bm{x} \in {\Omega}
\end{aligned}
\label{def:formulation1}
\end{equation}
with
\begin{equation}
\centering
 \bm{f}^{c} \big(\bm{x}\big):  {f}^c_1 \big(\bm{x}\big), \cdots,  {f}^c_{M^c} \big(\bm{x}\big),
\label{def:formulation2}
\end{equation}
and
\begin{equation}
\bm{f}^{\mathcal{H}} \big(\bm{x}\big):
\left\{
\begin{array}{l}
{f}^{h_1}_1 \big(\bm{x}, \big), \cdots,  {f}^{h_1}_{M^h_1} \big(\bm{x}\big),\\
\hspace{5mm} \vdots\\
{f}^{h_{|\mathcal{H}|}}_1 \big(\bm{x}, \big),\cdots, {f}^{h_{|\mathcal{H}|}}_{M^h_{|\mathcal{H}|}} \big(\bm{x}\big)
\end{array} 
\right.
,
\label{formulation3}
\end{equation}
where ${\Omega}$ is the search space, $\bm{x}$ and $\bm{w}(\bm{x})$ denote an encoded network architecture (i.e. decision vector) and the corresponding weights of the decoded network respectively, and $\bm{w}^*(\bm{x})$ denotes the optima weights achieving the minimal loss on the training set $\mathcal{D}_{trn}$;
$f^e$ denotes the objective function indicating the \emph{prediction error} of the model, $\bm{f}^{c}$ and $\bm{f}^{\mathcal{H}}$ denote the objective functions indicating the performance related to \emph{model complexity} and \emph{hardware devices} respectively.

In practice, the objective functions can be formulated in various ways.
For example, to formulate $f^e$, the most straightforward approach is to associate it with the validation loss on $\mathcal{D}_{vld}$;
to formulate $\bm{f}^{c}$, we usually refer to the number of parameters (i.e. weights) and floating point of operations (FLOPs) respectively;
to formulate $\bm{f}^{c}$, the most commonly considered metrics are hardware latency and hardware energy consumption.
Besides, since it can be computationally prohibitive to fully train a model on $\mathcal{D}_{trn}$ to obtain the exact validation loss on $\mathcal{D}_{vld}$ for $f^e$, surrogate-modeling approaches are also adopted.

Based on the formulation above, the following subsections will elaborate on the complex characteristics of NAS tasks from the optimization point of view.

\begin{figure}[t]
    \centering
    \begin{subfigure}[b]{0.24\textwidth}
        \centering
        \includegraphics[trim={0 0 0 0}, clip, width=.98\textwidth]{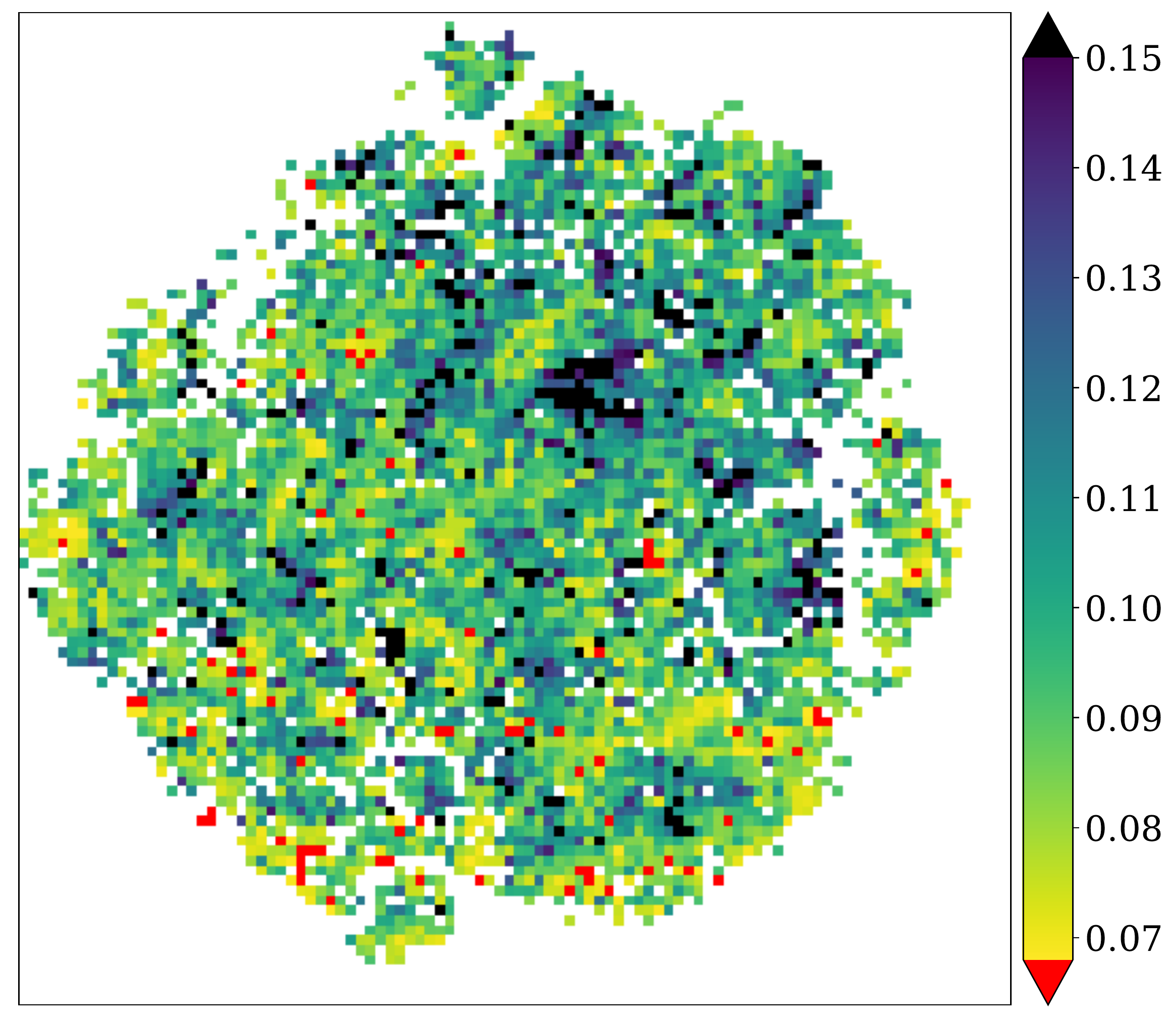}
        \caption{$\Omega$ = NB101\label{fig:nb101_landscape_tsne}}
    \end{subfigure} \hfill
    \centering
    \begin{subfigure}[b]{0.24\textwidth}
        \centering
        \includegraphics[trim={0 0 0 0}, clip, width=.98\textwidth]{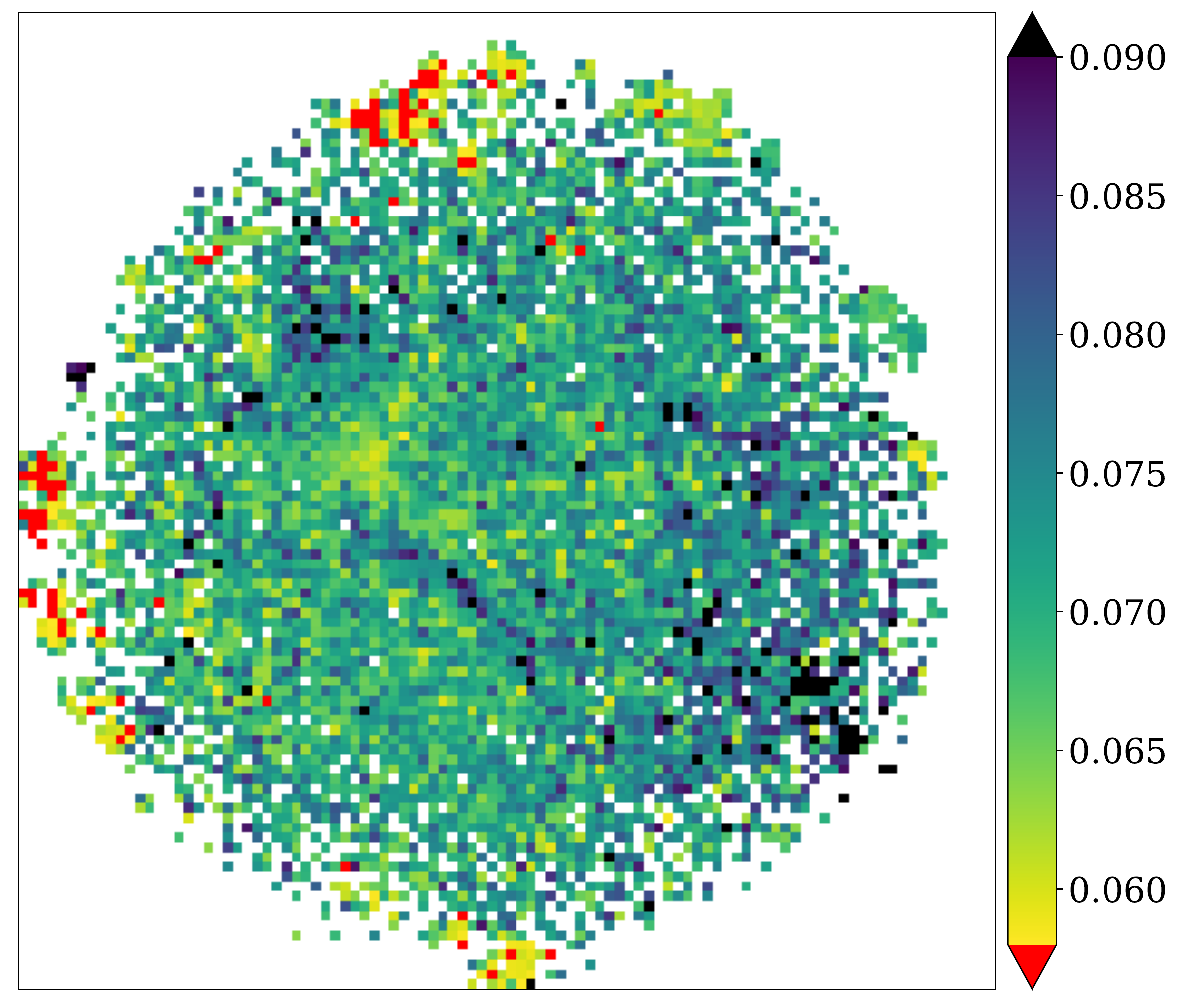}
        \caption{$\Omega$ = DARTS \label{fig:darts_landscape_tsne}}
    \end{subfigure}\\
    \centering
    \begin{subfigure}[b]{0.15\textwidth}
        \centering
        \includegraphics[trim={0 0 0 0}, clip, width=\textwidth]{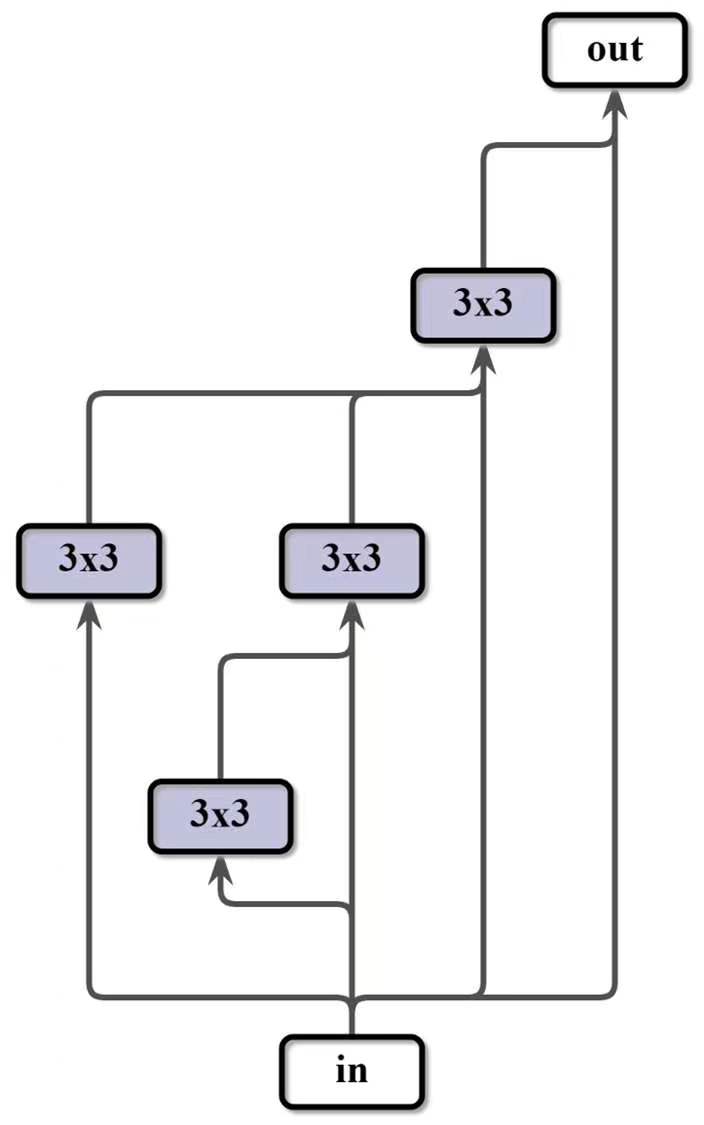}
        \caption{${f}^e=0.0576$
        \label{fig:nb201_arch1}}
    \end{subfigure}\hfill
    \begin{subfigure}[b]{0.15\textwidth}
        \centering
        \includegraphics[trim={0 0 0 0}, clip, width=\textwidth]{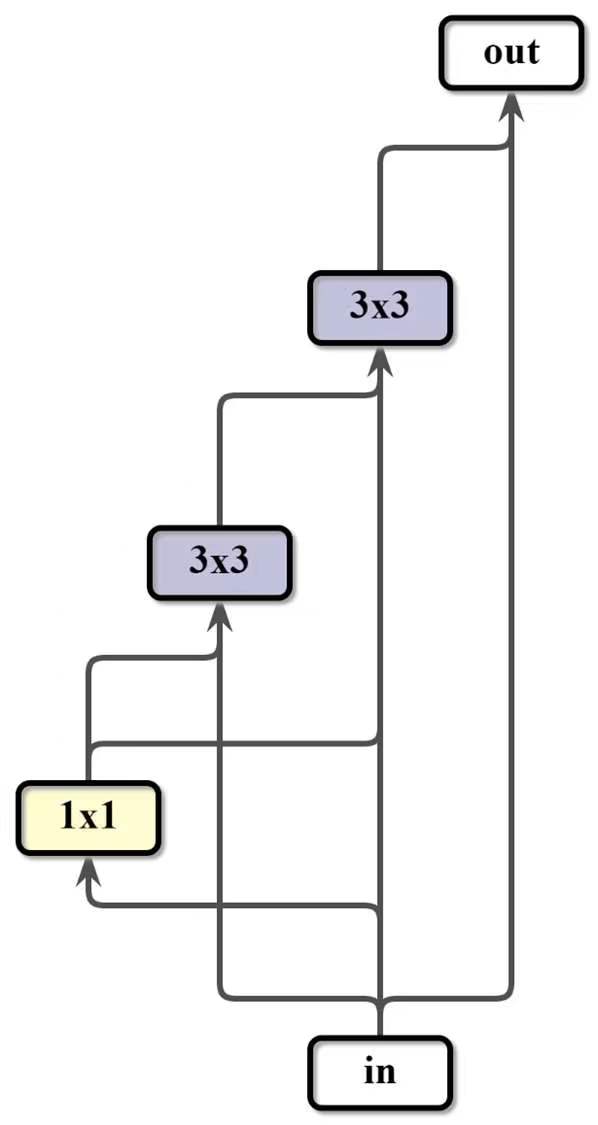}
        \caption{${f}^e=0.0578$
        \label{fig:nb201_arch2}}
    \end{subfigure}\hfill
    \begin{subfigure}[b]{0.15\textwidth}
        \centering
        \includegraphics[trim={0 0 0 0}, clip, width=\textwidth]{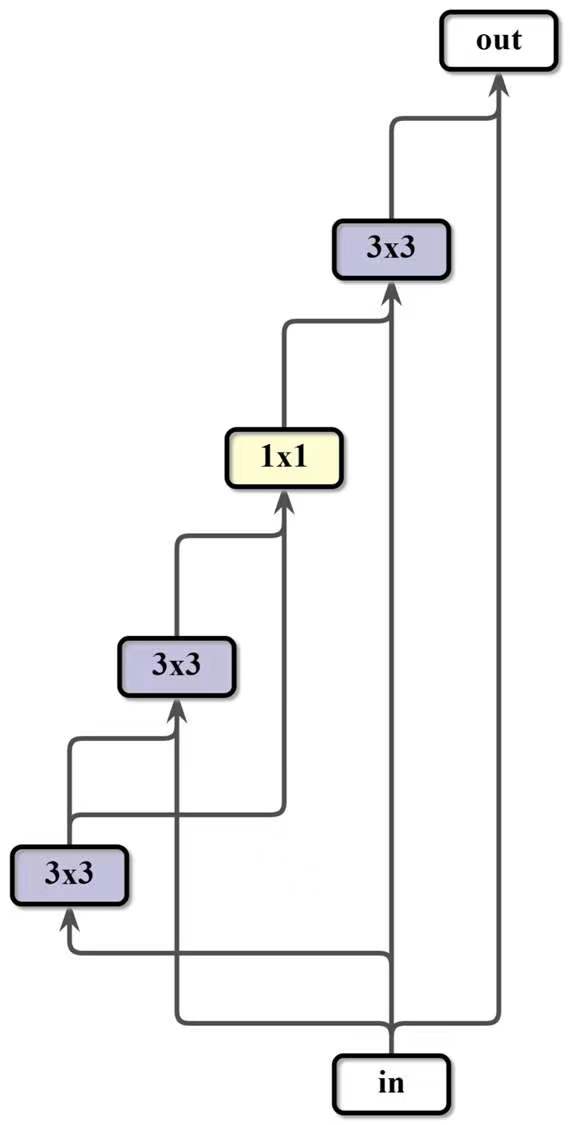}
        \caption{${f}^e=0.0579$
        \label{fig:nb201_arch3}}
    \end{subfigure}
    \caption{Fitness landscapes of $f^e$ (i.e. prediction error) on (a) NB101 and (b) DARTS. We project the original high-dimensional decision space to 2D latent space via t-SNE \cite{tsne}. We random sample 10K solutions from each search space and average ${f}^e$ within each small area. The top-performing solutions are highlighted in red. (c)-(e) are three solutions with similar ${f}^e$ but different decision vectors (i.e., architectures) sampled from the NB101 search space. 
    \label{fig:multimodal_lanscape}}
\end{figure}

\subsection{Multi-modal Fitness Landscapes}
The fitness landscape of an optimization problem depicts how fitness values change over the decision space.
Specifically, a \emph{multi-modal} landscape refers to the case involving multiple optimum solutions (a.k.a multi-modality) having very close (or even the same) fitness values~\cite{kerschke2019search}.
When searching over a multi-modal fitness landscape, the target is often to find the multiple optimum solutions once, such that the decision-maker is able to choose among those solutions according to personal preferences.
Particularly, there are emerging research interests in studying multi-modal MOPs in the EMO community~\cite{tanabe2019review}.

In the context of NAS, it is very likely that different network architectures can lead to very close (or even the same) prediction accuracy or hardware-related metrics such as latency.
As evidenced in Fig.~\ref{fig:multimodal_lanscape}, for example, the fitness landscapes of $f^e$ on search spaces NB101 and NB201 contain a number of optima having very close fitness values (i.e., prediction errors);
correspondingly, the solutions having very close fitness values could represent very different architectures.

\subsection{Noisy Objectives \label{sec:noisy_objs}}
In practical optimization problems, a solution is often evaluated through stochastic simulations, physical experiments, or even interactions with users. 
As a result, the outputs for repeated evaluations at the same decision vector are not deterministic. 
From an optimization point of view, we characterize such an evaluation process as a \emph{noisy objective}~\cite{jin2005evolutionary, rakshit2017noisy}. 
Evolutionary computation is believed to be well suited for tackling noisy objectives given its population-based nature and randomized search heuristics. 
Accordingly, developing effective algorithms for solving noisy optimization problems has always been an active research topic in the EMO community~\cite{4220676,6730915}. 

\begin{figure}[ht]
    \centering
    \begin{subfigure}[b]{0.24\textwidth}
        \centering
        \includegraphics[trim={0 0 0 0}, clip, width=\textwidth]{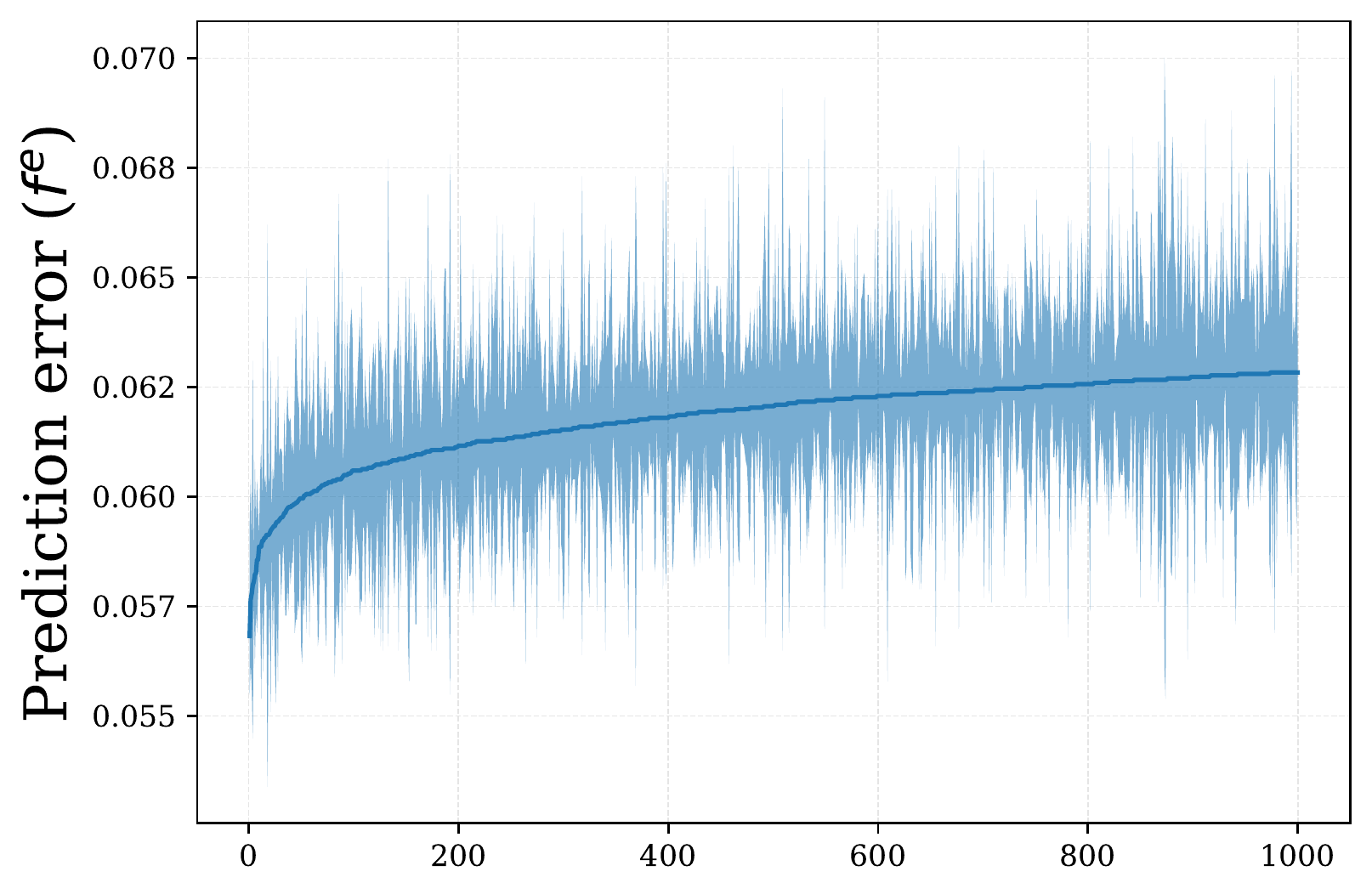}
        \caption{$\Omega = $ NB101\label{fig:nb101_noisy_test_error}}
    \end{subfigure} \hfill
    \begin{subfigure}[b]{0.24\textwidth}
        \centering
        \includegraphics[trim={0 0 0 0}, clip, width=\textwidth]{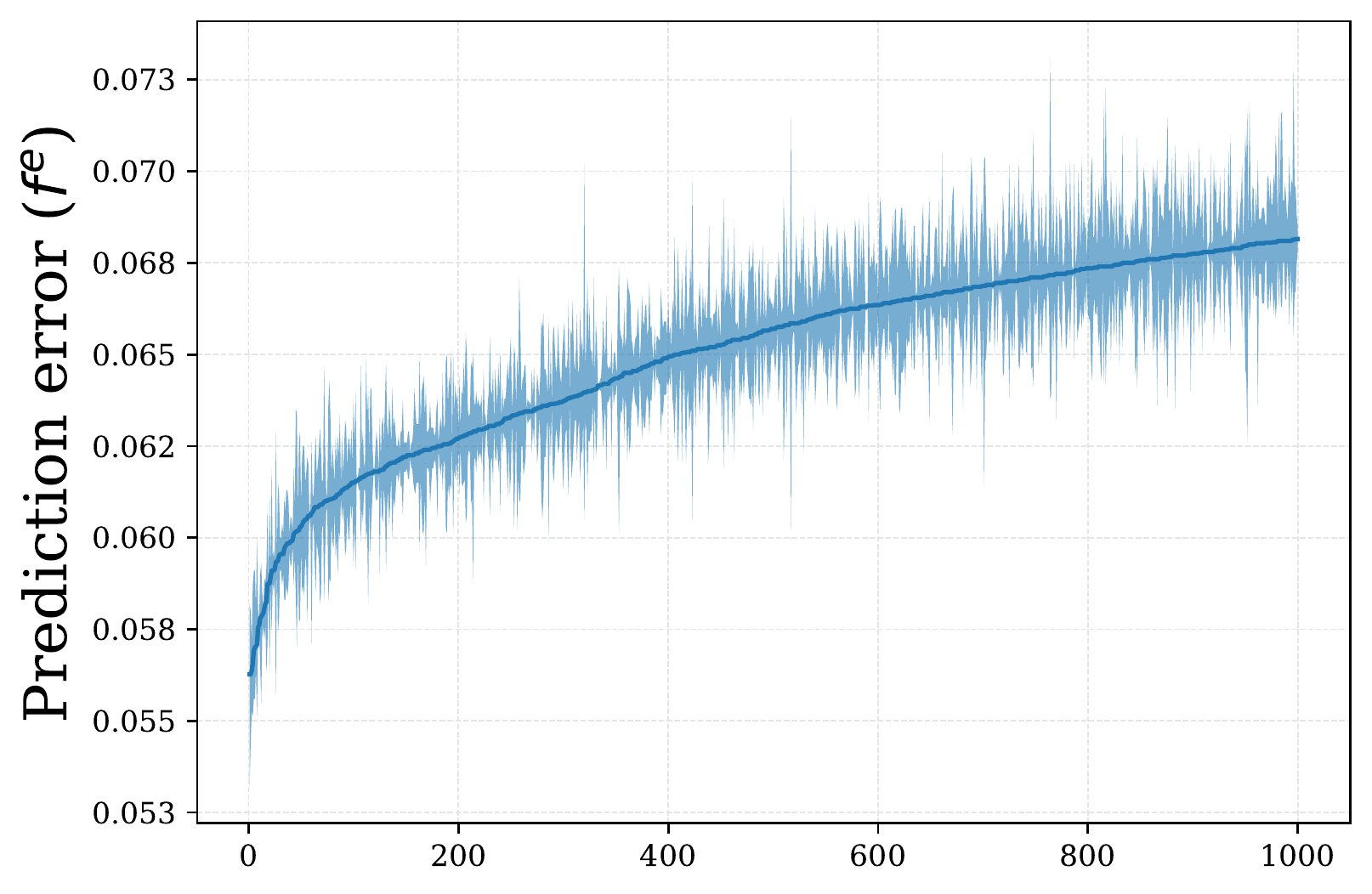}
        \caption{$\Omega = $ NB201\label{fig:nb201_noisy_test_error}}
    \end{subfigure}
    \caption{The mean ${f}^e$ (prediction error) of the top 1,000 solutions (according to ${f}^e$) on (a) NB101 and (b) NB201 search spaces from three repetitions. The standard deviations (i.e., \emph{noise}) of ${f}^e$ are visualized with the shaded error bars.\label{fig:noisy_objective}}
\end{figure}

In the context of NAS, the evaluation of ${f}^e$ (prediction error) of an architecture is almost always noisy due to various stochastic components involved in training the corresponding weights of the architecture, such as randomized data augmentation and loading order. 
Fig.~\ref{fig:noisy_objective} provides examples visualizing the noises in evaluating ${f}^e$ on NB101 and NB201 search spaces respectively.
Furthermore, hardware-related objectives also critically depend on the operating conditions of the hardware, such as current loading, ambient temperature, etc. 
Hence, the evaluation of $\bm{f}^{\mathcal{H}}$ is also subject to high variance. 


\subsection{Many Objectives}

The number of objectives in a MOP is an important character to be considered in EMO.
Particularly, if the number of objectives increases to more than three, an MOP is known to be a \emph{many-objective optimization problem (MaOP)}, posing great challenges to the EMO algorithms~\cite{li2015many}.
On the one hand, it is challenging to obtain the sparsely distributed candidate solutions when considering both convergence and diversity in the high-dimensional objective space; on the other hand, it is challenging to perform decision-making when trading off among many objectives.

\begin{figure}[ht]
    \centering
    \includegraphics[trim={0.3cm 0 0.3cm 0}, clip, width=.49\textwidth]{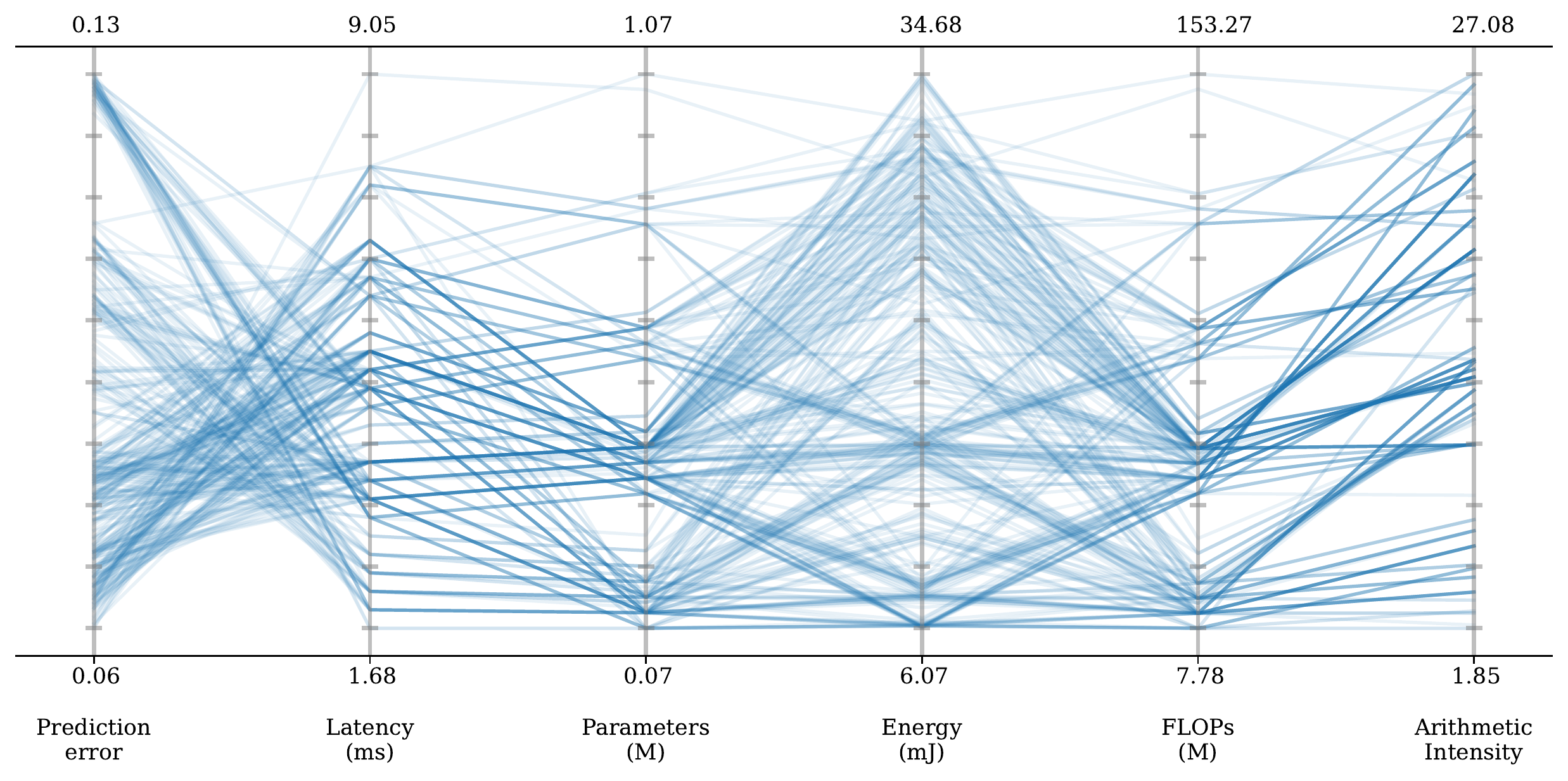}
\caption{The parallel coordinates of non-dominated solutions of architectures sampled from the NB021 search space with hardware device of Eyeriss (i.e., $\Omega$ = MNV3, $\mathcal{H} = \{h_1 = \mbox{Eyeriss} \}$).
The six optimization objectives: ${f}^e$: prediction error, ${f}^c_1$: number of parameters (M), ${f}^c_2$: number of FLOPs (M), ${f}^{\mathcal{H}}_1$: latency (ms), ${f}^{\mathcal{H}}_3$: energy (mJ), ${f}^{\mathcal{H}}_3$: Arithmetic Intensity (ops/byte).
\label{fig:many_objectives}}
\end{figure}

As formulated above, a NAS task can involve up to a number of $1 + M^c + \sum_{i=1}^{|\mathcal{H}|} M^h_i$ objective functions to be optimized simultaneously, where $M_c$ denotes the number of objective functions related to model complexity, and $M^h_i$ denotes the number of objective functions related to the $i$-th hardware device in $\mathcal{H}$.
As exemplified in Fig.~\ref{fig:many_objectives}, a NAS task on the search space of NB201 consists of
six objectives with $M^c = 2$ and $M^h_1 = 3$.

\subsection{Badly Scaled Objectives}

Ideally, the objective functions of a MOP should be scaled to the same (or similar) values such that EMO algorithms could work effectively.
This is particularly important to the decomposition-based EMO algorithms whose performances largely rely on the predefined weight/reference vectors.
Without any preference or \textit{a priori} knowledge, the weight/reference vectors are uniformly sampled  from a unit hyper-plan/sphere, assuming that the objective functions of the MOP to be solved are also well normalized.
In practice, however, the objective functions can be of very different scales and there is no way to formulate them in a normalized manner.
An MOP of such a character is often known to be \emph{badly scaled}~\cite{he2021survey}.

\begin{figure}[ht]
    \centering
    \begin{subfigure}[b]{0.24\textwidth}
        \centering
        \includegraphics[trim={0 0 0 0}, clip, width=.98\textwidth]{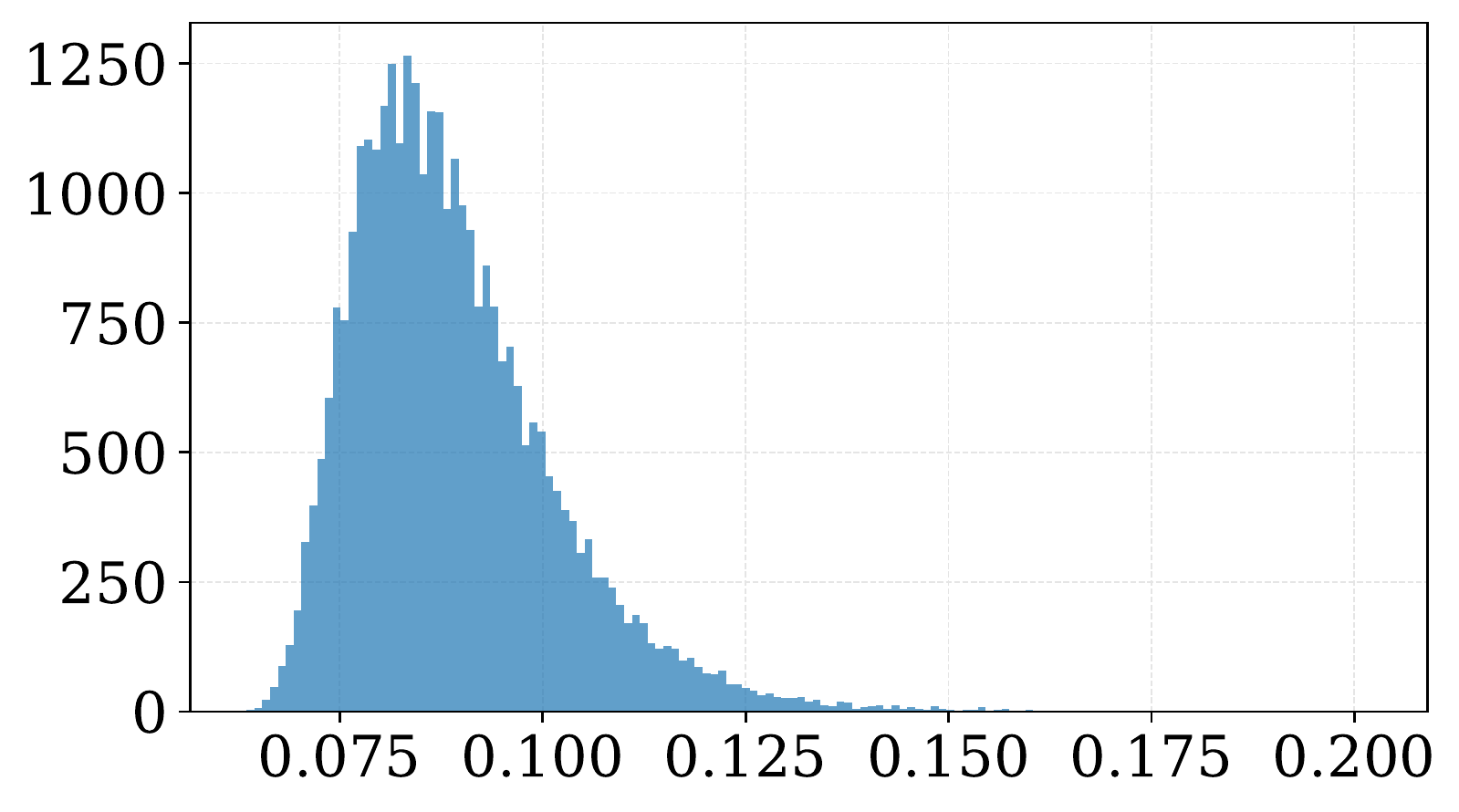}
        \caption{${f}^e$: prediction error \label{fig:nats_err_hist}}
    \end{subfigure} \hfill
    \begin{subfigure}[b]{0.24\textwidth}
        \centering
        \includegraphics[trim={0 0 0 0}, clip, width=.98\textwidth]{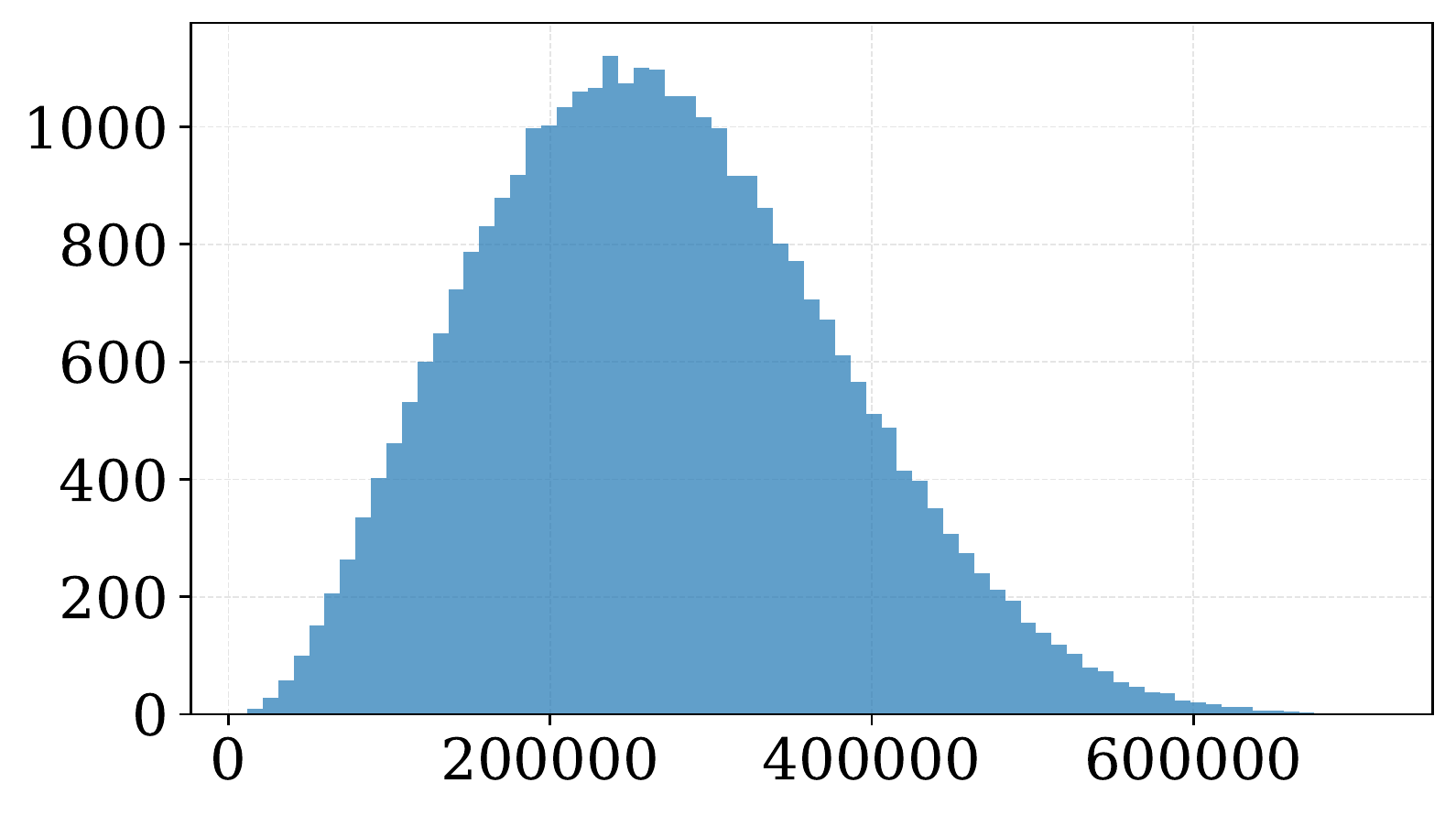}
        \caption{${f}^c_1$: number of parameters\label{fig:nats_params_hist}}
    \end{subfigure}\\
    \centering
    \begin{subfigure}[b]{0.24\textwidth}
        \centering
        \includegraphics[trim={0 0 0 0}, clip, width=.98\textwidth]{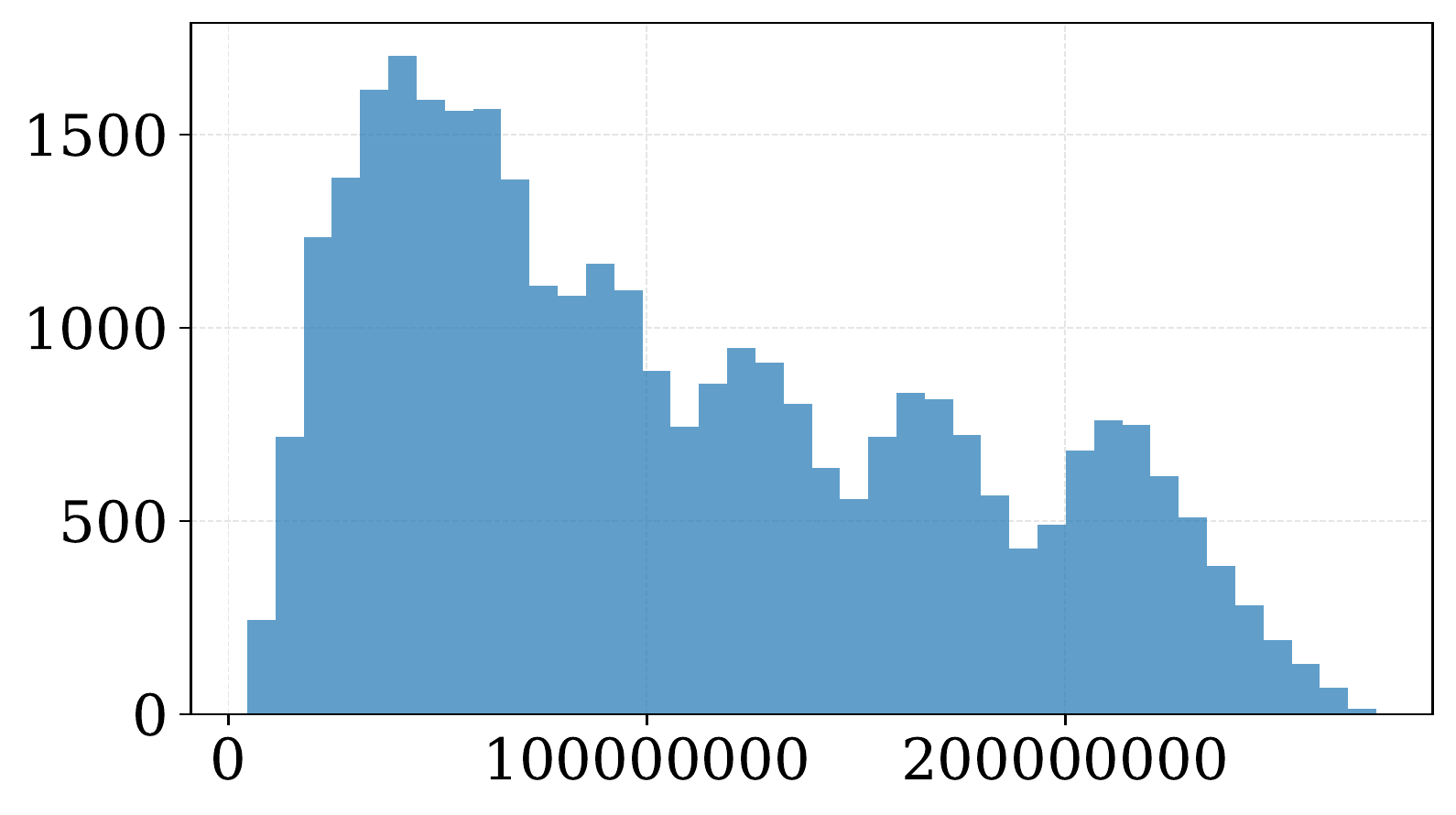}
        \caption{${f}^c_2$: number of FLOPs \label{fig:nats_flops_hist}}
    \end{subfigure} \hfill
    \begin{subfigure}[b]{0.24\textwidth}
        \centering
        \includegraphics[trim={0 0 0 0}, clip, width=.98\textwidth]{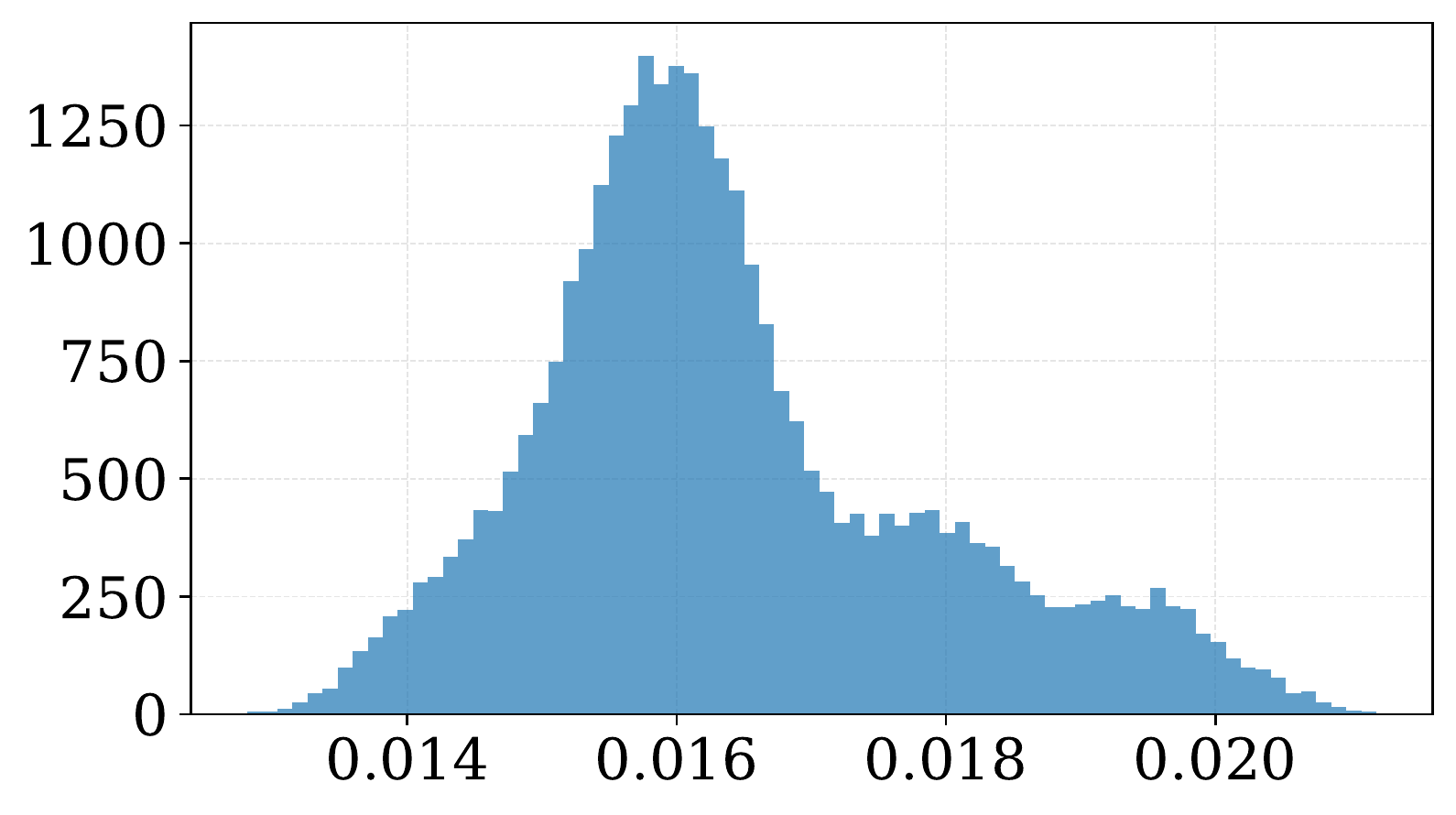}
        \caption{${f}^{\mathcal{H}}_1$: latency (seconds)\label{fig:nats_latency_hist}}
    \end{subfigure}\\
    \caption{Histogram of the objective values of architectures sampled from the NATS search space with hardware device of GPU (i.e., $\Omega$ = NATS, $\mathcal{H} = \{h_1 =\mbox{GPU}\}$). The x-axis of each subfigure shows the \emph{range} of the objective values, while the y-axis shows the number of corresponding architectures. \label{fig:badly_scale}}
\end{figure}

Considering the three types of objective functions in a general formulation of a NAS task, the corresponding MOPs are intrinsically badly scaled.
As exemplified in Fig.~\ref{fig:badly_scale}, the prediction error ($f^e$) falls into the range of $[0,1]$; by contrast, since the complexity related objectives ($\bm{f}^c$) and hardware related objectives ($\bm{f}^\mathcal{H}$) are formulated to indicate different physical quantities, their objective values are of very different scales.
Moreover, it is difficult to normalize these objective values into the same scale due to the unknown properties of various search spaces and hardware devices.

\subsection{Degenerate Pareto fronts}

In multiobjective optimization, an important assumption is that the objectives to be optimized are often in conflict, such that there does not exist a single solution achieving optima on all the objectives. 
Alternatively, an EMO algorithm aims to find a set of solutions as an approximation to the Pareto front (PF), where the solutions on the PF trade-off between the conflicting objective.
Ideally, the Pareto front (PF) of an $m$-objective MOP is an $m-1$-dimensional manifold iff all the objectives are conflicting with each other.
In practice, however, this property may not always hold since some objectives of a MOP may be positively correlated to each other, thus leading to a \emph{degenerate PF}~\cite{ishibuchi2015pareto, elarbi2019approximating}.

\begin{figure}[ht]
    \centering
    \begin{subfigure}[b]{0.24\textwidth}
        \centering
        \includegraphics[trim={0 .7cm 0 1cm}, clip, width=\textwidth]{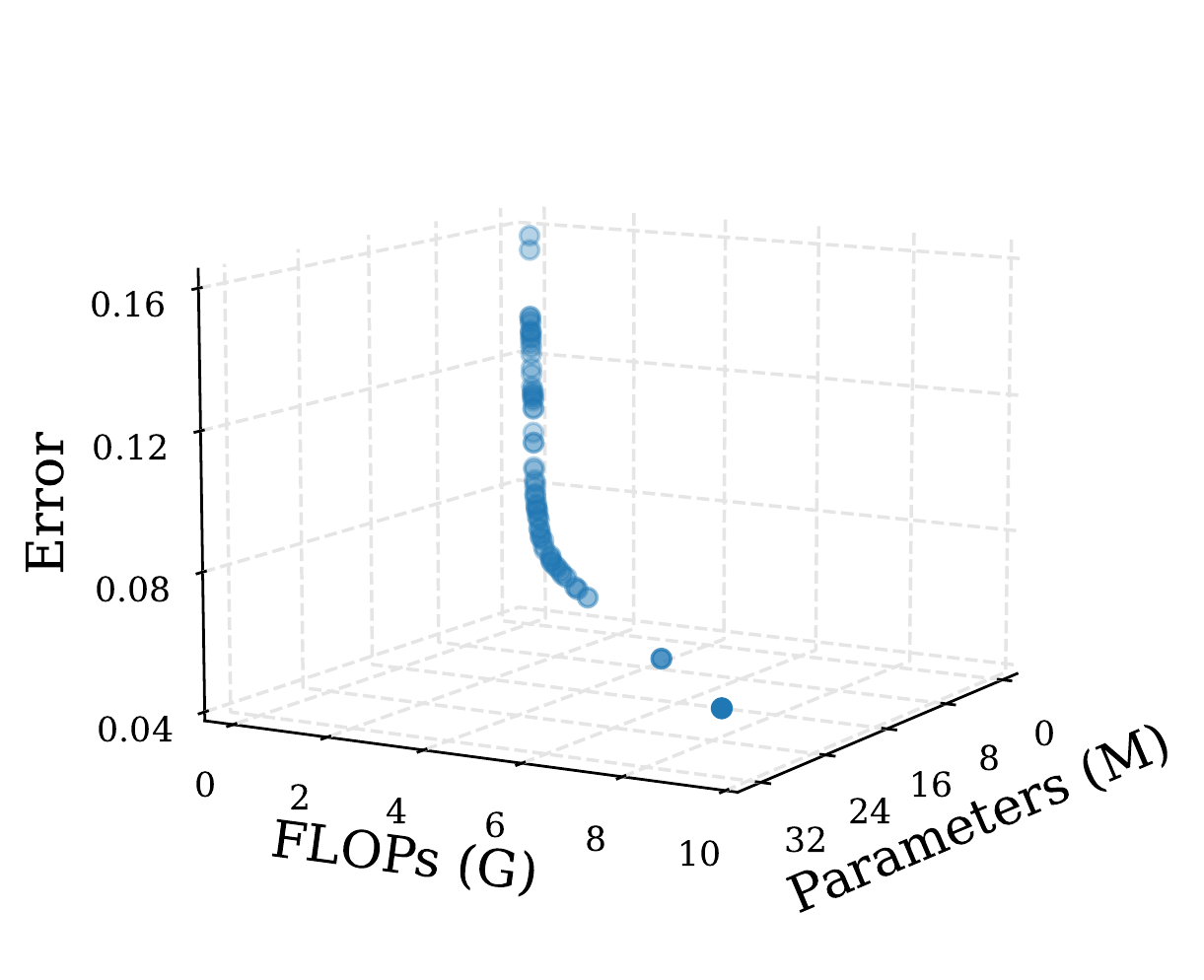}
        \caption{$\Omega = $ NB101 \label{fig:nb101_err_params_flops_sideview}}
    \end{subfigure} \hfill
    \begin{subfigure}[b]{0.24\textwidth}
        \centering
        \includegraphics[trim={0 .7cm 0 1cm}, clip, width=\textwidth]{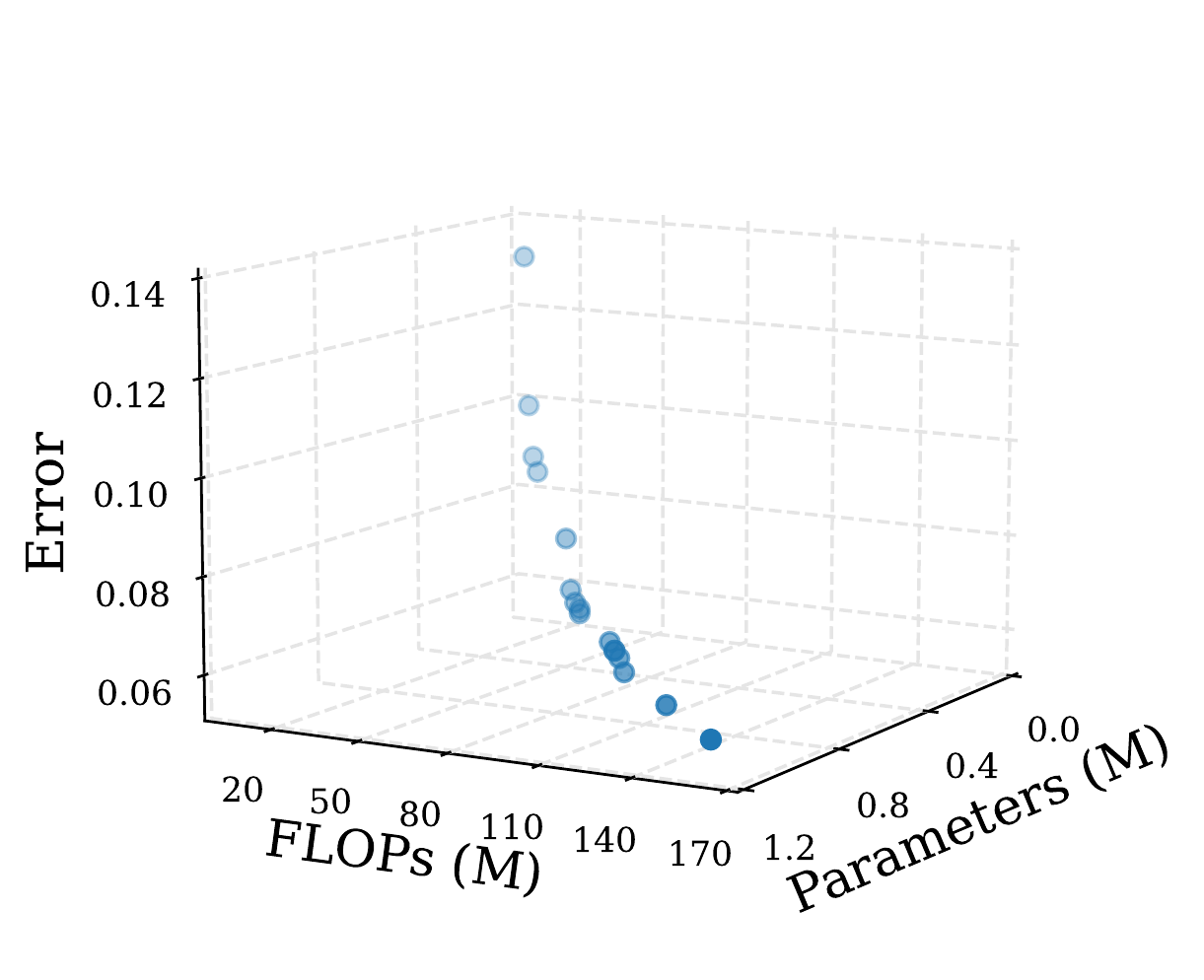}
        \caption{$\Omega = $ NB201 \label{fig:nb201_err_params_flops_sideview}}
    \end{subfigure}
    \caption{The Pareto fronts of two three-objective NAS tasks on the (a) NB101 and (b) NB201 search spaces. The three optimization objectives are: ${f}^e$: prediction error, ${f}^c_1$: number of parameters, ${f}^c_2$: number of FLOPs.
    \label{fig:degenerated_pf}}
\end{figure}

As formulated above, there are generally three types of objectives in a NAS problem: the \emph{error}-related objective $\bm{f}^e$, the \emph{complexity}-related objectives $\bm{f}^{c}$ ,and the \emph{hardware}-related objectives $\bm{f}^{\mathcal{H}}$.
Normally, $f^e$ is conflicting with $\bm{f}^{c}$ and $\bm{f}^{c}$ respectively, i.e., a model with lower prediction error usually has higher complexity and lower hardware performance.
By contrast, however, the objectives in $\bm{f}^{c}$ or $\bm{f}^{\mathcal{H}}$ may not be fully conflicting among themselves.
As exemplified in \ref{fig:nb101_err_params_flops_sideview}, ${f}^e$ is conflicting with ${f}^c_1$ and ${f}^c_2$ respectively, while ${f}^c_1$ and ${f}^c_2$ are positively correlated, thus leading to a degenerate PF (i.e. the one-dimensional curve) in the three objective space.





%% file: 4-benchmark.tex
\section{Benchmark Design\label{sec:design}}
In this section, we introduce the overall pipeline of \ourbenchmark{}, followed by detailed design principles of each main component. 

\subsection{Overview}
\begin{figure}[t]
    \centering
    \begin{subfigure}[b]{0.5\textwidth}
        \centering
        \includegraphics[trim={0 0 0 0}, clip, width=\textwidth]{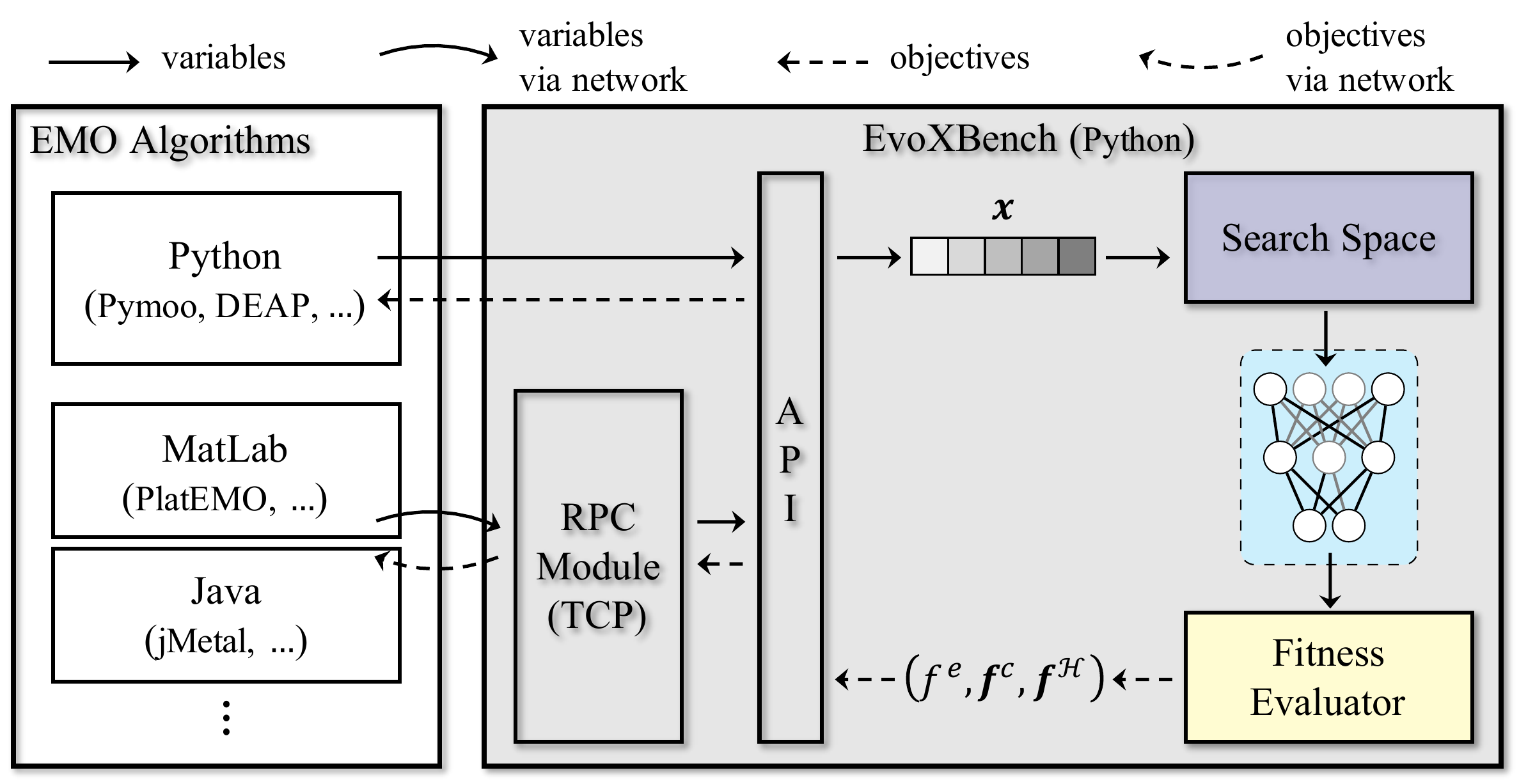}
    \end{subfigure}
    \caption{Overall pipeline of \ourbenchmark{}. See text for details. \label{fig:evoxbench_pipeline}}
\end{figure}

In order to facilitate seamless communication, an API layer is designed to handle (i) the reception of the candidate decision vectors generated by an EMO algorithm and (ii) the feedback of the fitness values evaluated by our \ourbenchmark{}. 
As shown in Fig.~\ref{fig:evoxbench_pipeline}, given a candidate decision vector received by the API layer, \ourbenchmark{} first passes it through the search space module to create its corresponding architecture; then, the architecture is processed by the fitness evaluator module; finally, the optimization objectives are routed back to the API layer for output. 
For EMO algorithms implemented in non-Python programming languages (e.g., MATLAB or Java), the communication is routed through an extra remote procedure call (RPC) module built on top of the widely-used transmission control protocol (TCP). 

\subsection{Design of Search Spaces}
As formulated in (\ref{def:formulation1}), a search space (a.k.a. decision space) $\Omega$ defines all attainable solutions to an optimization algorithm. 
In general, every search space follows the schema (outlined in Algorithm~1 from supplementary materials). 
Specifically, a string that defines a DNN architecture is referred to as a \emph{phenotype}, from which an actual DNN model can be created; 
an integer-valued vector is referred to as a \emph{genotype} on which genetic operators, such as crossovers and mutations, are carried out; and the interfaces between genotypes and phenotypes are referred to as \emph{encode} and \emph{decode}, respectively. 

\begin{figure}[t]
    \centering
    \begin{subfigure}[b]{0.42\textwidth}
        \centering
        \includegraphics[trim={0 0 1cm 0}, clip, width=.98\textwidth]{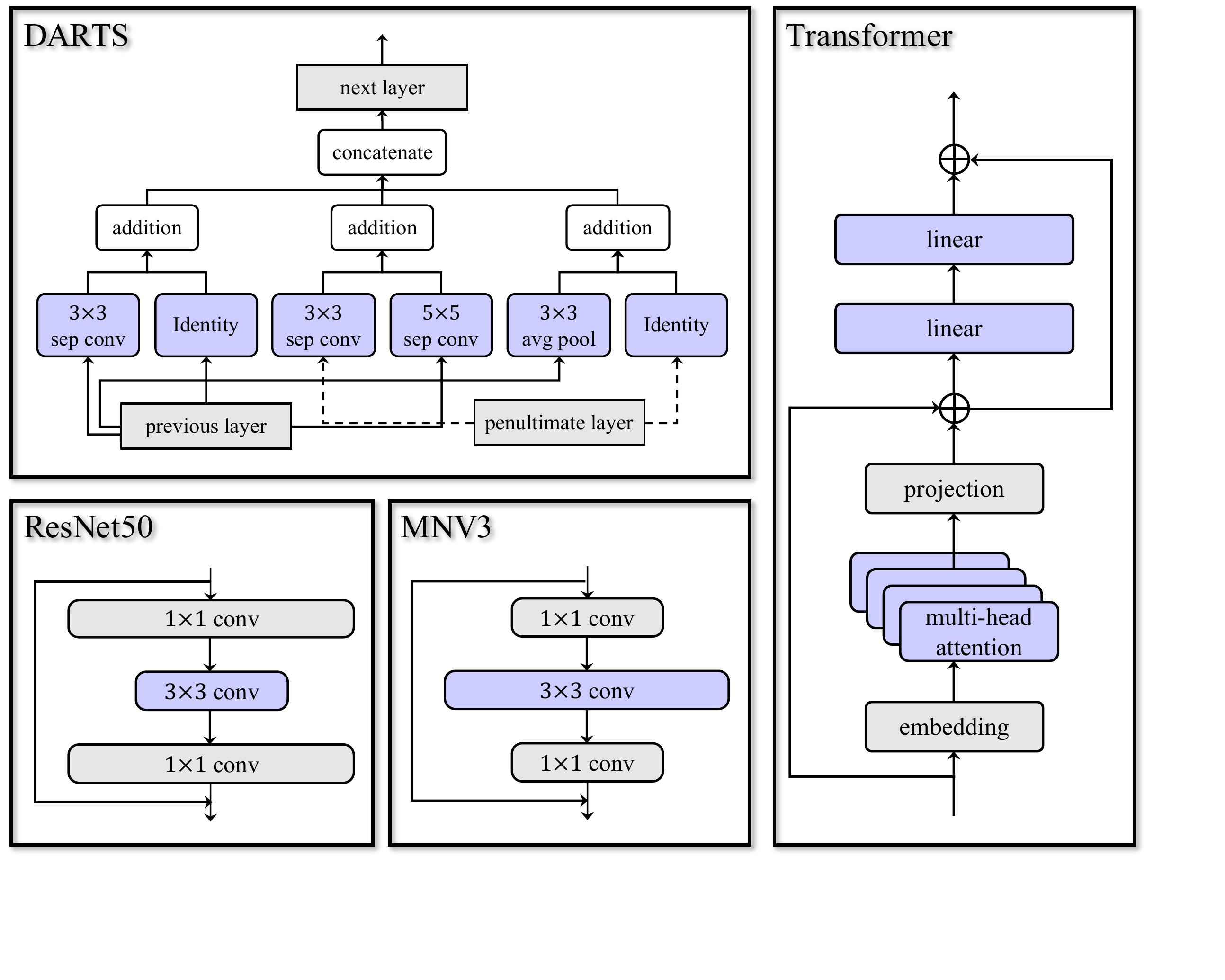}
    \end{subfigure}
    \caption{Visualization of architectures from search spaces supported in \ourbenchmark{}. \label{fig:search_space_instances}}
\end{figure}

In \ourbenchmark{}, we currently consider seven search spaces for image classification on CIFAR-10 and ImageNet datasets, covering both convolutional DNNs and Transformers from micro and macro search spaces.
A summary of these search spaces is provided in Table~\ref{tab:search_space_overview}, and visualizations of architectures from these search spaces are provided in Fig.~\ref{fig:search_space_instances}.

It is worth noting that, apart from the existing seven search spaces, any other user-specific search space can also be easily incorporated into \ourbenchmark{} by following a routine pipeline.
Due to the page limit, readers are referred to Section IV in the supplementary materials for more details.

\begin{table}[ht]
\centering
\caption{An overview of the search spaces supported in \ourbenchmark{}. The fitness landscapes of the search spaces from the top section are completely known via exhaustive evaluations, while the fitness landscapes of the search spaces from the bottom section are approximated via surrogate models trained on 60$K$ uniformly sampled architectures. \label{tab:search_space_overview}}
\resizebox{.47\textwidth}{!}{%
\begin{tabular}{@{\hspace{2mm}}cccccc@{\hspace{2mm}}}
\toprule
$\Omega$ & Type & Dataset & $D$ & $\vert\Omega\vert$ & Objectives\\ \midrule
NB101\cite{ying2019bench}  & micro & C-10 & 26 & 423K & $f^e$, $\bm{f}^{c}$ \\
NB201\cite{Dong2020NAS-Bench-201,li2021hwnasbench} & micro & C-10 & 6 & 15.6K & $f^e$, $\bm{f}^{c}$, $\bm{f}^{h_1}$ \\
NATS\cite{dong2021nats} & macro & C-10 & 5 & 32.8K & $f^e$, $\bm{f}^{c}$, ${f}^{h_1}$\\ \midrule
DARTS\cite{zela2022surrogate} & micro & C-10 & 32 & $\sim10^{21}$ & $f^e$, $\bm{f}^{c}$ \\
ResNet50\cite{Cai2020Once-for-All} & macro & IN-1K & 25 & $\sim10^{14}$ & $f^e$, $\bm{f}^{c}$\\
Transformer\cite{chen2021autoformer} & macro & IN-1K & 34 & $\sim10^{14}$ & $f^e$, $\bm{f}^{c}$ \\
MNV3\cite{Cai2020Once-for-All} & macro & IN-1K & 21 & $\sim10^{20}$ & $f^e$, $\bm{f}^{c}$, ${f}^{h_1}$\\
\bottomrule
\end{tabular}%
}
\end{table}

\subsection{Design of Fitness Evaluator}
As formulated in (\ref{def:formulation1}), we consider three categories of objectives, i.e., the prediction error objective ($f^e$), the model complexity related objectives ($\bm{f}^{c}$) and hardware efficiency ($\bm{f}^{\mathcal{H}}$). 
Dedicated methods are devoted to handling these objectives efficiently and reliably. 

\begin{figure}[ht]
    \centering
    \begin{subfigure}[b]{0.49\textwidth}
        \centering
        \includegraphics[trim={0 0 0 0}, clip, width=\textwidth]{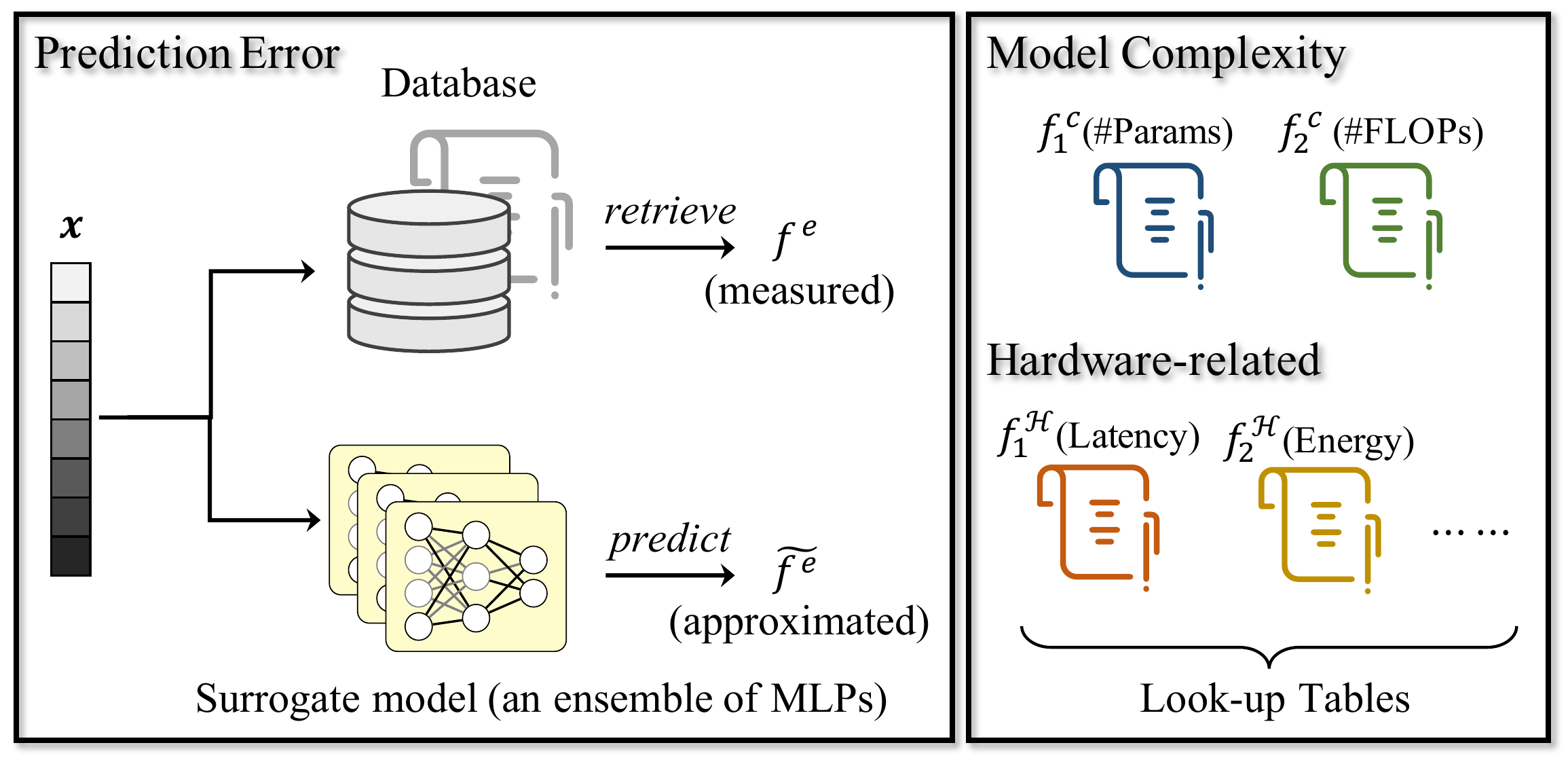}
    \end{subfigure}
    \caption{Overview of fitness evaluators in \ourbenchmark{}.\label{fig:evoxbench_evaluator}}
\end{figure}

\subsubsection{Evaluation of $f^e$}
As illustrated in Section~\ref{sec:noisy_objs}, $f^e$ is a noisy optimization objective. 
To simulate the noise in evaluating $f^e$, the following two strategies are employed:

\vspace{2pt}
\noindent\textbf{--}
 For NB101, NB201, and NATS search spaces, we leverage the exhaustive evaluation results provided by the original papers \cite{ying2019bench,Dong2020NAS-Bench-201,li2021hwnasbench,dong2021nats}, where all solutions (architectures) are thoroughly trained from scratch with three repetitions. 
 On top of these results, we construct a unified database solely based on Python. 
 In other words, the $f^e$ values of solutions from these search spaces can be queried via a canonical interface without configurations of sophisticated software such as TensorFlow or PyTorch. 
Thereafter, each evaluation value of $f^{e}$ is randomly selected from the three repetitions of the corresponding evaluation results stored in the database.

\vspace{2pt}
\noindent\textbf{--}
Considering the total volume of the DARTS, ResNet50, Transformer, and MNV3 search spaces, we resort to surrogate models for evaluating $f^e$. 
We choose a multi-layer perceptron (MLP) as our surrogate model in this work. 
Apart from its well-established track record for predicting the $f^e$ values of DNNs accurately \cite{Cai2020Once-for-All,xu2021renas,dai2021fbnetv3}, implementing an MLP (i.e., a loop of matrix multiplications and additions) is straightforward and canonical in most programming languages without additional dependencies. 
Furthermore, we construct a pool of MLPs from $k$-fold cross-validation of the training data, where $k$ is set to ten in this work. Thereafter, each evaluation value of $f^e$ is predicted by a single MLP randomly selected from the pool. 

For generating samples (i.e., variable-objective pairs) to train a MLP, we utilize the \emph{supernet} idea that has emerged as a standard technique in state-of-the-art (SotA) NAS methods \cite{ren2021comprehensive}. Specially, first, we follow the progressive shrinking algorithm \cite{Cai2020Once-for-All} to train a supernet from which the optimal weights of a candidate architecture solution (i.e., $\bm{w}^*(\bm{x})$ in Eq~\ref{def:formulation1}) can be directly inherited without the costly iterations of SGD\footnote{SGD is a standard method for solving the inner optimization problem in Eq~\ref{def:formulation1}, i.e., $\min~\mathcal{L}_{trn}(\bm{x}; \bm{w})$.}; then, we use the weights provided by the trained supernet to evaluate the $f^e$ values of the samples. As opposed to generating samples by independently training every one of them from scratch, this pipeline offers two appealing properties: (i) it is more scalable to a large number of samples which is crucial for learning an accurate approximation\footnote{Assuming that the training cost of a large number of sample architectures, 60$K$ in our case, is much greater than the training cost of a single supernet.}; (ii) the post-search re-training of the obtained architectures can be avoided as the corresponding weights are readily available from the supernet. 

\begin{figure}[ht]
    \centering
    \begin{subfigure}[b]{0.49\textwidth}
        \centering
        \includegraphics[trim={0 0 0 0}, clip, width=.32\textwidth]{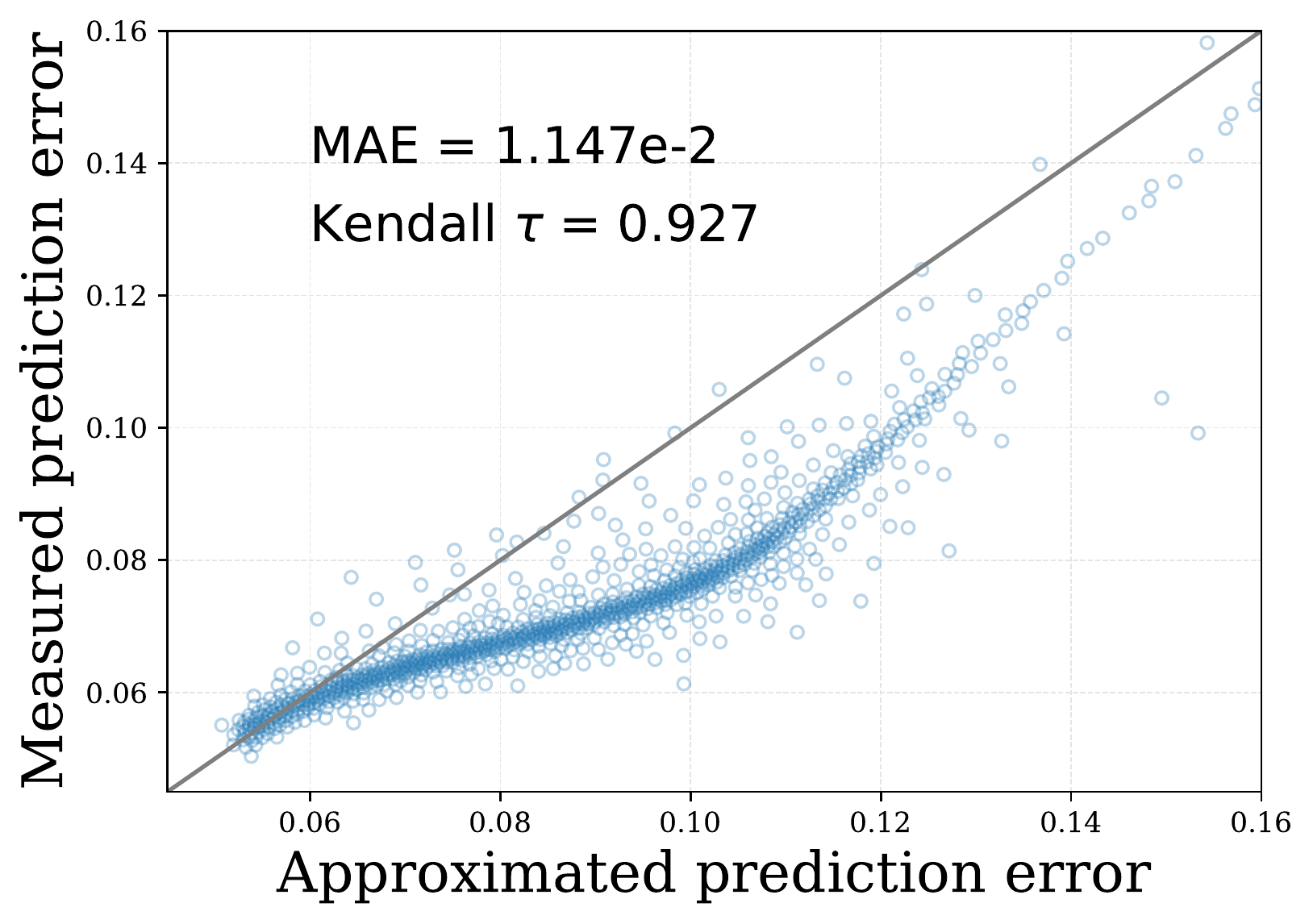} \hfill
        \includegraphics[trim={0 0 0 0}, clip, width=.32\textwidth]{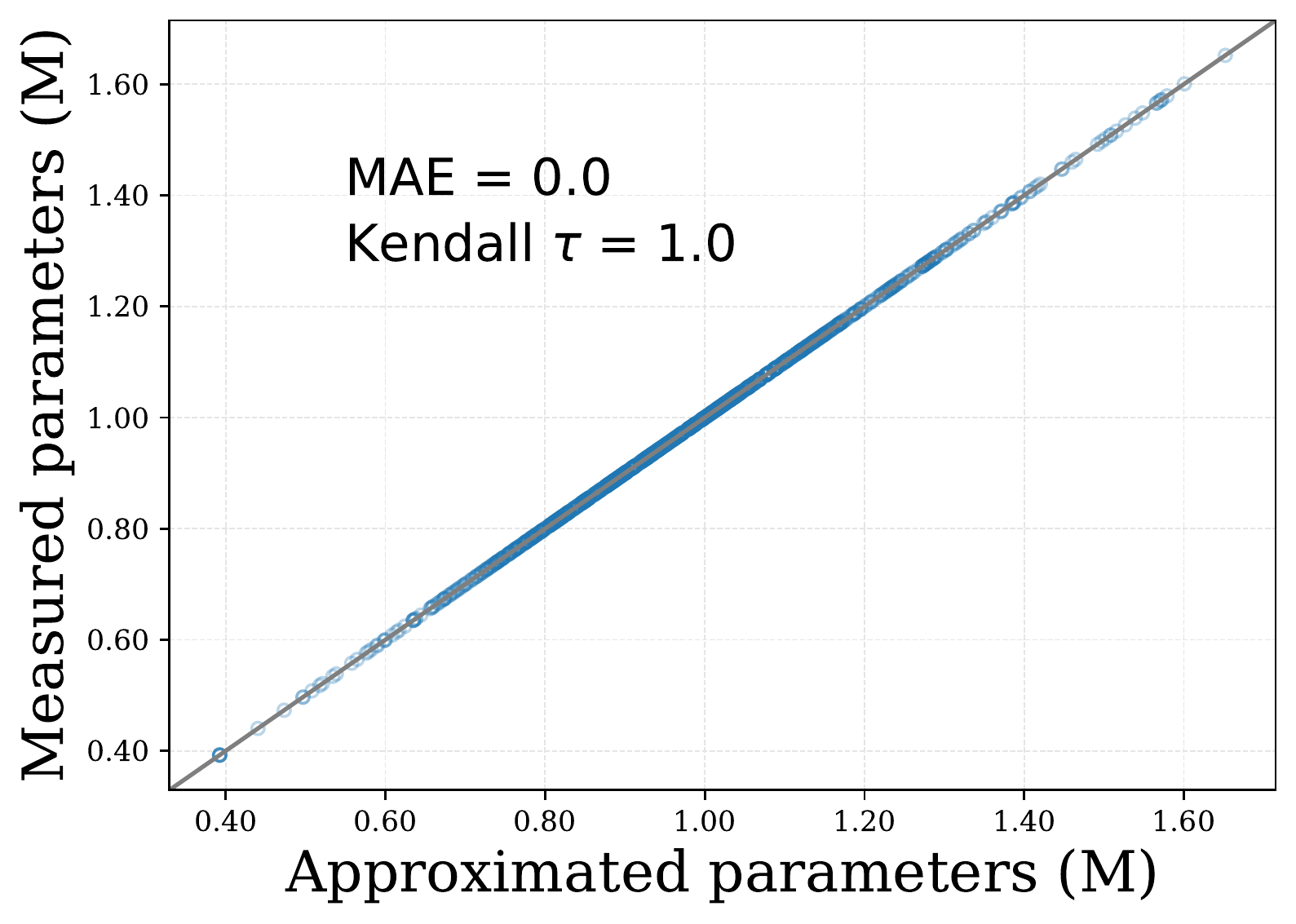} \hfill
        \includegraphics[trim={0 0 0 0}, clip, width=.32\textwidth]{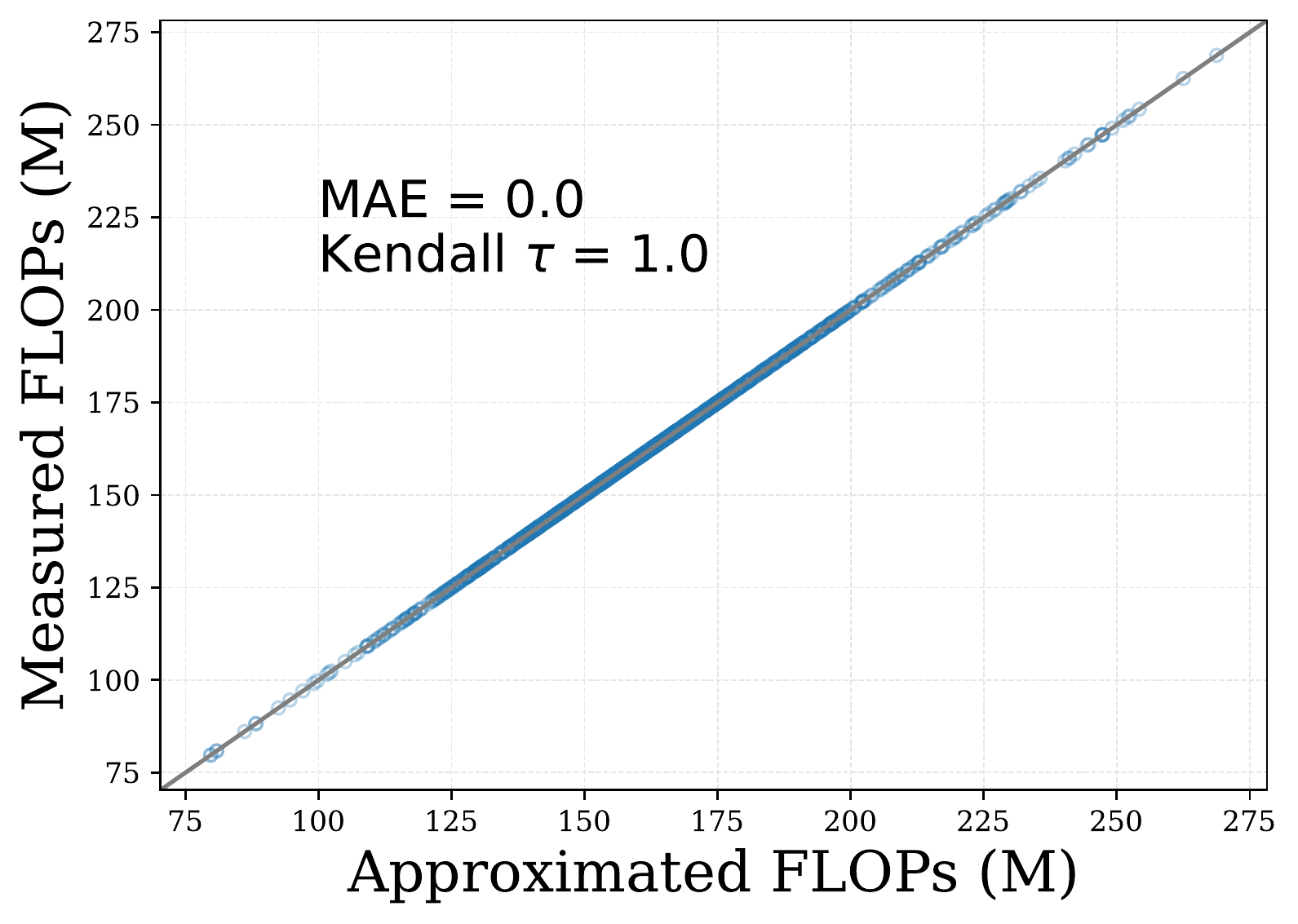}
        \vspace{-.5em}
        \caption{DARTS search space\label{fig:datrs_surrogate}}
    \end{subfigure} \\ \vspace{.5em}
    \centering
    \begin{subfigure}[b]{0.49\textwidth}
        \centering
        \includegraphics[trim={0 0 0 0}, clip, width=.32\textwidth]{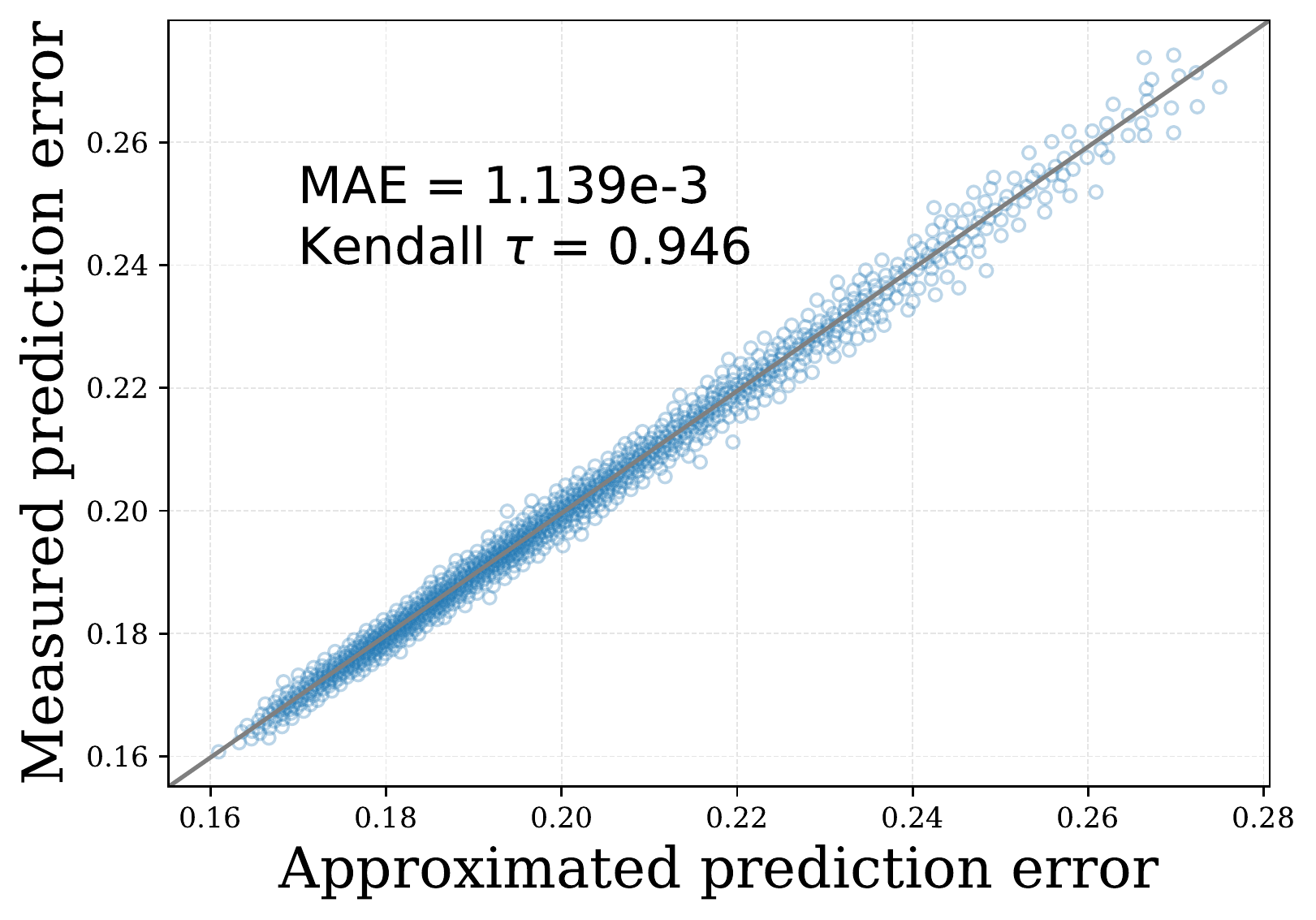} \hfill
        \includegraphics[trim={0 0 0 0}, clip, width=.32\textwidth]{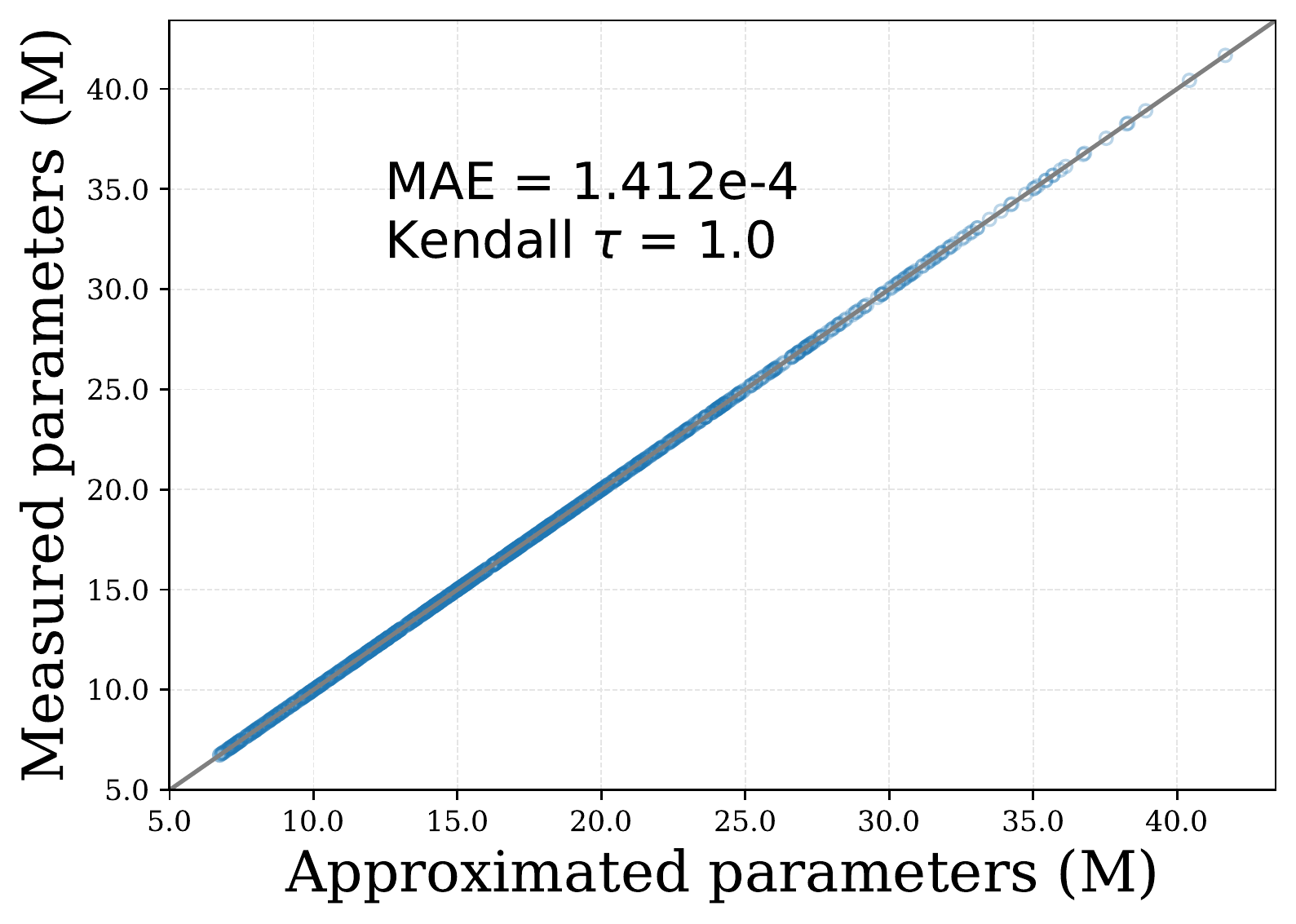} \hfill
        \includegraphics[trim={0 0 0 0}, clip, width=.32\textwidth]{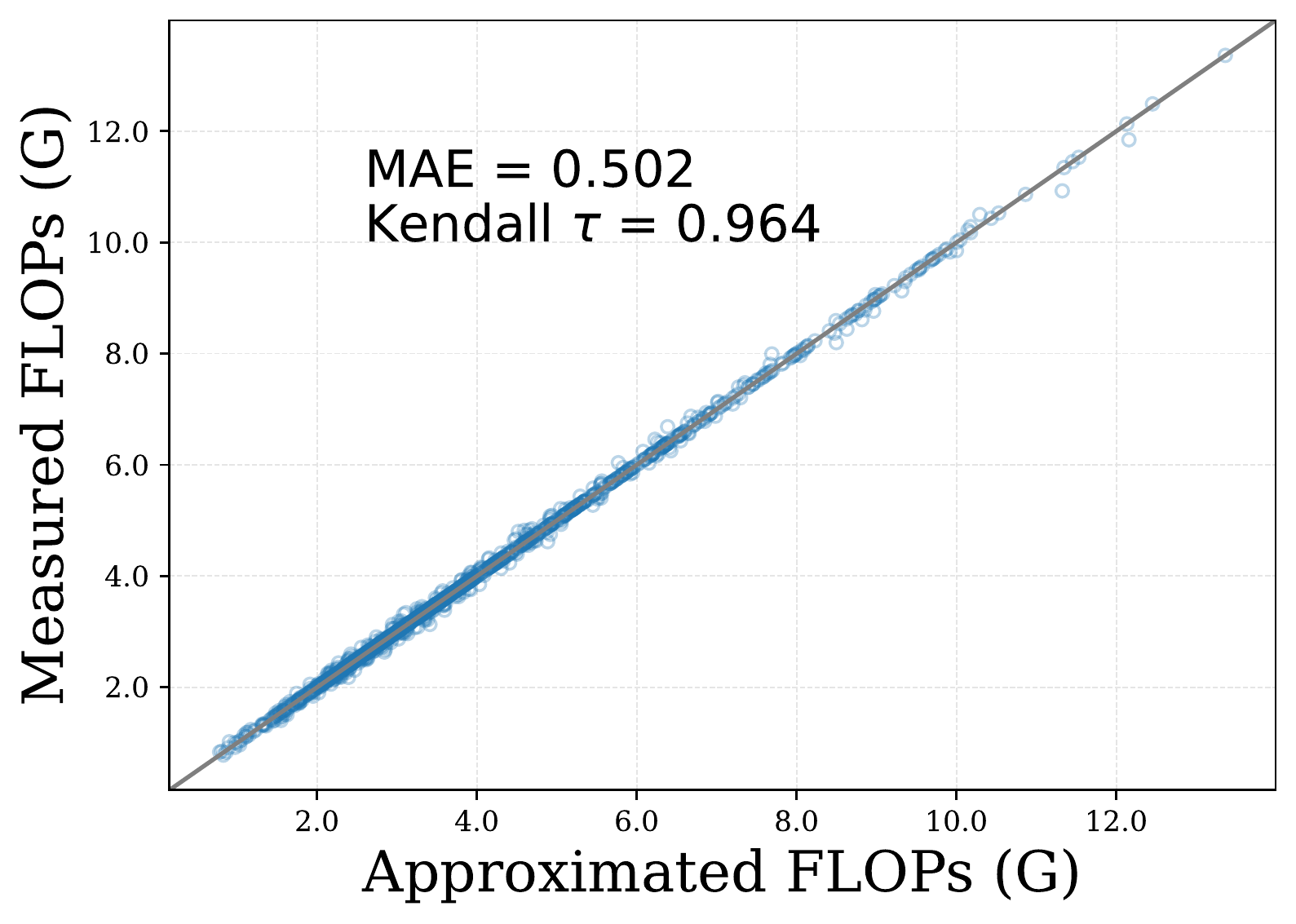}
        \vspace{-.5em}
        \caption{ResNet50 search space\label{fig:r50_surrogate}}
    \end{subfigure} \\ \vspace{.5em}
    \centering
    \begin{subfigure}[b]{0.49\textwidth}
        \centering
        \includegraphics[trim={0 0 0 0}, clip, width=.32\textwidth]{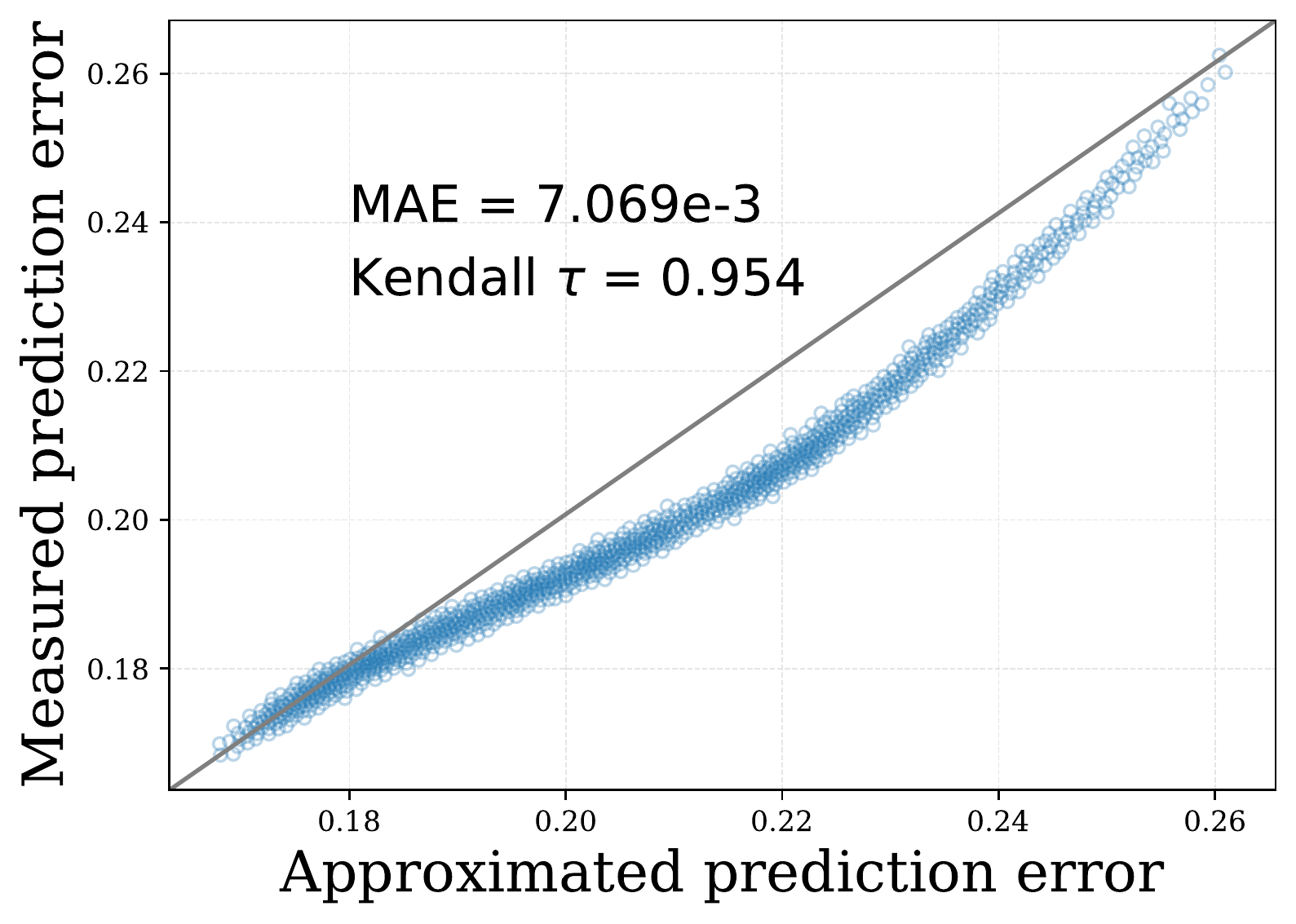} \hfill
        \includegraphics[trim={0 0 0 0}, clip, width=.32\textwidth]{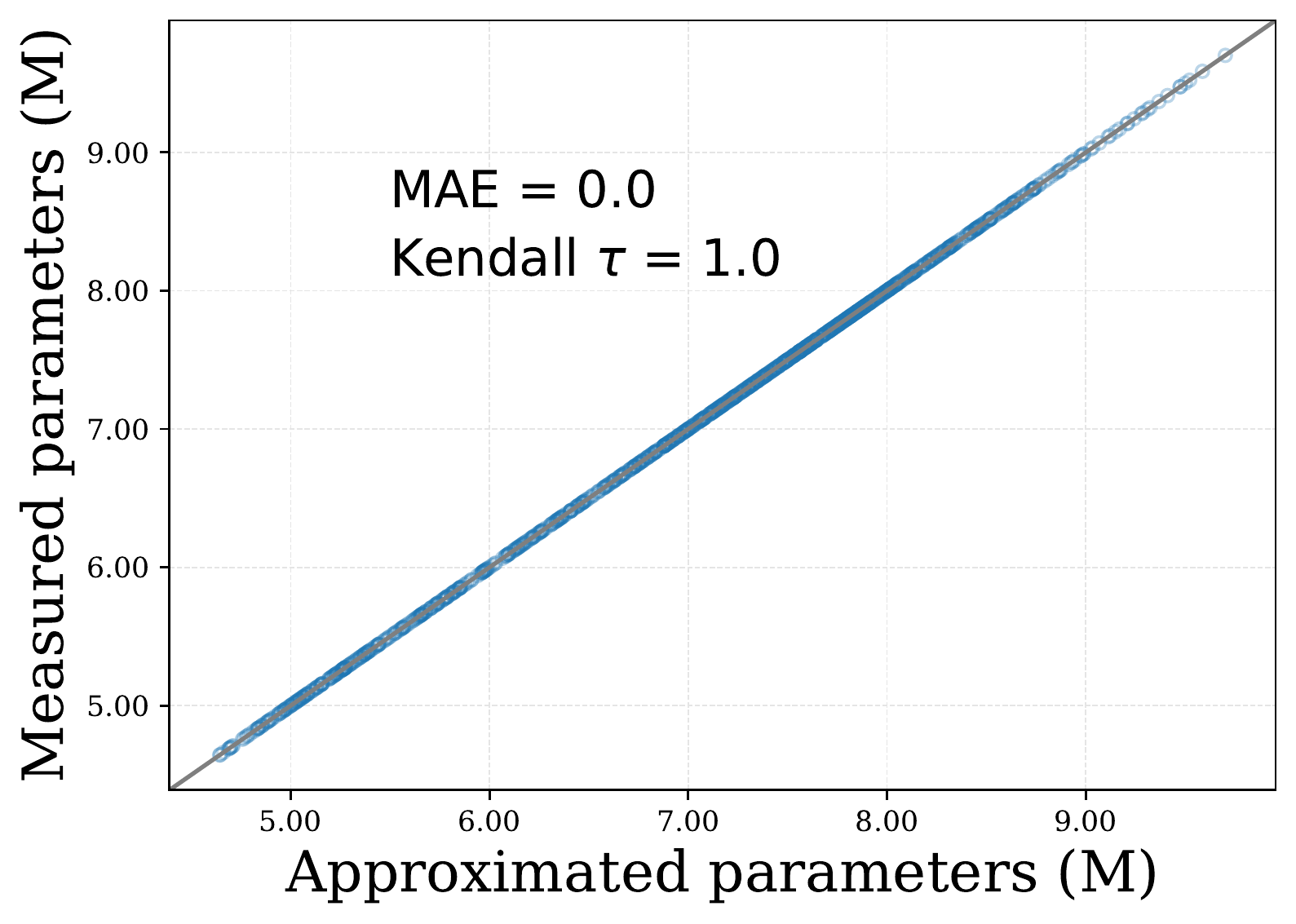} \hfill
        \includegraphics[trim={0 0 0 0}, clip, width=.32\textwidth]{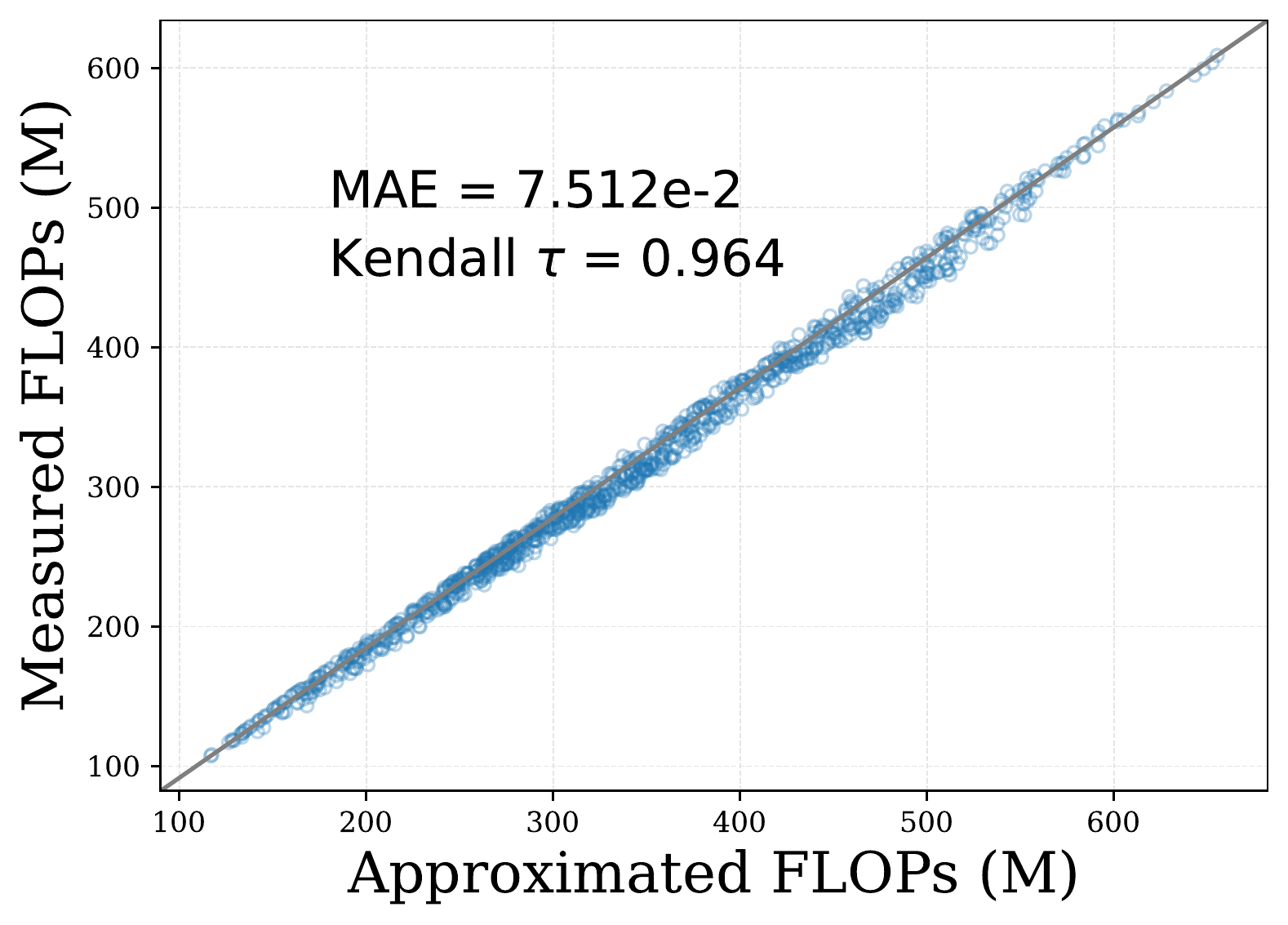}
        \vspace{-.5em}
        \caption{MNV3 search space\label{fig:mnv3_surrogate}}
    \end{subfigure}
    \caption{Performance of our fitness evaluators. For each search space, we visualize the correlation between real measurements and approximations by our surrogate model for prediction error ($f^{e}$), look-up tables for the number of parameters (${f}_1^{c}$) and FLOPs (${f}_2^{c}$), from left to right. The mean absolute error (MAE) and Kendall rank correlation values are annotated in every sub-figure. \label{fig:surrogate_performance}}
\end{figure}

A pictorial illustration of the $f^e$ evaluators in \ourbenchmark{} is provided in the top part of Fig.~\ref{fig:evoxbench_evaluator}. And the performance of the adopted surrogate model, an ensemble of MLPs, is provided in Fig.~\ref{fig:surrogate_performance}. 

\subsubsection{Evaluation of $\bm{f}^{c}$ and $\bm{f}^{\mathcal{H}}$}
Similar to the evaluation of $f^e$, we also leverage the exhaustive evaluation history provided by the original papers \cite{ying2019bench,Dong2020NAS-Bench-201,li2021hwnasbench,dong2021nats} for NB101, NB201 and NATS search spaces. These $\bm{f}^{c}$ and $\bm{f}^{\mathcal{H}}$ values are, again, stored in a unified database along with $f^e$ values. 

For the DARTS, ResNet50, Transformer, and MNV3 search spaces, we opt for the \emph{loop-up table} route---a standard approach among existing NAS methods where an exhaustive exploration of the search space is not possible \cite{cai2018proxylessnas,li2021hwnasbench}. A pictorial overview is provided in the bottom part of Fig.~\ref{fig:evoxbench_evaluator}. In general, on one hand, the efficiency of a cell structure can be decomposed into the accumulative efficiency of the participated operations for a micro search space (i.e., DARTS); while on the other hand, the efficiency of a network can be decomposed into the accumulative efficiency of each layer for macro search spaces (i.e., ResNet50, Transformer, and MNV3). 

Under such assumptions, there exist limited operations and layer variations to enumerate, respectively for micro and macro search spaces. Hence, we can exhaustively evaluate the $\bm{f}^{c}$ and $\bm{f}^{\mathcal{H}}$ values of all operations and layer variations. Afterward, the efficiency of any given operation or layer variation can be queried from a look-up table with a designated key. Then the $\bm{f}^{c}$ and $\bm{f}^{\mathcal{H}}$ values of an architecture solution can be obtained by summing over all operations or layer variations. 

\subsection{Design of RPC Module}

As depicted in Fig.~\ref{fig:evoxbench_pipeline}, EMO algorithms written in Python can call \ourbenchmark{} directly, while queries from EMO algorithms in other programming languages are processed through the remote procedure call (RPC) module. 
For compatibility reasons, the communication functionalities of RPC are built upon the transmission control protocol (TCP). 
Technically, it allows programs written in almost any programming language on the operating system to call \ourbenchmark{} via localhost. 
Furthermore, by setting up \ourbenchmark{} once on a central server, it allows additional users to use \ourbenchmark{} remotely as a web service without any local installation/configuration. 

More specifically, after establishing a successful TCP connection, EMO algorithms (as the users) and \ourbenchmark{} communicate by exchanging JSON strings. 
Since this protocol itself is stateless, we assign an integer string to every object created in \ourbenchmark{} as its unique identifier. 
And this identifier needs to be included as a part of the JSON strings to facilitate the communication between users and \ourbenchmark{}.  

As depicted in Table~\ref{tab:evoxbench_speed}, we empirically demonstrate that real-time feedback of fitness values is archived by \ourbenchmark{} via the RPC module. 
Specifically, the evaluation of an architecture in \ourbenchmark{} is facilitated by either querying a database or a surrogate model. 
On one hand, querying a database is intrinsically fast; 
on the other hand, the surrogate model used in \ourbenchmark{} is a simple yet effective three-layer perceptron.
Additionally, we batch all evaluations in a matrix to further boost the speed. 
Readers may refer to supplementary materials for more technical details. 

\begin{table}[ht]
\centering
\caption{Total time cost (in \textbf{seconds})  for calling \ourbenchmark{} to evaluate $\textbf{1,000}$ solutions by EMO algorithms implemented in three different programming languages on a local machine. Results are averaged over 31 runs on a single \textbf{CPU} core.\label{tab:evoxbench_speed}}
\resizebox{.48\textwidth}{!}{%
\begin{tabular}{@{\hspace{2mm}}ccccc@{\hspace{2mm}}}
\toprule
Search space & \begin{tabular}[c]{@{}c@{}}Query\\ method\end{tabular} & Python & MATLAB & Java \\ \midrule
NB101 & Database & 0.1139 $\pm$ 0.013 & 0.1716 $\pm$ 0.031 & 0.1970 $\pm$ 0.036 \\
MNV3 & Surrogate & 0.0380 $\pm$ 0.002 & 0.0528 $\pm$ 0.018 & 0.0574 $\pm$ 0.020 \\ \bottomrule
\end{tabular}%
}
\end{table}




%% file: 5-testsuite.tex
\section{Benchmark Test Suite Generation\label{sec:test_suite}}
On the basis of \ourbenchmark{}, specific test suites (i.e., collections of test instances/problems) can be generated by combining search spaces and their supported objectives. For the purpose of demonstration, we generate two tailored multi-objective NAS test suites for image classification on datasets CIFAR-10 and ImageNet respectively.
Following the formulations in Section~\ref{sec:formulation}, the test instances are defined by specifying the search space, the hardware, and the metrics for measuring model complexity and hardware efficiency. 
In the remaining of this section, we present these two test suites in detail. 

\subsection{\firsttestsuite{} Test Suite}

As summarized in Table~\ref{tab:c10mop}, there are nine instances in \firsttestsuite{} tailored for image classification on CIFAR-10, considering various search spaces and hardware devices. 
The test instances are arranged in the order of ascending number of objectives, i.e., from two to eight objectives. 
For most test instances, we consider only one targeted hardware (i.e., GPUs or Eyeriss \cite{chen2016eyeriss}) except \firsttestsuite{}7 where both hardware are considered simultaneously. 

\begin{table}[ht]
\centering
\caption{Definition of the proposed \firsttestsuite{} test suite.\label{tab:c10mop}}
\resizebox{.49\textwidth}{!}{%
\begin{threeparttable}
\begin{tabular}{@{\hspace{2mm}}ccccc@{\hspace{2mm}}}
\toprule
 \begin{tabular}[c]{@{}c@{}}Problem\end{tabular} & \begin{tabular}[c]{@{}c@{}}$\Omega$\end{tabular} & $D$ & $M$ & Objectives \\ \midrule
    \firsttestsuite{}1 & NB101  & 26 & 2 & $f^e$, $f_1^c$ \\
    \firsttestsuite{}2 & NB101  & 26 & 3 & $f^e$, $f_1^c$, $f_2^c$ \\
    \firsttestsuite{}3 & NATS & 5 & 3 & $f^e$, $f_1^c$, $f_2^c$ \\
    \firsttestsuite{}4 & NATS & 5 & 4 & $f^e$, $f_1^c$, $f_2^c$, $f_1^{h_1}$ \\
    \firsttestsuite{}5 & NB201 & 6 & 5 & $f^e$, $f_1^c$, $f_2^c$, $f_1^{h_1}$, $f_2^{h_1}$ \\
    \firsttestsuite{}6 & NB201  & 6 & 6 & $f^e$, $f_1^c$, $f_2^c$, $f_1^{h_2}$, $f_2^{h_2}$, $f_3^{h_2}$ \\
    \firsttestsuite{}7 & NB201  & 6 & 8 & $f^e$, $f_1^c$, $f_2^c$, $f_1^{h_1}$, $f_2^{h_1}$, $f_1^{h_2}$, $f_2^{h_2}$, $f_3^{h_2}$ \\
    \firsttestsuite{}8 & DARTS  & 32 & 2 & $^{\dagger}f^e$, $f_1^c$ \\
    \firsttestsuite{}9 & DARTS & 32 & 3 & $^{\dagger}f^e$, $f_1^c$, $f_2^c$ \\ 
\bottomrule
\end{tabular}%
\begin{tablenotes}
\setlength\labelsep{0pt}
\small
\item \hspace{1em}$^{\dagger}$ indicates that $f^e$ (validation error) is based on surrogate modeling. 
\item \hspace{1em}$^{\ddagger}$ hardware $h_1 = \mbox{GPU}$ and $h_2 = \mbox{Eyeriss}$.
\end{tablenotes}
\end{threeparttable}
}
\end{table}

The properties of the test instances are summarized in Table~\ref{tab:c10mop_properties} in correspondence with the characters illustrated in Section~\ref{sec:formulation}. 
In general, the fitness landscapes of prediction error ($f^{e}$) are noisy and multi-modal, and the Pareto fronts are degenerated for most test instances.
Since the fitness landscape of \firsttestsuite{}1 to \firsttestsuite{}7 are completely known via exhaustive samples, the ranges of all objectives are known \emph{a priori} and thus can be properly scaled
By contrast, the objectives of \firsttestsuite{}8 and \firsttestsuite{}9 remain un-normalized (i.e., badly scaled) as the ranges cannot be estimated via exhaustive samples.

\begin{table}[ht]
\centering
\caption{Property of test instances in \firsttestsuite{}.\label{tab:c10mop_properties}}
\resizebox{.45\textwidth}{!}{%
\begin{tabular}{@{\hspace{2mm}}cccccc@{\hspace{2mm}}}
\toprule
    Problem & \begin{tabular}[c]{@{}c@{}}Multi-\\ modal\end{tabular} & \begin{tabular}[c]{@{}c@{}}Many \\ objectives\end{tabular} & \begin{tabular}[c]{@{}c@{}}Noisy \\ objectives\end{tabular} & \begin{tabular}[c]{@{}c@{}}Badly-\\ scaled\end{tabular} & \begin{tabular}[c]{@{}c@{}}Degenerated\\ PF\end{tabular} \\ \midrule
    \firsttestsuite{}1 & $\checkmark$ &  & $\checkmark$ &  & \\
    \firsttestsuite{}2 & $\checkmark$ &  & $\checkmark$ &  & $\checkmark$\\
    \firsttestsuite{}3 &  &  &  &  & $\checkmark$\\
    \firsttestsuite{}4 &  & $\checkmark$ &  &  & $\checkmark$\\
    \firsttestsuite{}5 & $\checkmark$ & $\checkmark$ & $\checkmark$ &  & $\checkmark$ \\
    \firsttestsuite{}6 & $\checkmark$ & $\checkmark$ & $\checkmark$ &  & $\checkmark$\\
    \firsttestsuite{}7 & $\checkmark$ & $\checkmark$ & $\checkmark$ &  & $\checkmark$\\
    \firsttestsuite{}8 & $\checkmark$ &  & $\checkmark$ & $\checkmark$ &  \\
    \firsttestsuite{}9 & $\checkmark$ &  & $\checkmark$ & $\checkmark$ &  \\ \bottomrule
\end{tabular}%
}
\end{table}

\subsection{\secondtestsuite{} Test Suite}
The \secondtestsuite{} test suite also comprises nine test instances. 
Since the search spaces of ResNet50 and Transformer are resource-intensive and thus unsuitable for efficient hardware deployment, we do not consider hardware-related objectives as architectures from them.
As a result, most test instances are multiobjective problems except \secondtestsuite{}9 where an additional objective of the latency on a mobile phone is considered. A summary of \secondtestsuite{} test suite is provided in Table~\ref{tab:in1kmop}.

\begin{table}[ht]
\centering
\caption{Definition of the proposed \secondtestsuite{} test suite.\label{tab:in1kmop}}
\resizebox{.44\textwidth}{!}{%
\begin{threeparttable}
\begin{tabular}{@{\hspace{1.2em}}cccc@{\hspace{1em}}c@{\hspace{1em}}c@{\hspace{1.2em}}}
\toprule
\begin{tabular}[c]{@{}c@{}}Problem\end{tabular} & \begin{tabular}[c]{@{}c@{}}$\Omega$\end{tabular} & $D$ & $M$ & Objectives \\ \midrule
    \secondtestsuite{}1 & ResNet50 & 25 & 2 & $f^e$, $f_1^c$ \\
    \secondtestsuite{}2 & ResNet50 & 25 & 2 & $f^e$, $f_2^c$ \\
    \secondtestsuite{}3 & ResNet50 & 25 & 3 & $f^e$, $f_1^c$, $f_2^c$ \\
    \secondtestsuite{}4 & Transformer & 34 & 2 & $f^e$, $f_1^c$ \\
    \secondtestsuite{}5 & Transformer & 34 & 2 & $f^e$, $f_2^c$ \\
    \secondtestsuite{}6 & Transformer & 34 & 3 & $f^e$, $f_1^c$, $f_2^c$ \\
    \secondtestsuite{}7 & MNV3 & 21 & 2 & $f^e$, $f_1^c$ \\
    \secondtestsuite{}8 & MNV3 & 21 & 3 & $f^e$, $f_1^c$, $f_2^c$ \\
    \secondtestsuite{}9 & MNV3 & 21 & 4 & $f^e$, $f_1^c$, $f_2^c$, $f_1^{h_1}$ \\ \bottomrule
\end{tabular}%
\begin{tablenotes}
\setlength\labelsep{0pt}
\small
\item $^{\dagger}$ All $f^e$ (validation errors) are based on surrogate modeling. 
\item $^{\ddagger}$ hardware $h_1 = \mbox{Samsung Note10}$.
\end{tablenotes}
\end{threeparttable}
}
\end{table}

The properties of the test instances of \secondtestsuite{} are summarized in Table~\ref{tab:in1kmop_properties}. 
Given the volume of the involved search spaces, it is impossible to evaluate every attainable solution. 
Accordingly, for each test instance, the fitness landscape of is approximated via a surrogate model, and the objectives are badly-scaled (i.e., without normalization).
Note that the objectives can be normalized to the bounds derived from the sampled architectures (for building the surrogate models). 
However, due to the fact that the sampled architectures may not exactly cover the full range of the search spaces, the  ``normalized'' objectives (according to the bounds derived from the sampled architectures) may not be strictly between zero and one. 
Accordingly, we opt for leaving the objectives as un-normalized. 
The character of multi-modality for all test instances is verified on the basis of extensive optimization using a large number of fitness evaluations.

\begin{table}[ht]
\centering
\caption{Property of test instances in \secondtestsuite{}.\label{tab:in1kmop_properties}}
\resizebox{.4\textwidth}{!}{%
\begin{tabular}{@{\hspace{2mm}}ccccc@{\hspace{2mm}}}
\toprule
    Problem & \begin{tabular}[c]{@{}c@{}}Multi-\\ modal\end{tabular} & \begin{tabular}[c]{@{}c@{}}Many \\ objectives\end{tabular} & \begin{tabular}[c]{@{}c@{}}Noisy \\ objectives\end{tabular} & \begin{tabular}[c]{@{}c@{}}Badly-\\ scaled\end{tabular} 
    \\ \midrule
    \secondtestsuite{}1 & $\checkmark$ &  & $\checkmark$  & $\checkmark$ \\
    \secondtestsuite{}2 & $\checkmark$ &  & $\checkmark$  & $\checkmark$ \\
    \secondtestsuite{}3 & $\checkmark$ &  & $\checkmark$  & $\checkmark$  \\
    \secondtestsuite{}4 &  &  & $\checkmark$  & $\checkmark$  \\
    \secondtestsuite{}5 &  &  & $\checkmark$  & $\checkmark$  \\
    \secondtestsuite{}6 &  &  & $\checkmark$  & $\checkmark$  \\
    \secondtestsuite{}7 & $\checkmark$ &  & $\checkmark$  & $\checkmark$  \\
    \secondtestsuite{}8 & $\checkmark$ &  & $\checkmark$  & $\checkmark$ \\
    \secondtestsuite{}9 & $\checkmark$ & $\checkmark$  & $\checkmark$  & $\checkmark$ \\ \bottomrule
\end{tabular}%
}
\end{table}

%% file: 6-experiment.tex
\section{Experimental Study\label{sec:experiments}}
In this section, we provide an experimental study using the two test suites proposed in the previous section. 
First, we explain the experimental setup, including the selected EMO algorithms and the performance metric; then, we present experimental results followed by a discussion of the observations. Constrained by space, readers are referred to supplementary materials for implementation details.

\subsection{Experimental Settings}
A plethora of algorithms have been proposed in the EMO literature for solving MOPs (involving two/three objectives) or MaOPs (involving more than three objectives). 
%
%
\begin{wraptable}{r}{3.8cm}
\caption{Population size settings.\label{tab:pop_size_setting}}
\centering
\resizebox{.18\textwidth}{!}{%
    \begin{tabular}{@{\hspace{2mm}}ccc@{\hspace{2mm}}}
    \toprule
    $M$ & ($H_1$, $H_2$) & $N$ \\ \midrule
    2 & (99, 0) & 100 \\
    3 & (13, 0) & 105 \\
    4 & (7, 0) & 120 \\
    5 & (5, 0) & 126 \\
    6 & (4, 1) & 132 \\
    8 & (3, 2) & 156 \\ \bottomrule
    \end{tabular}%
}
\end{wraptable} 
Accordingly, we select six representative algorithms from each category as introduced in Section~\ref{sec:2.1}, including: NSGA-II~\cite{nsga2} (dominance-based), IBEA~\cite{zitzler2004indicator} (indicator-based), MOEA/D~\cite{moead} (decomposition-based), NSGA-III\cite{nsga3} (dominance-based), HypE~\cite{HypE} (indicator-based), and RVEA\footnote{We use the version with the reference vector regeneration strategy, which is also known as the RVEA$^\ast$.}~\cite{rvea} (decomposition-based), where the first three were classic ones for solving MOPs and the rest were tailored for solving MaOPs. 
%
%

All experiments are conducted on PlatEMO~\cite{platemo} -- an EMO algorithm library implemented in MATLAB. The population size is set in correspondence with the number objectives, as shown in Table~\ref{tab:pop_size_setting}. We perform 31 independent runs for each algorithm on each test instance using 10,000 fitness evaluations, and the statistical results are compared using Wilcoxon rank sum test.

\subsection{Performance Indicator}

In this paper, we adopt hypervolume (HV) to quantitatively compare the performance among the considered algorithms. Let us denote $\bm{y}^{\mbox{\tiny ref}} = (y_1, \ldots, y_m)$ as a reference point dominated by all Pareto-optimal solutions in the objective space, and $\bm{Y}$ as the Pareto-front approximated by an algorithm.
The HV value of $\bm{Y}$ (with respect to $\bm{y}^{\mbox{\tiny ref}}$) is the volume of the region dominating $\bm{y}^{\mbox{\tiny ref}}$ and dominated by $\bm{Y}$.

Specifically, we follow two rules to set the reference point for calculating HV: (i) for problems derived from search spaces that are exhaustively evaluated (i.e., $\bm{\Omega} = \{\mbox{NB101, NB201, NATS}\}$), we set $\bm{y}^{\mbox{\tiny ref}}$ to the nadir point since the actual Pareto front is available; (ii) for problems derived on the basis of surrogate models (i.e., $\bm{\Omega} = \{\mbox{DARTS, MNV3, ResNet50, Transformer}\}$), we set $\bm{y}^{\mbox{\tiny ref}}$ to the worst point among the samples collected for training the surrogate models. 
Readers are referred to Table A.I in the supplementary materials for the specific reference point for each test instance.

\subsection{Results}
In the following, we present results achieved by each of the considered algorithm on the two test suites. 

\subsubsection{Results on \firsttestsuite{}} 
\begin{figure*}[ht]
    \centering
    \begin{subfigure}[b]{\textwidth}
        \centering
        \includegraphics[trim={0 0 0 0}, clip, width=.16\textwidth]{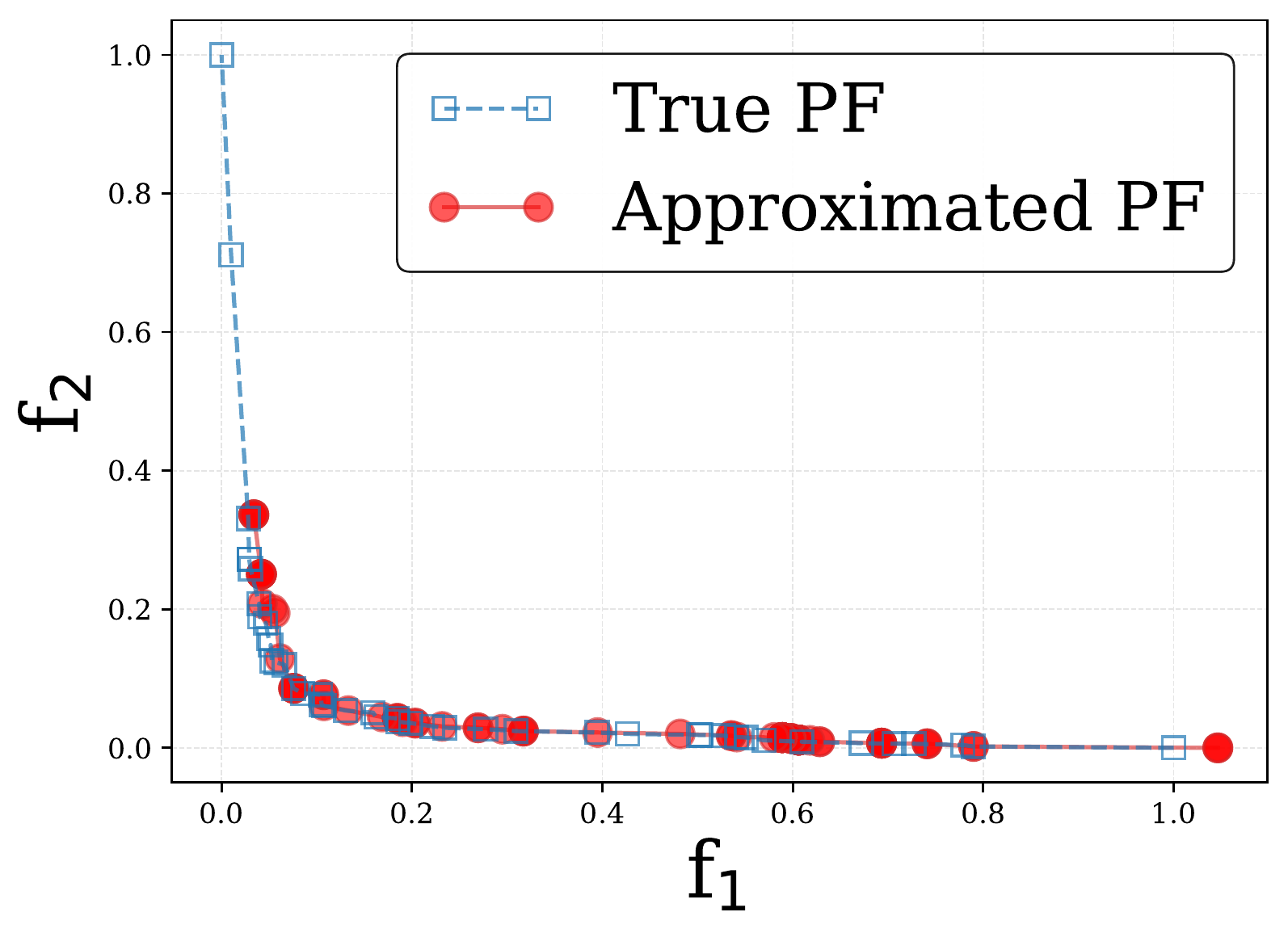}\hfill
        \includegraphics[trim={0 0 0 0}, clip, width=.16\textwidth]{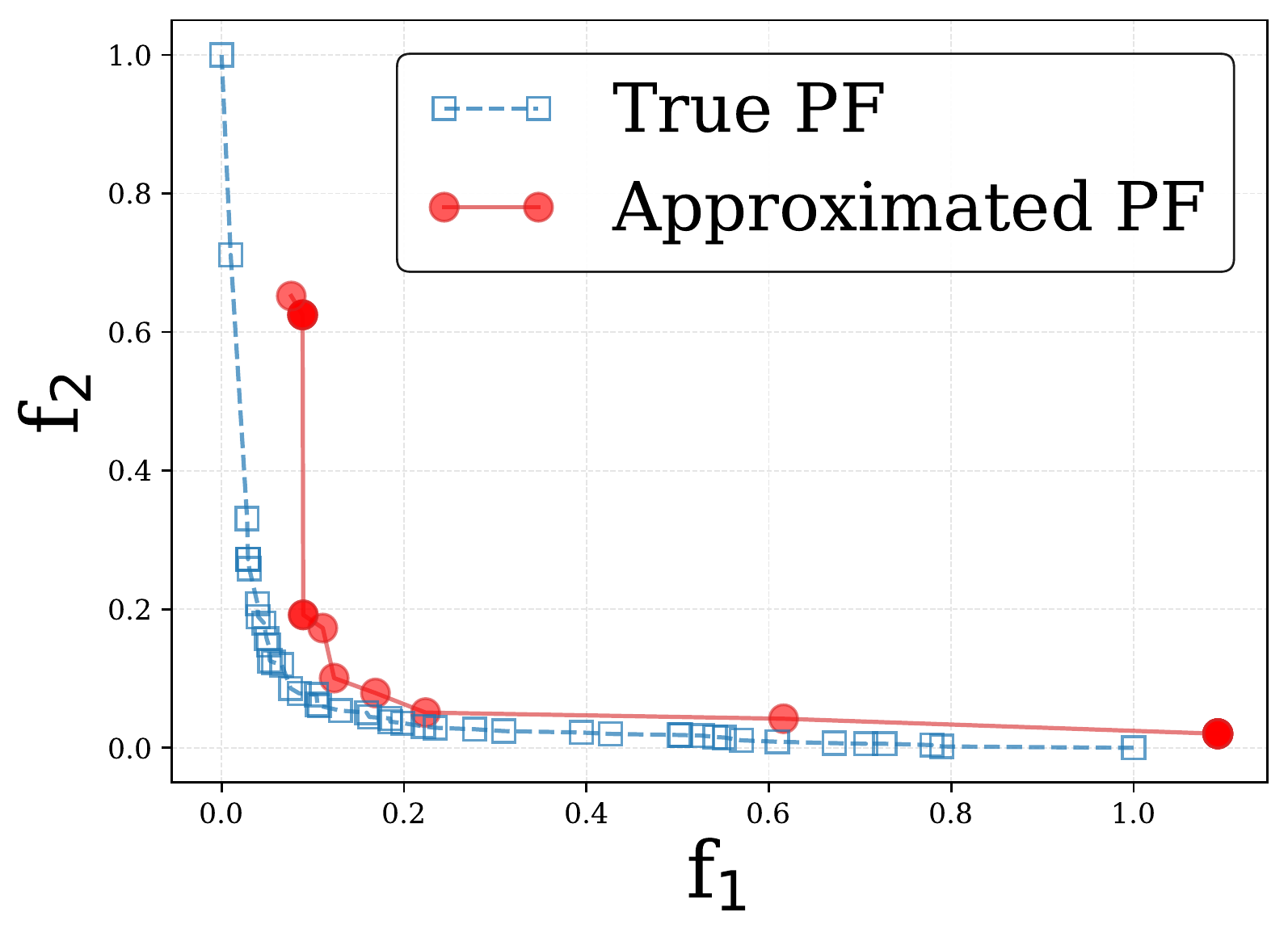}\hfill
        \includegraphics[trim={0 0 0 0}, clip, width=.16\textwidth]{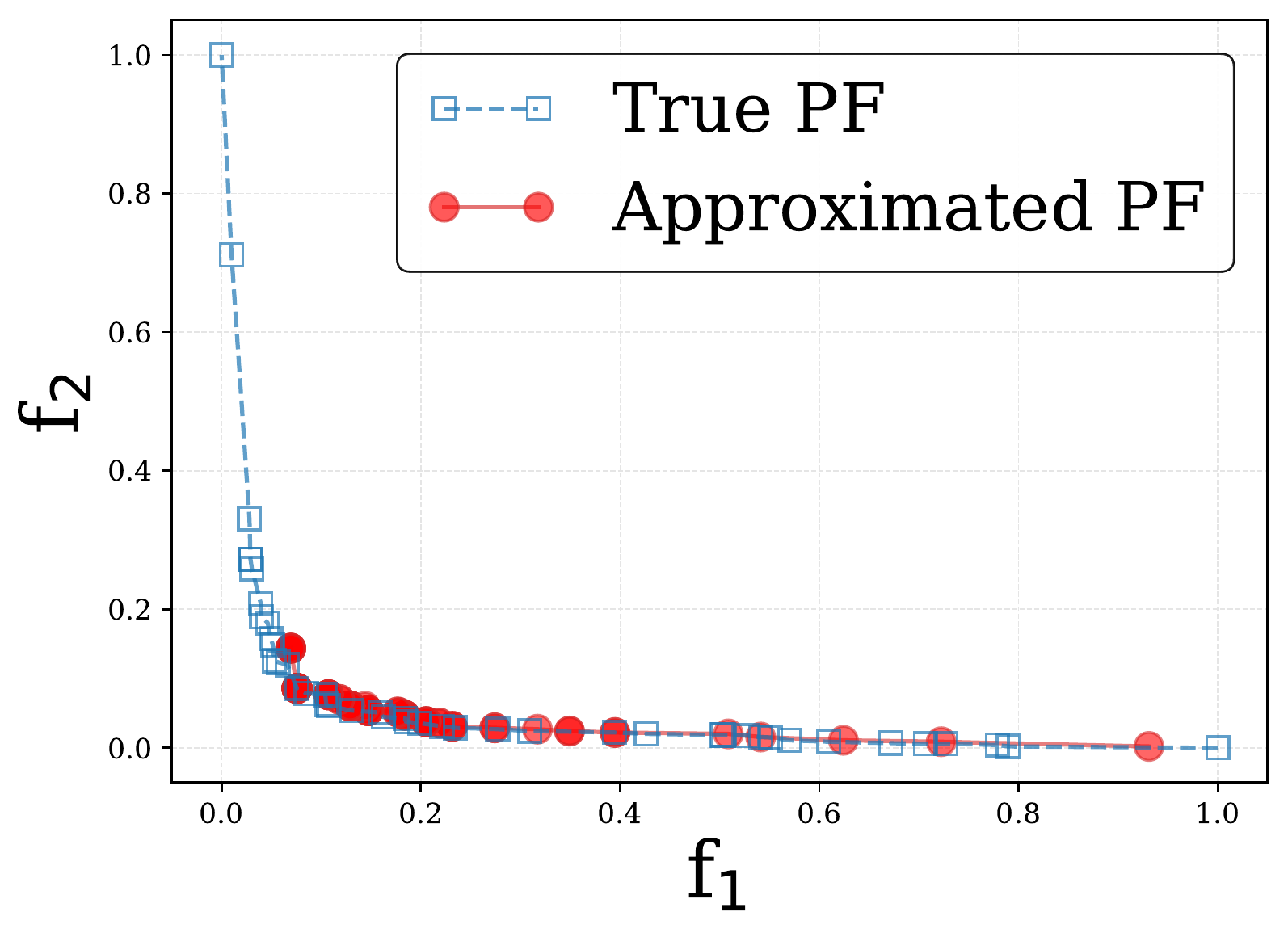}\hfill
        \includegraphics[trim={0 0 0 0}, clip, width=.16\textwidth]{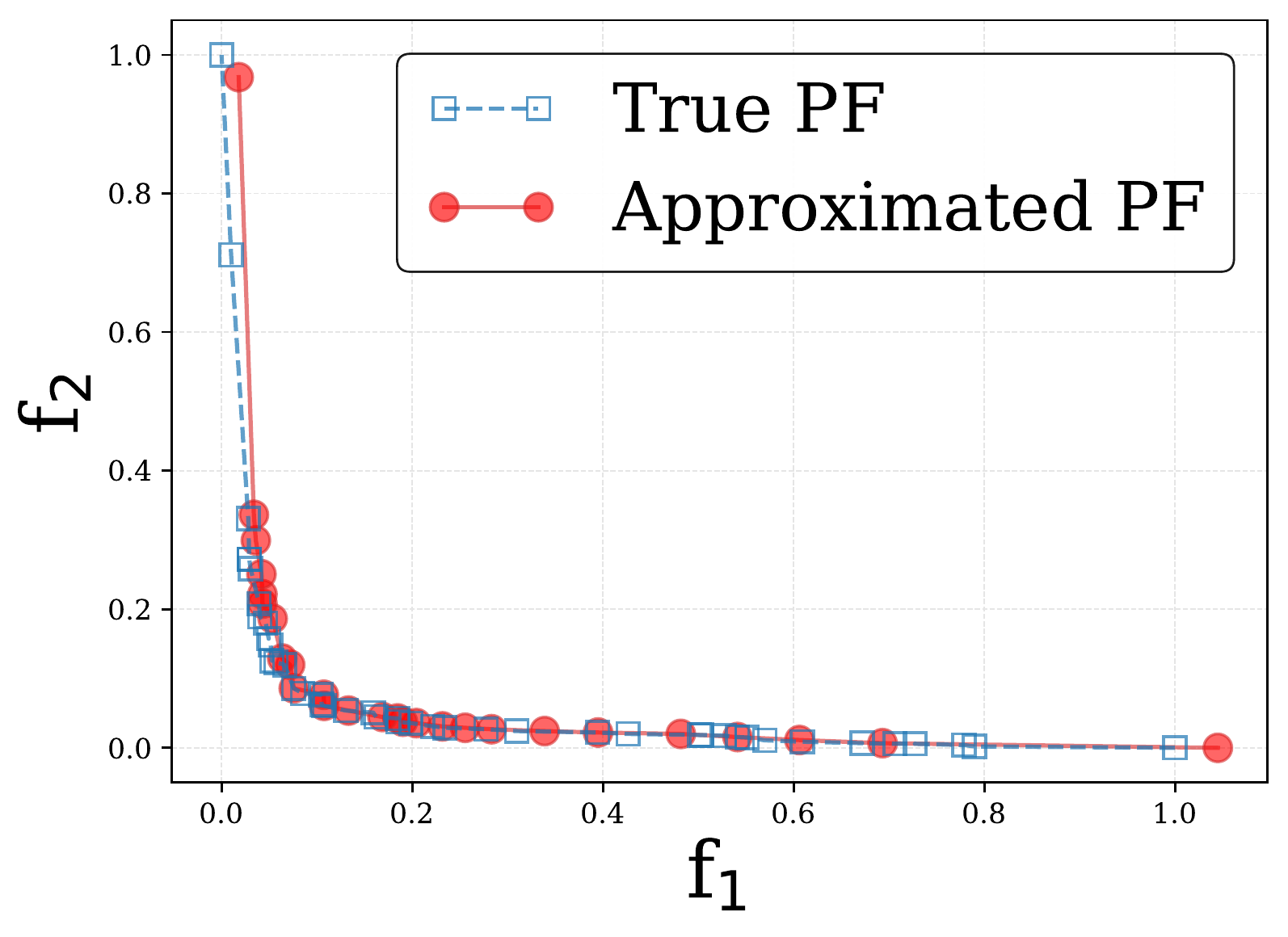}\hfill
        \includegraphics[trim={0 0 0 0}, clip, width=.16\textwidth]{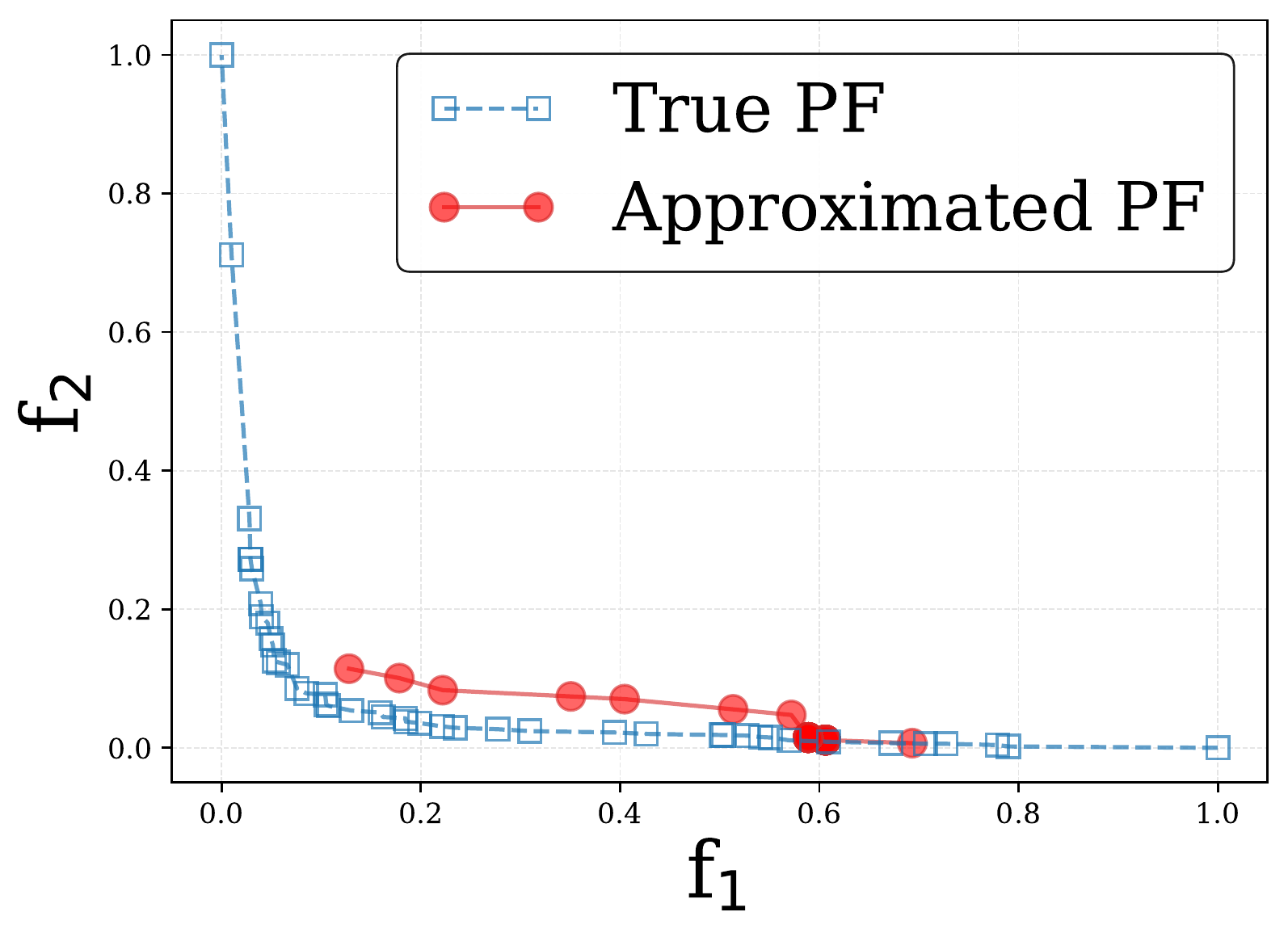}\hfill
        \includegraphics[trim={0 0 0 0}, clip, width=.16\textwidth]{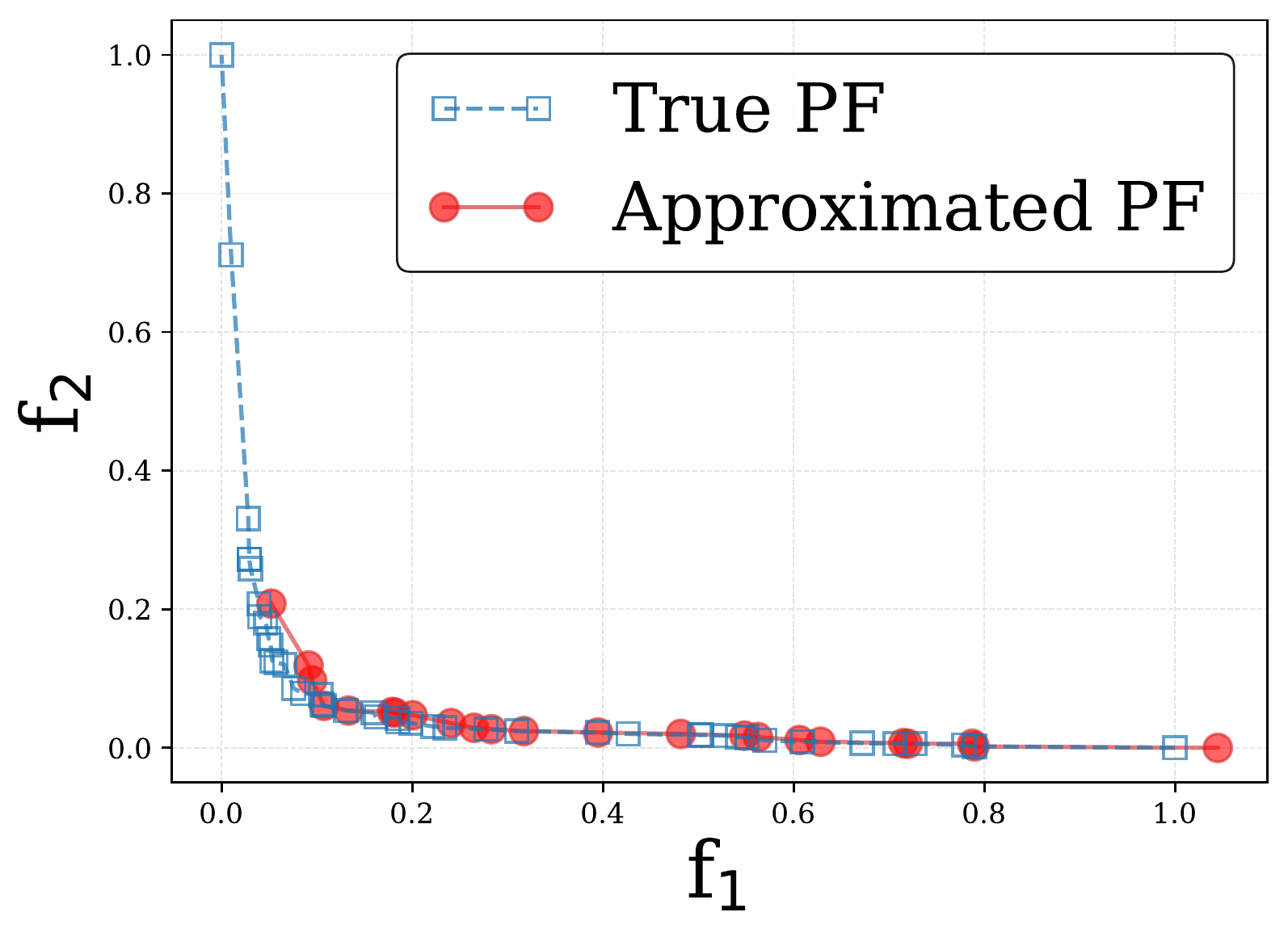}\hfill
        \vspace{-5pt}
        \caption{\firsttestsuite{}1 \label{fig:c10mop1_pf}}
    \end{subfigure} \\ \vspace{5pt}
    \begin{subfigure}[b]{\textwidth}
        \centering
        \includegraphics[trim={0 0 0 0}, clip, width=.16\textwidth]{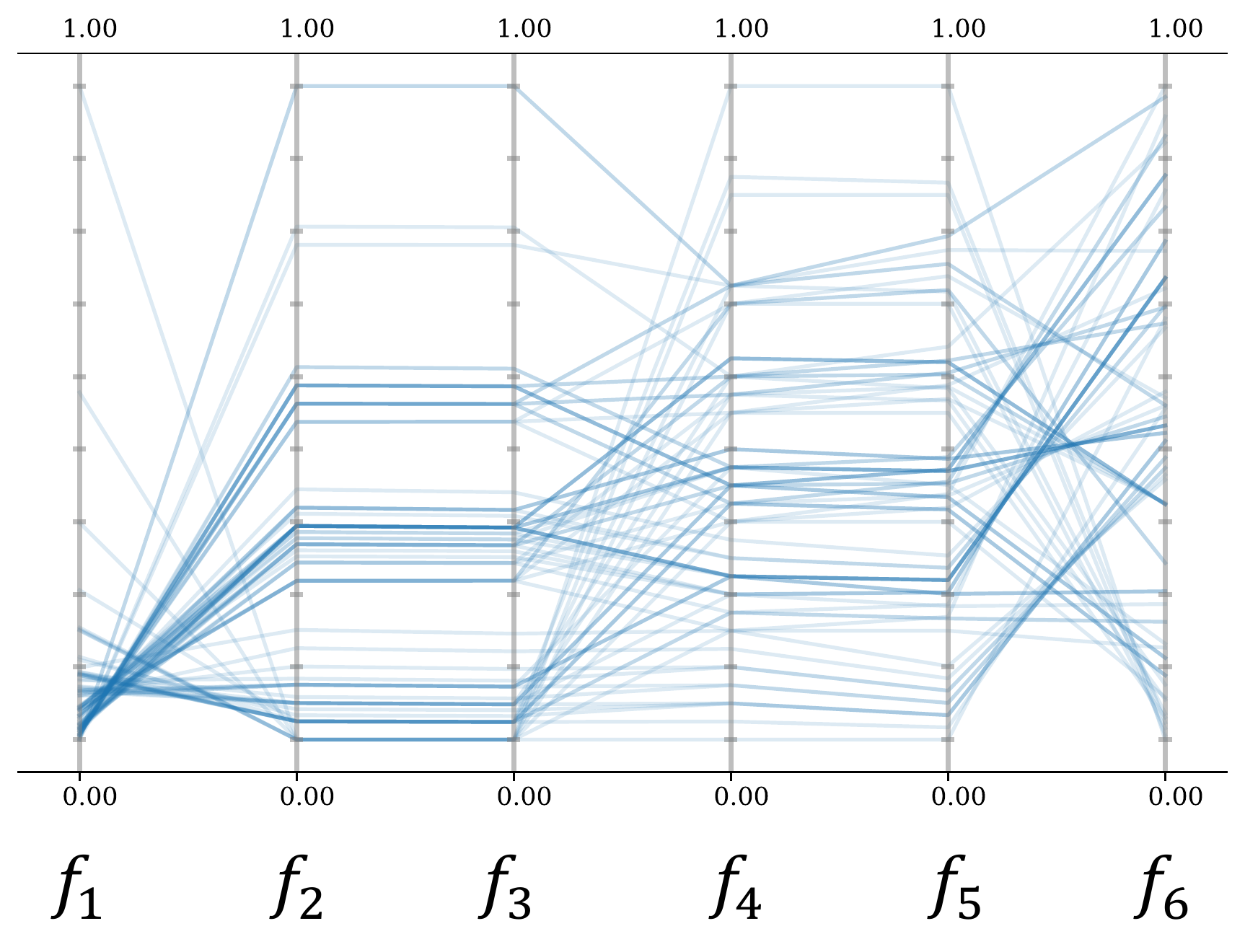}\hfill
        \includegraphics[trim={0 0 0 0}, clip, width=.16\textwidth]{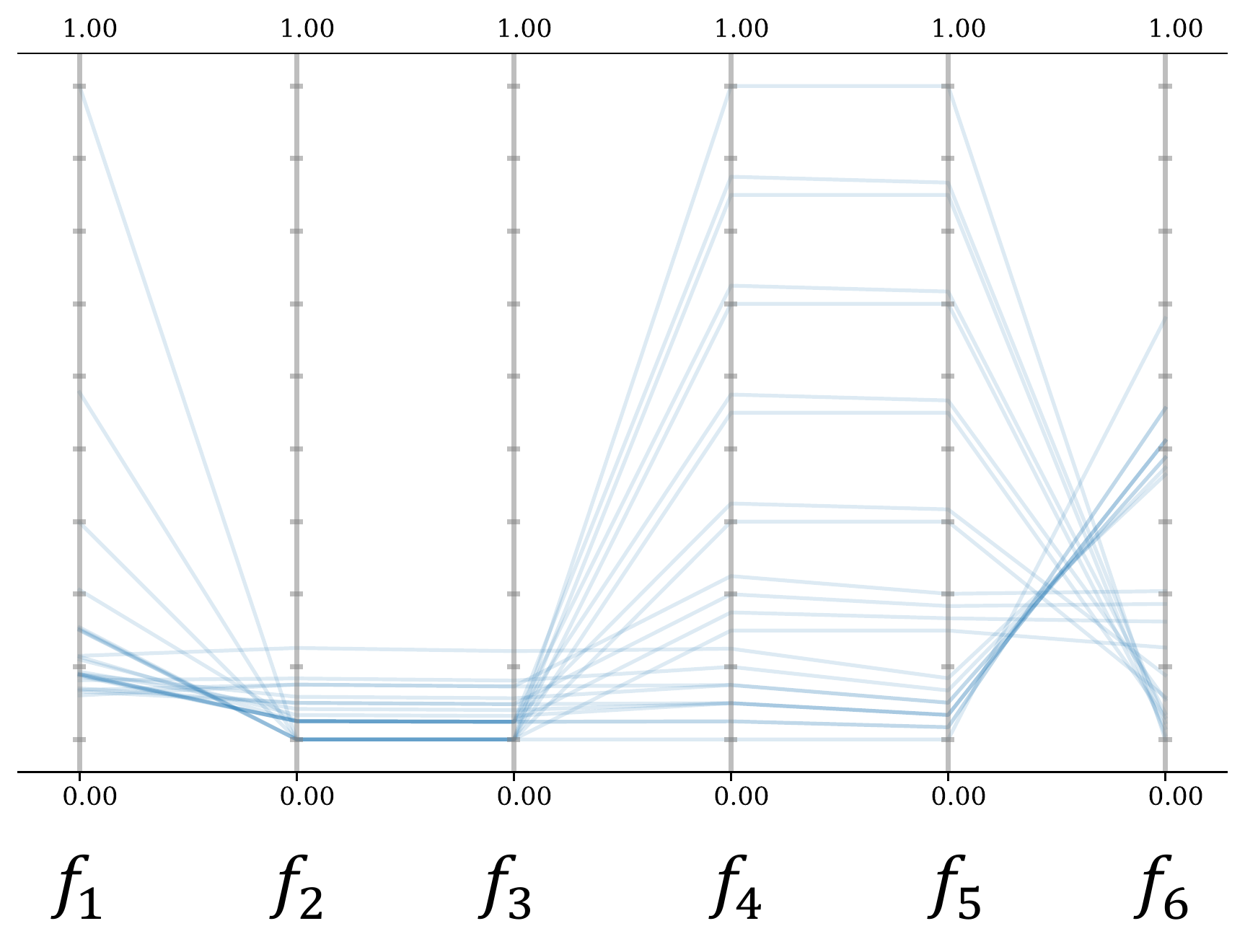}\hfill
        \includegraphics[trim={0 0 0 0}, clip, width=.16\textwidth]{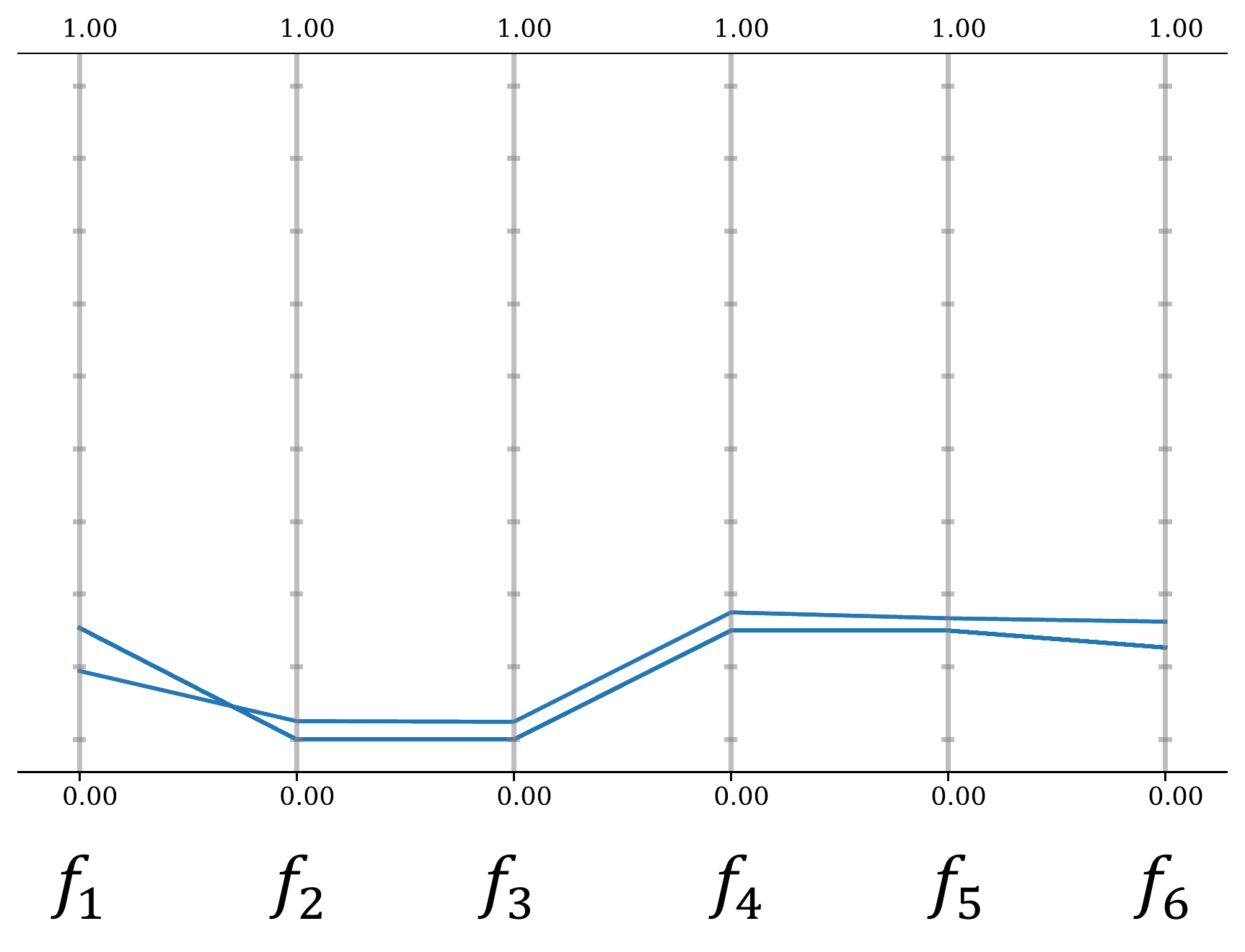}\hfill
        \includegraphics[trim={0 0 0 0}, clip, width=.16\textwidth]{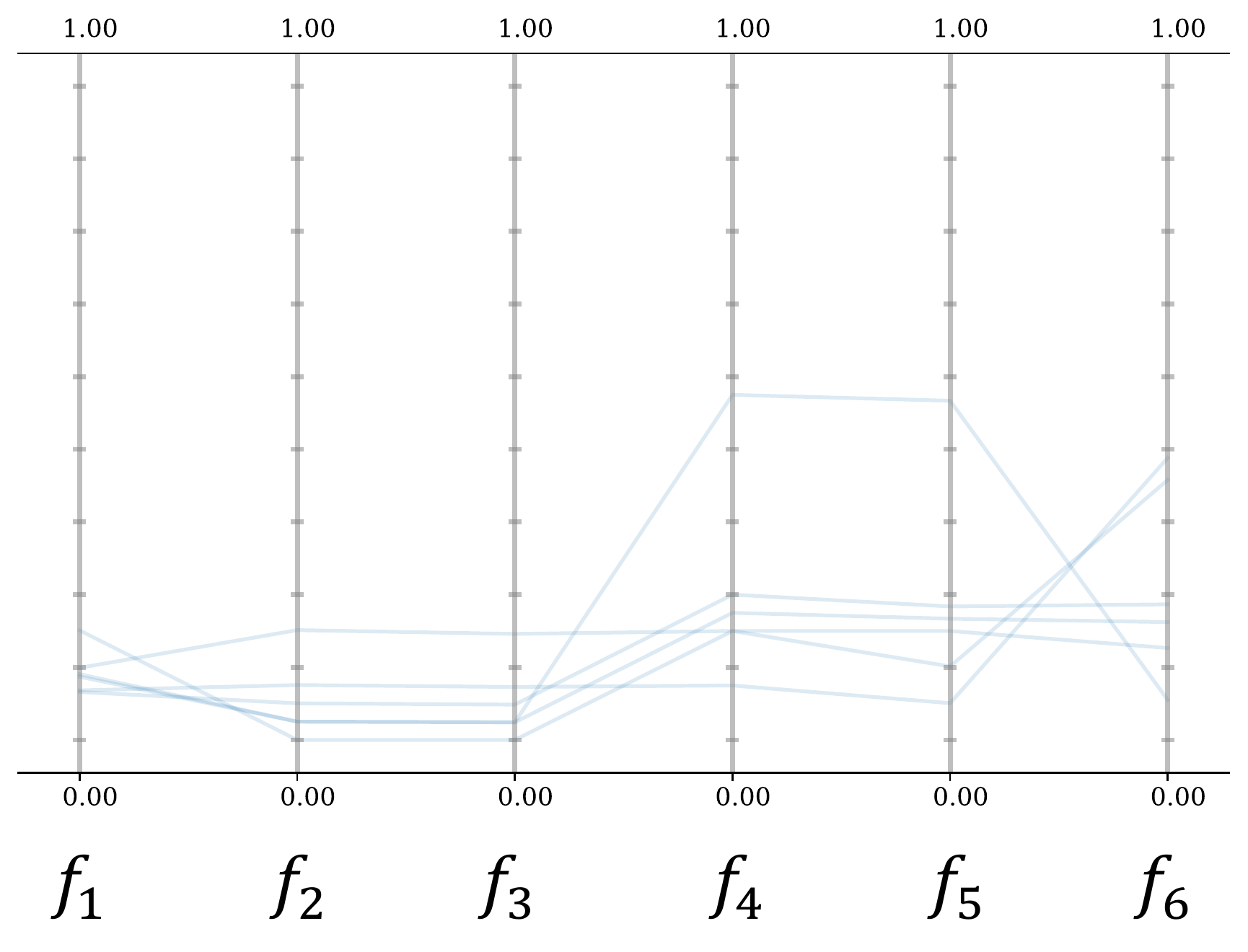}\hfill
        \includegraphics[trim={0 0 0 0}, clip, width=.16\textwidth]{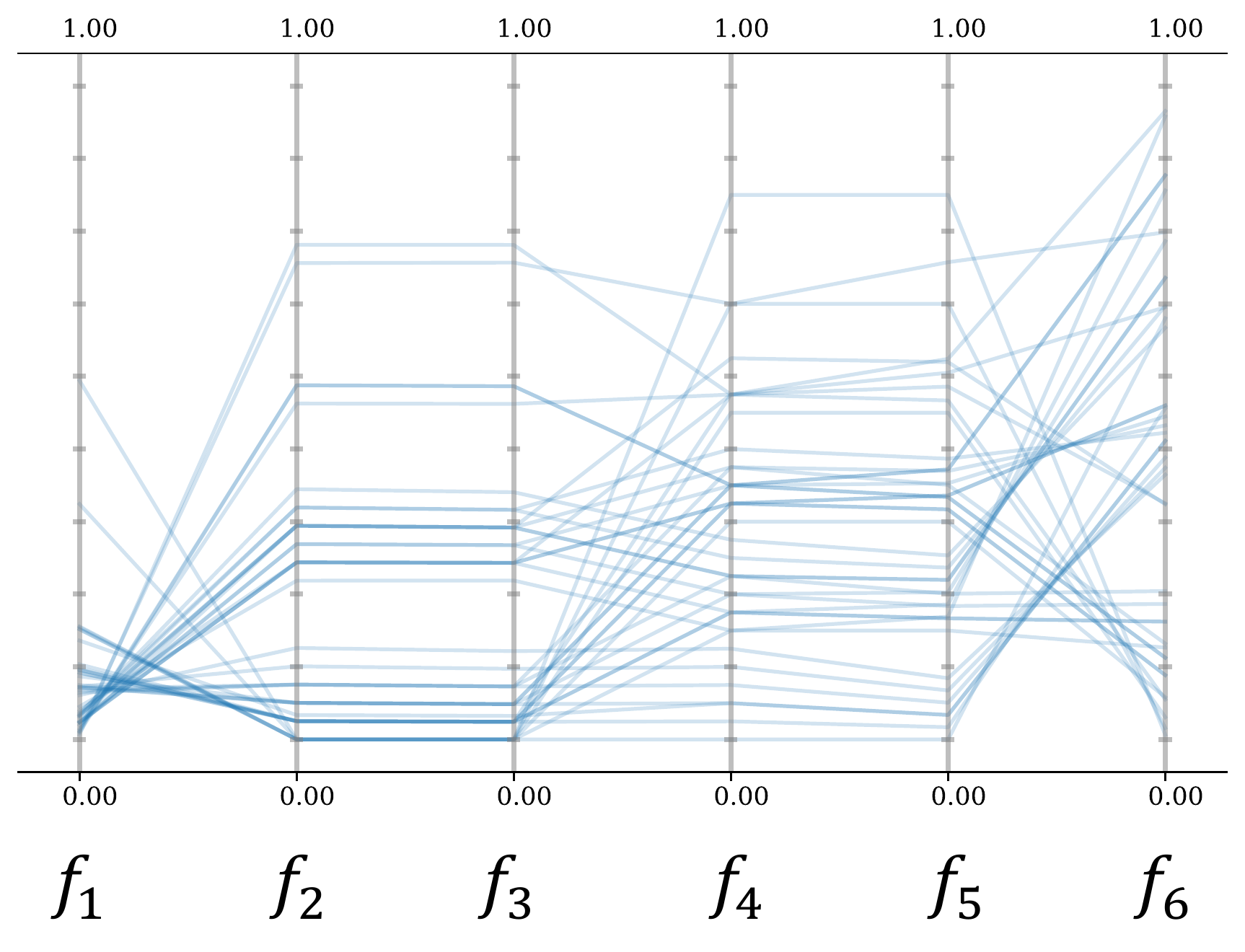}\hfill
        \includegraphics[trim={0 0 0 0}, clip, width=.16\textwidth]{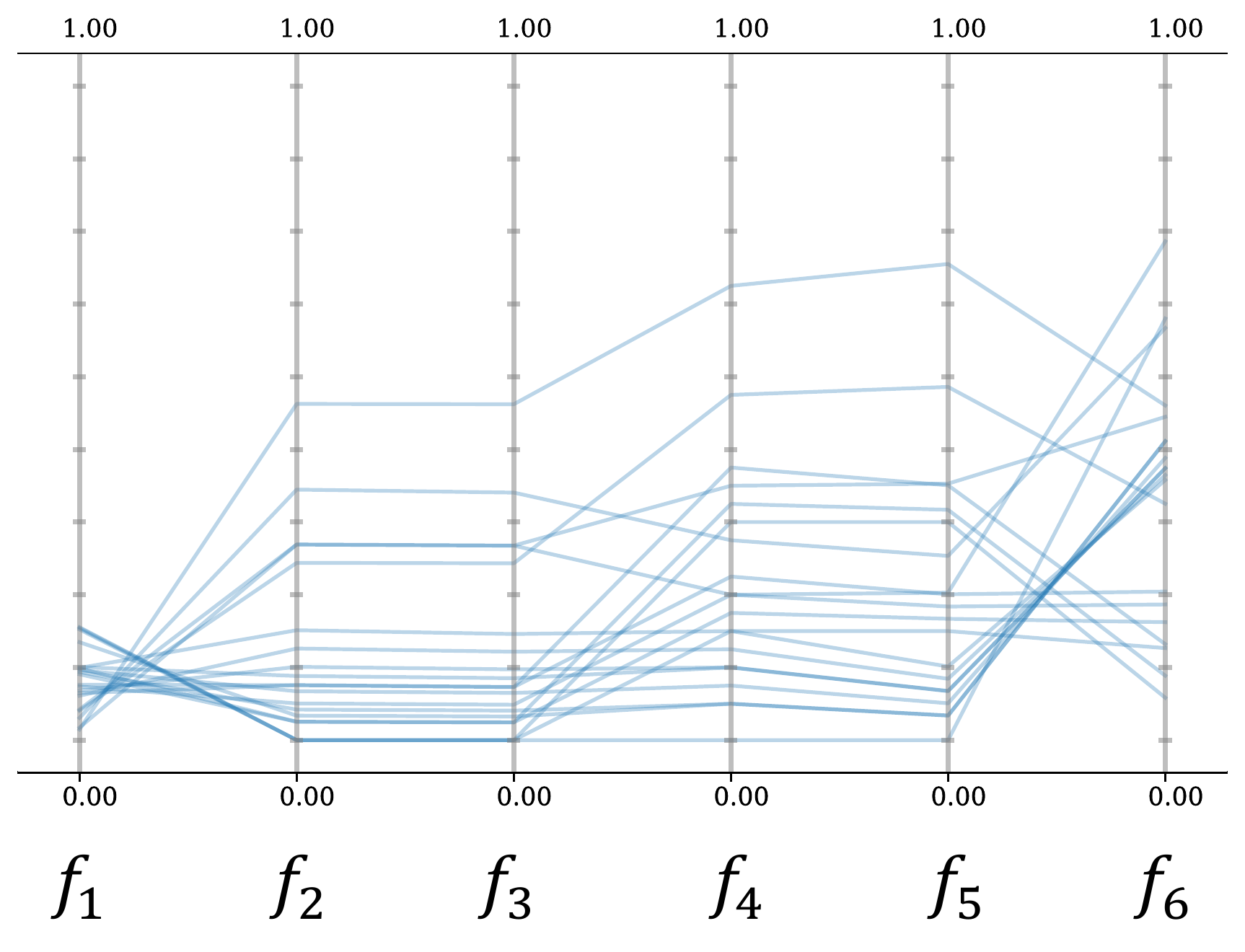}\hfill
        \vspace{-5pt}
        \caption{\firsttestsuite{}6 \label{fig:c10mop6_pf}}
    \end{subfigure} 
    \caption{Nondominated solutions obtained by each algorithm on (a) \firsttestsuite{}1 and (b) \firsttestsuite{}6. We select the run associated with the median HV value. For each row, the subfigures correspond to NSGA-II, IBEA, MOEA/D, NSGA-III, HypE, and RVEA, respectively. \label{fig:c10mop_visualization}}
\end{figure*}

Table~\ref{tab:c10mop_hv} summarizes the statistical values of the HV metric achieved by the six algorithms. 
In general, we can observe that none of the six algorithms can effectively solve all instances in the \firsttestsuite{} test suite. 
More specifically, on MOPs with two or three objectives (i.e., \firsttestsuite{}1, \firsttestsuite{}2, \firsttestsuite{}8, and \firsttestsuite{}9), we can observe that NSGA-II consistently outperforms other algorithms, except \firsttestsuite{}3 where IBEA performs slightly better, while on MaOPs involving more than three objectives (i.e., \firsttestsuite{}4 - \firsttestsuite{}7), we can observe that HypE yields generally the best performance
%
For further observations, we visualize the final nondominated solutions obtained by each algorithm in the median run on two typical test instances in Fig.~\ref{fig:c10mop1_pf} and Fig.~\ref{fig:c10mop6_pf} for \firsttestsuite{}1 and \firsttestsuite{}6 respectively. 
In the following, we provide some discussions based on the observations made from Fig.~\ref{fig:c10mop_visualization}.

\begin{table*}[ht]
\centering
\caption{Statistical results (median and standard deviation) of the HV values on \firsttestsuite{} test suite. The best results of each instance are in bold. \label{tab:c10mop_hv}}
\resizebox{.85\textwidth}{!}{%
\begin{threeparttable}
    \begin{tabular}{@{\hspace{2mm}}ccccccc@{\hspace{2mm}}}
    \toprule
    Problem & NSGA-II & IBEA & MOEA/D & NSGA-III & HypE & RVEA \\ \midrule
    \firsttestsuite{}1 & \textbf{0.9367 (0.0042)}$\bm^{\approx}$ & 0.8627 (0.0186)$^-$               & 0.9069 (0.0151)$^-$   & \textbf{0.9379 (0.0047)}$\bm{^\approx}$  & 0.8488 (0.0678)$^-$               & 0.9297 (0.0083)$^-$ \\
    \firsttestsuite{}2 & \textbf{0.9176 (0.0025)}$\bm{^+}$       & 0.8341 (0.0228)$^-$               & 0.8727 (0.0064)$^-$   & 0.7108 (0.2048)$^-$                      & 0.7879 (0.0848)$^-$               &  0.9091 (0.0071)$^-$\\
    \firsttestsuite{}3 & 0.8235 (0.0015)$^-$                     & \textbf{0.8323 (0.0009)}$\bm{^+}$ & 0.8000 (0.0047)$^-$   & 0.8083 (0.0044)$^-$                      & 0.8183 (0.0071)$^-$               &  0.8262 (0.0072)$^-$  \\
    \firsttestsuite{}4 & 0.7656 (0.0085)$^-$                     & \textbf{0.7941 (0.0061)}$\bm{^+}$ & 0.7277 (0.0125)$^-$   & 0.7633 (0.0091)$^-$                      & 0.7692 (0.0080)$^-$               &  0.7544 (0.0189)$^-$ \\
    \firsttestsuite{}5 & 0.7127 (0.0000)${^-}$                   & 0.7094 (0.0007)$^-$               & 0.6721 (0.0045)$^-$   & 0.7040 (0.0179)$^-$                      & \textbf{0.7569 (0.0017)}$\bm{^+}$ &  {0.7521 (0.0077)}${^-}$\\
    \firsttestsuite{}6 & 0.7404 (0.0001)${^-}$                   & 0.7302 (0.0005)$^-$               & 0.6807 (0.0560)$^-$   & 0.6387 (0.0149)$^-$                      & \textbf{0.7743 (0.0006)}$\bm{^+}$ &  {0.7549 (0.0145)}${^-}$\\
    \firsttestsuite{}7 & 0.5696 (0.0118)$^-$                     &  {0.5892 (0.0012)}${^-}$          & 0.5108 (0.0509)$^-$   & 0.5417 (0.0202)$^-$                      & \textbf{0.6322 (0.0068)}$\bm{^+}$ & {0.6122 (0.0146)}${^-}$ \\
    \firsttestsuite{}8 & \textbf{0.9769 (0.0043)}$\bm{^+}$       & 0.9751 (0.0080)$^-$               & 0.8870 (0.0250)$^-$   & 0.9741 (0.0064)$^-$                      & 0.9484 (0.0115)$^-$               & 0.9185 ((0.0168)$^-$ \\
    \firsttestsuite{}9 & \textbf{0.9630 (0.0097)}$\bm{^+}$       & 0.9609 (0.0136)$^-$               & 0.7633 (0.0404)$^-$   & 0.9579 (0.0066)$^-$                      & 0.9253 (0.0085)$^-$               & 0.8952 (0.0178)$^-$ \\ \bottomrule
    \end{tabular}%
\begin{tablenotes}
\setlength\labelsep{0pt}
\small
\item \hspace{1em}$^{+}$ indicates a method achieving significantly better performance.
\item \hspace{1em}$^{\approx}$ indicates a method achieving similar performance as the best-performing method.
\item \hspace{1em}$^{-}$ indicates a method achieving significantly worse performance.
\end{tablenotes}
\end{threeparttable}
}
\end{table*}

\firsttestsuite{}1 is a problem with a simple bi-objective convex PF; 
nevertheless, given its multi-modal and noisy fitness landscape, the problem is still non-trivial to solve. 
First, we can observe that all algorithms fail to converge to the low-$f_1$ regime (top-left corner), indicating the challenges for obtaining nondominated solutions with low $f_1$ (i.e., $f^e$; prediction error). 
Second, we can observe that indicator-based algorithms (i.e., IBEA and HypE) perform significantly worse than other peer methods. 
Third, both NSGA-II and NSGA-III achieve the best performance, where NSGA-II has slightly better coverage of the low-$f_2$ regime while NSGA-III has a slightly better converge of the low-$f_1$ regime. 

\firsttestsuite{}6 is relatively more complex with six (but partially correlated) objectives. 
The objective function of $f_1$ (i.e., $f^e$; prediction error) is also multi-modal and noisy, but the number of variables is small, i.e., six. 
First, we can observe that objective 2 ($f_1^c$; No. of parameters) and objective 3 ($f_2^c$: No. of FLOPs), objective 4 ($f_1^{h_2}$: latency on Eyeriss), and objective 5 ($f_2^{h_2}$: energy consumption on Eyeriss) are correlated, thus leading to a degenerated PF.
Second, we can observe that the decomposition-based algorithms (i.e., MOEA/D and RVEA) are significantly worse than those with purely Pareto-dominance-based (i.e., NSGA-II) or indicator-based (i.e., IBEA and HypE) ones. 
%

\subsubsection{Results on \secondtestsuite{}}
\begin{figure*}[ht]
    \centering
    \begin{subfigure}[b]{\textwidth}
        \centering
        \includegraphics[trim={0 0 0 0}, clip, width=.16\textwidth]{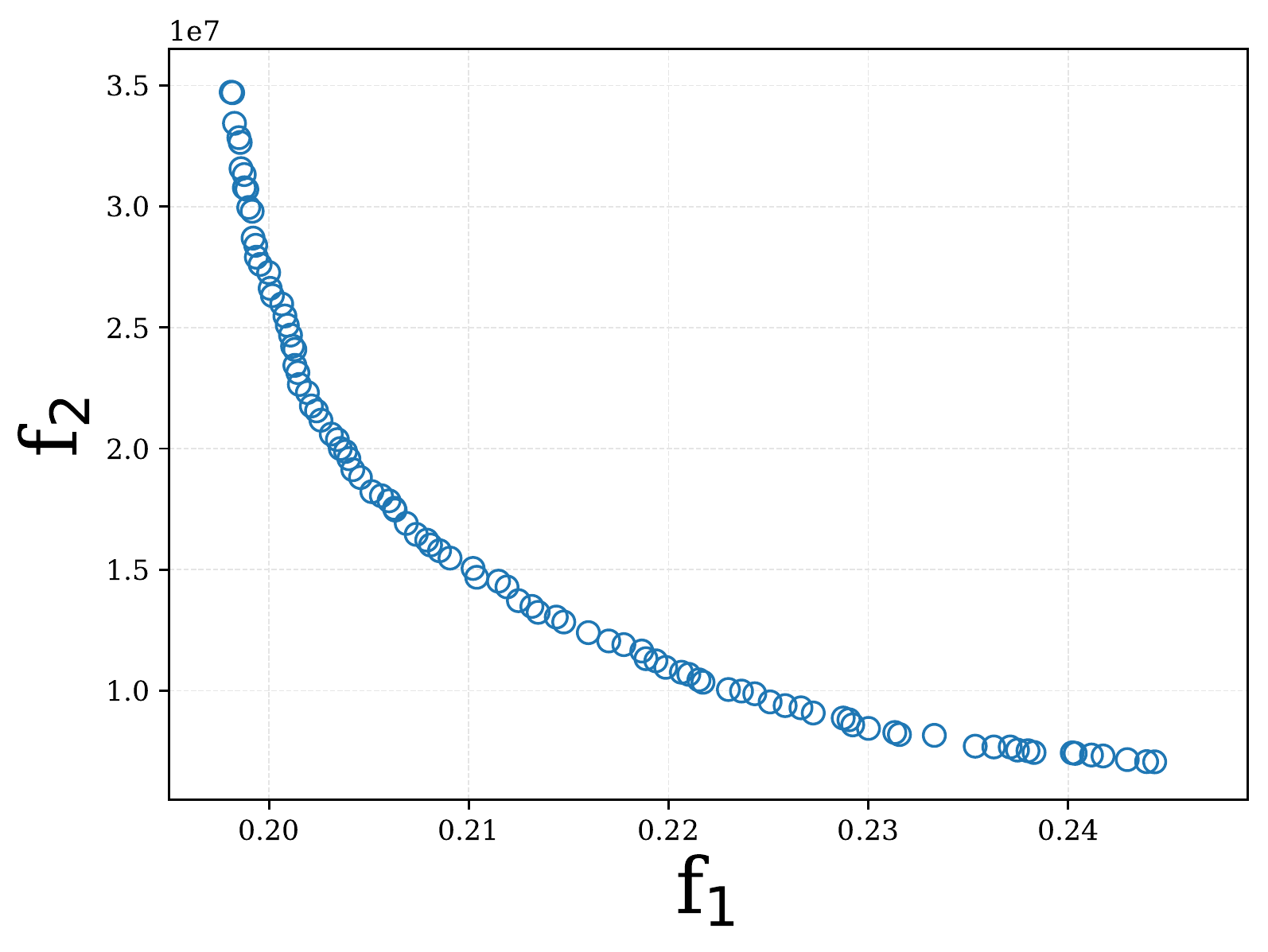}\hfill
        \includegraphics[trim={0 0 0 0}, clip, width=.16\textwidth]{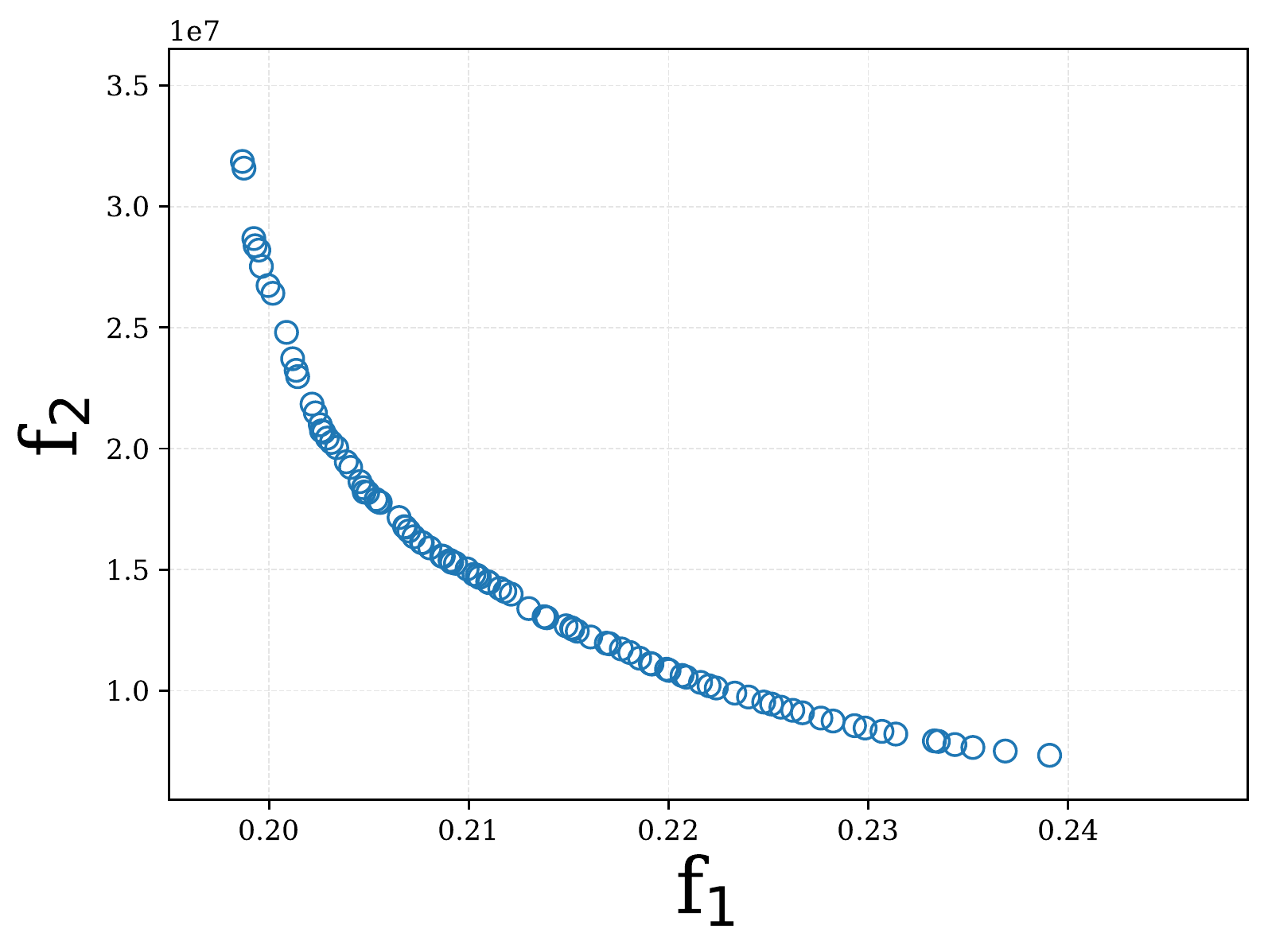}\hfill
        \includegraphics[trim={0 0 0 0}, clip, width=.16\textwidth]{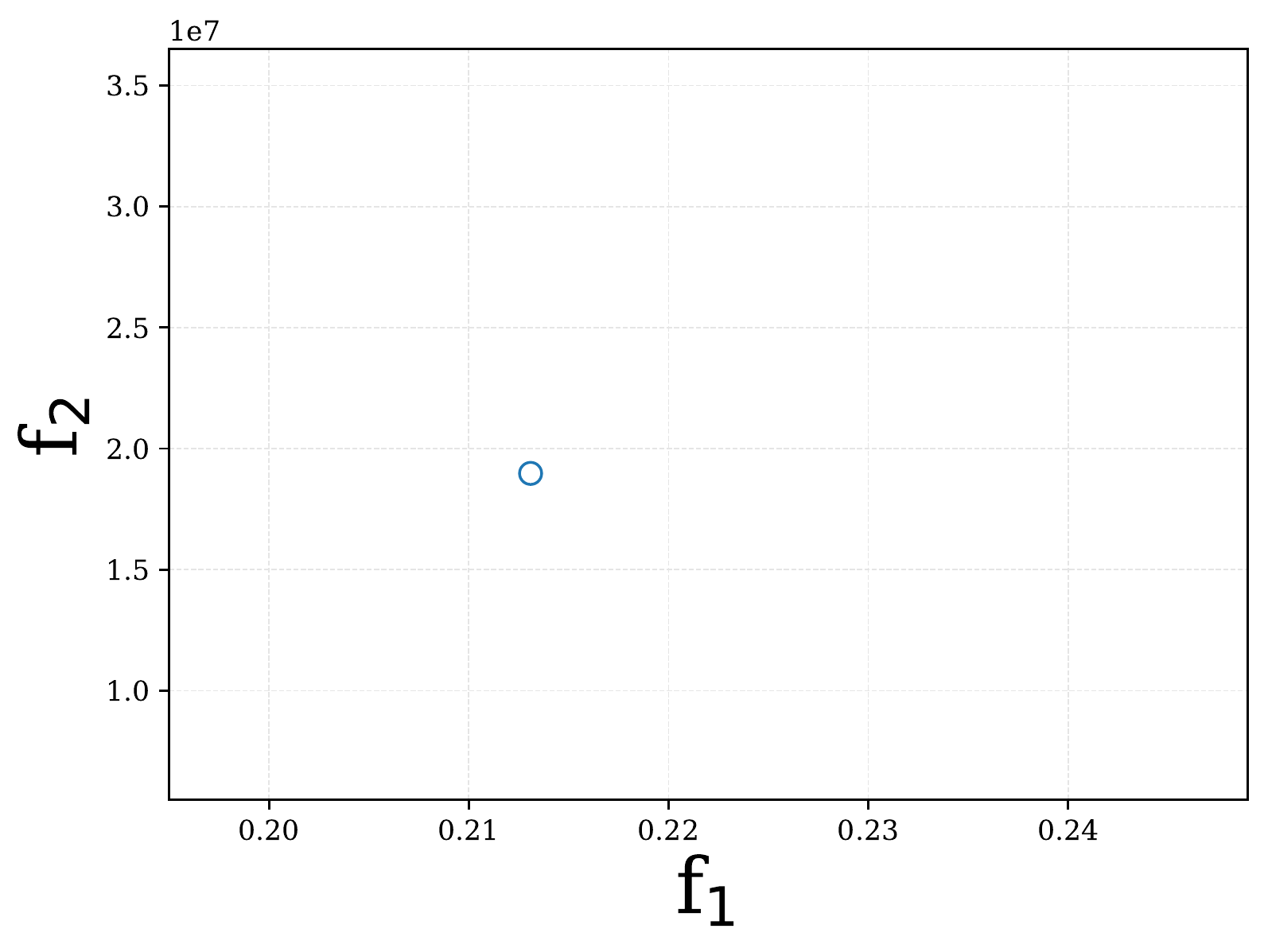}\hfill
        \includegraphics[trim={0 0 0 0}, clip, width=.16\textwidth]{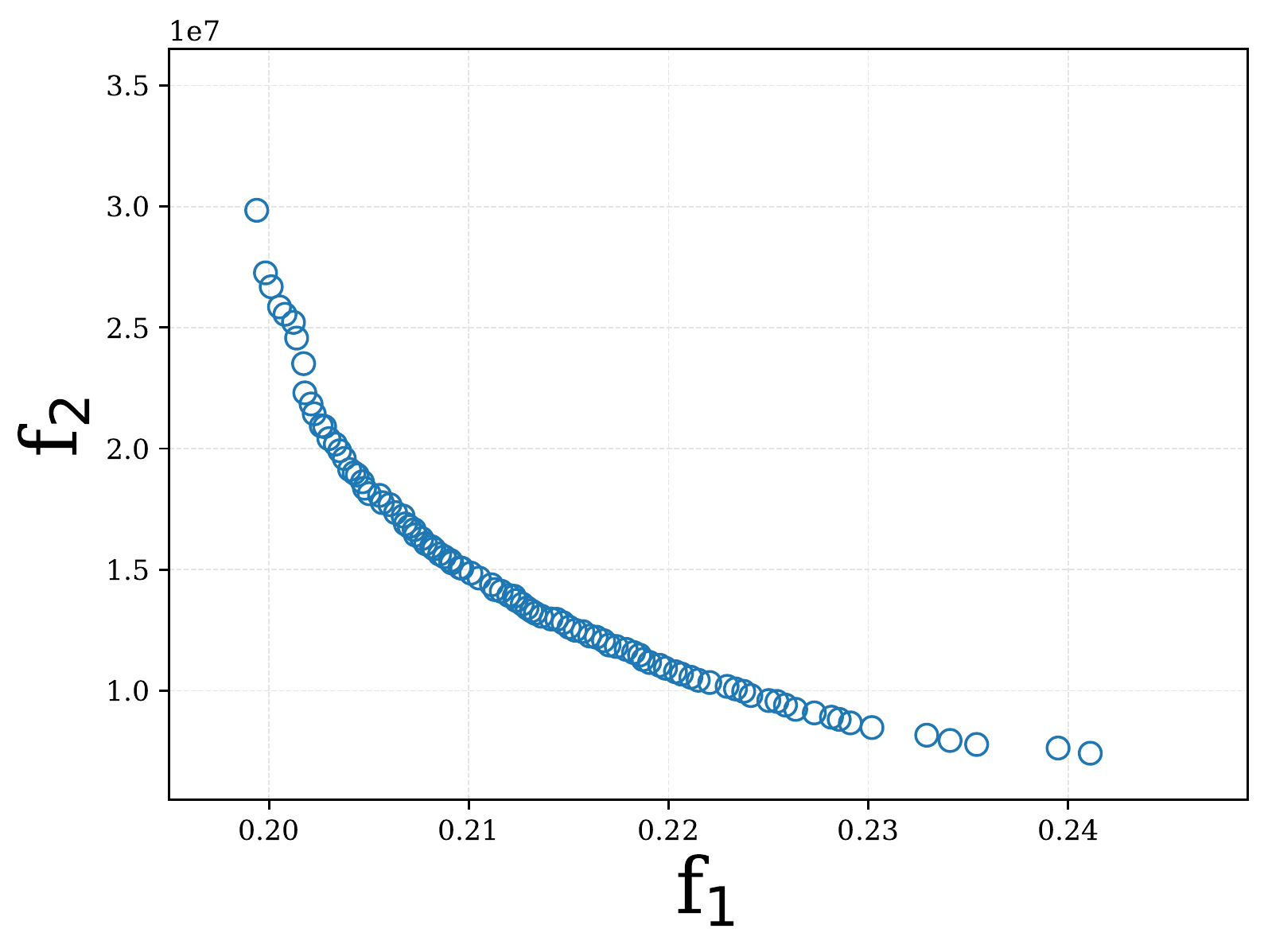}\hfill
        \includegraphics[trim={0 0 0 0}, clip, width=.16\textwidth]{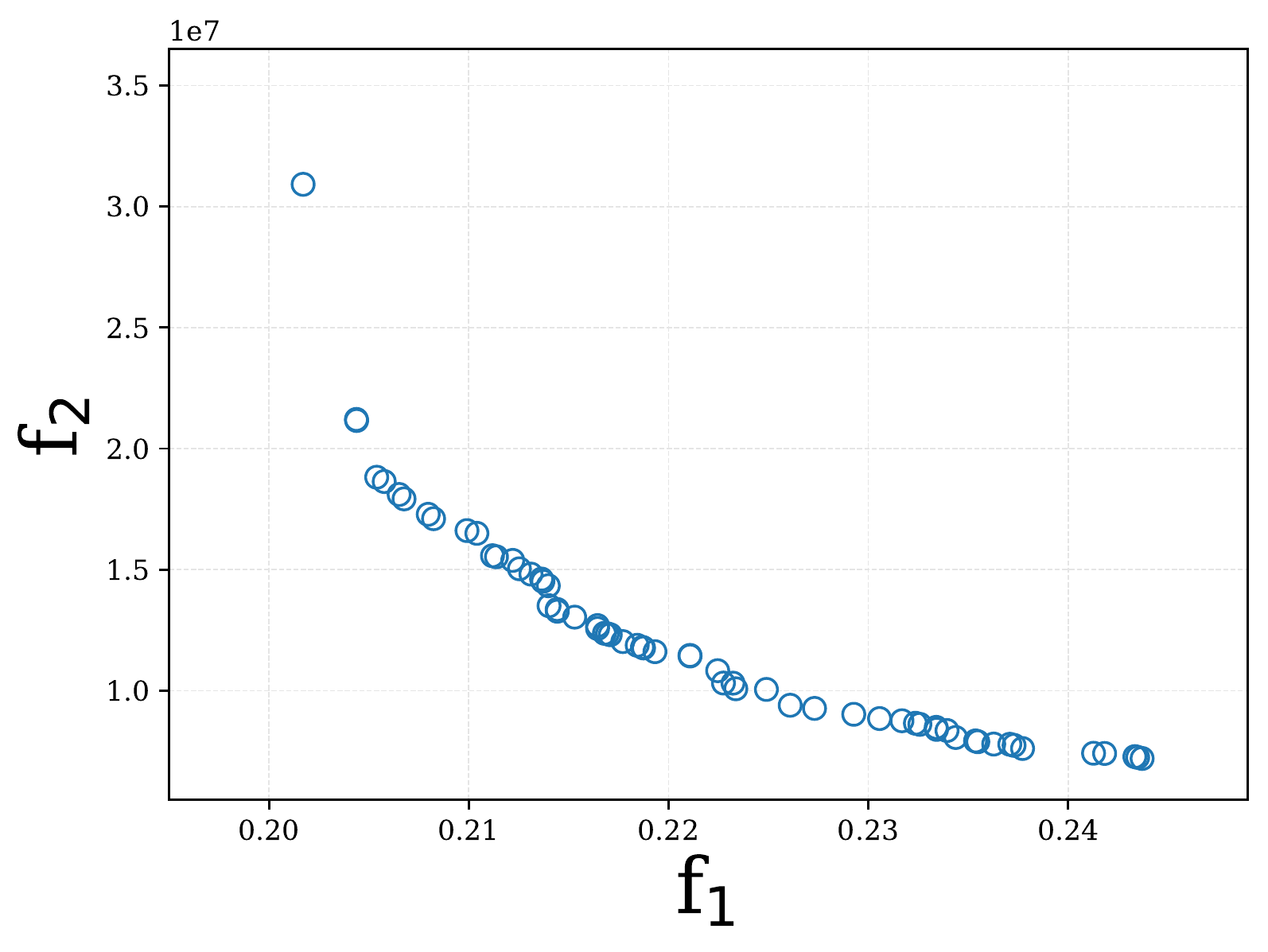}\hfill
        \includegraphics[trim={0 0 0 0}, clip, width=.16\textwidth]{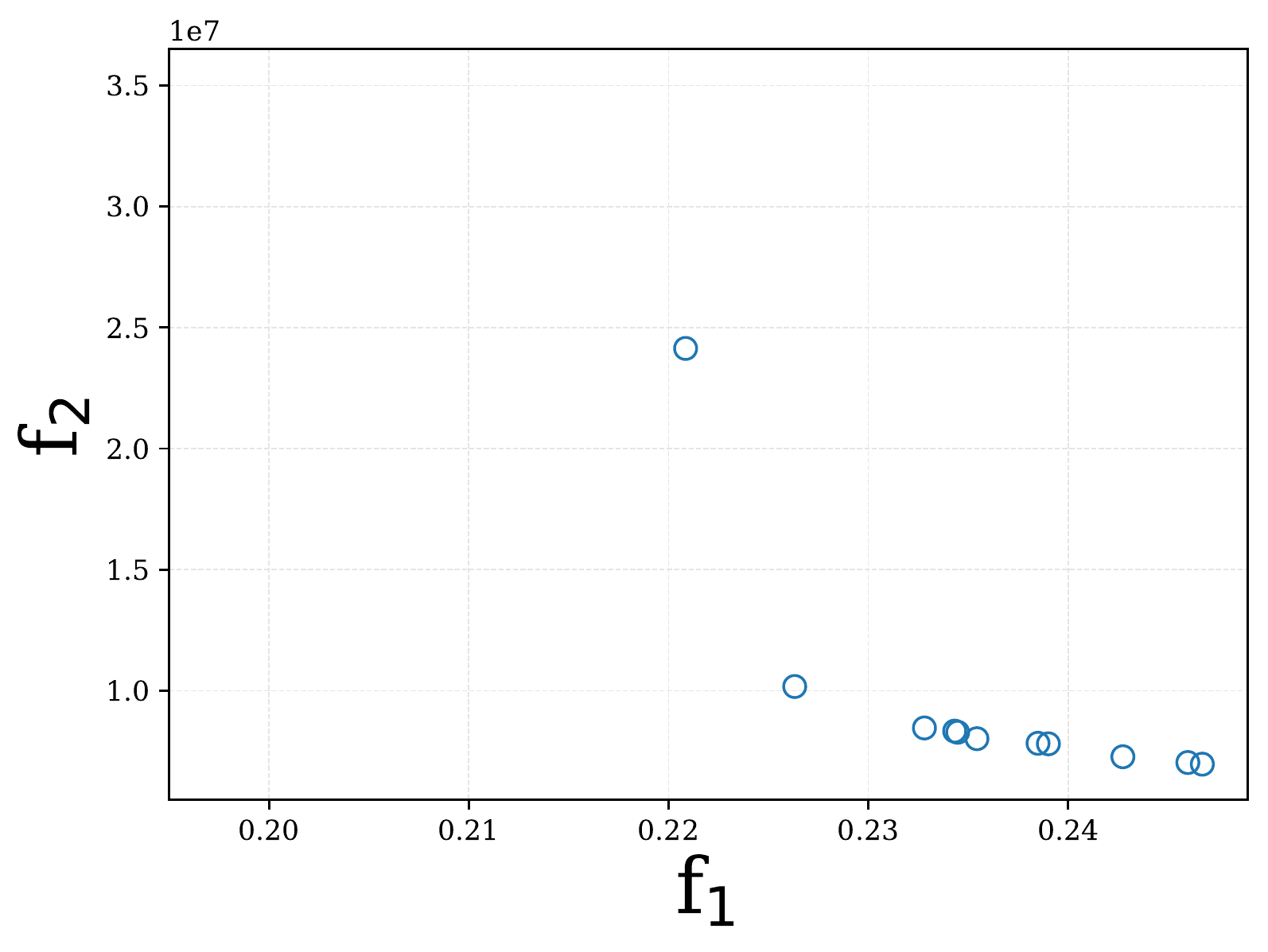}\hfill
        \vspace{-5pt}
        \caption{\secondtestsuite{}1 \label{fig:in1kmop1_pf}}
    \end{subfigure} \\ 
    \begin{subfigure}[b]{\textwidth}
        \centering
        \includegraphics[trim={0 0 0 0}, clip, width=.16\textwidth]{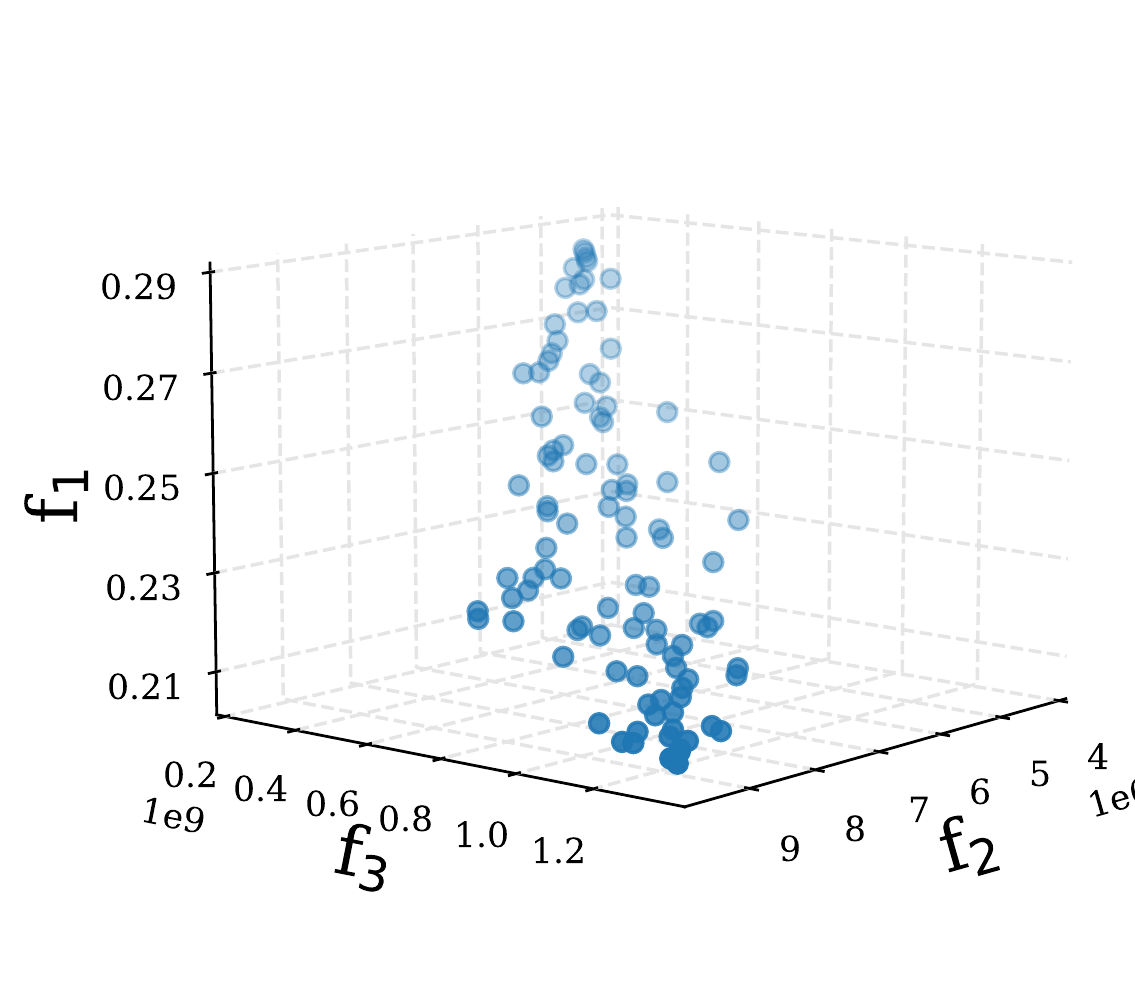}\hfill
        \includegraphics[trim={0 0 0 0}, clip, width=.16\textwidth]{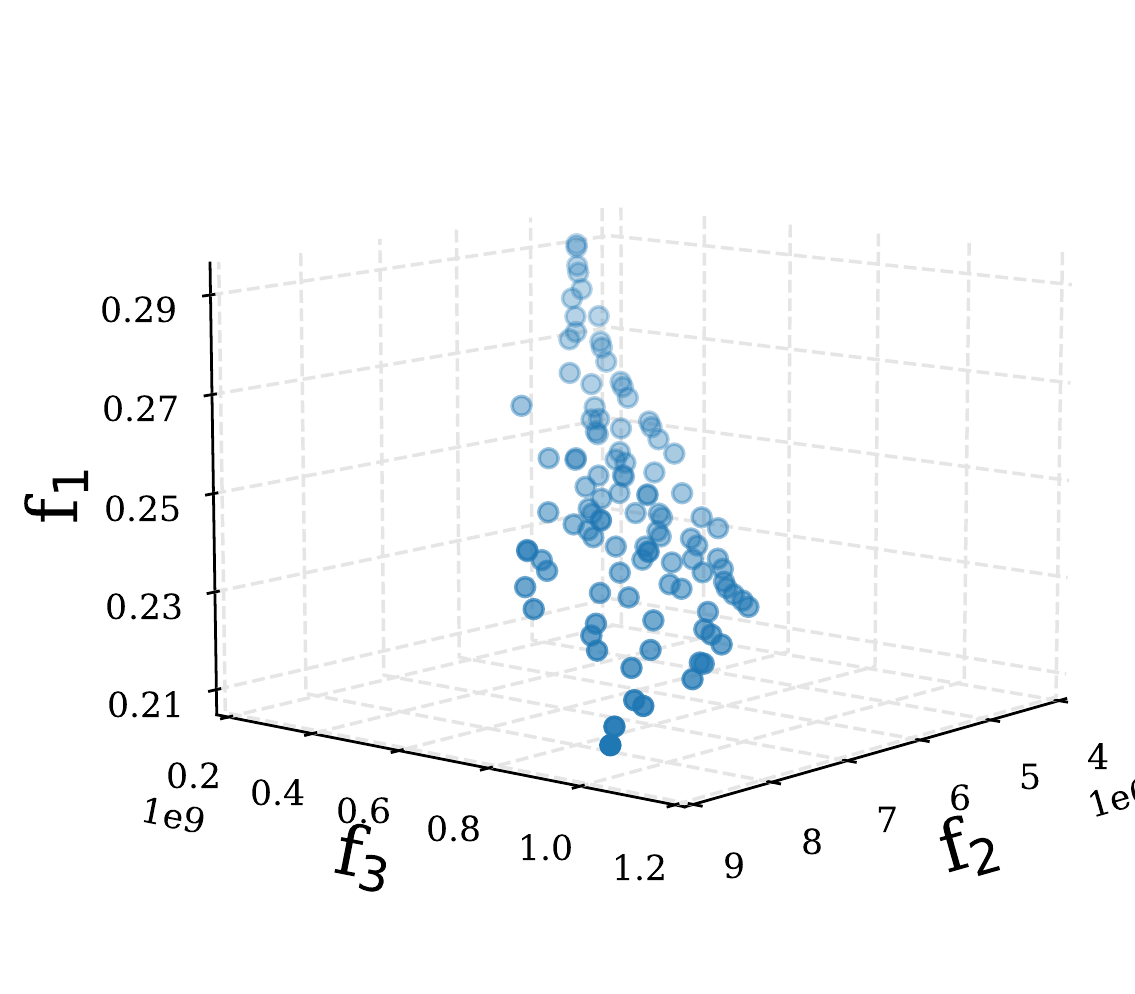}\hfill
        \includegraphics[trim={0 0 0 0}, clip, width=.16\textwidth]{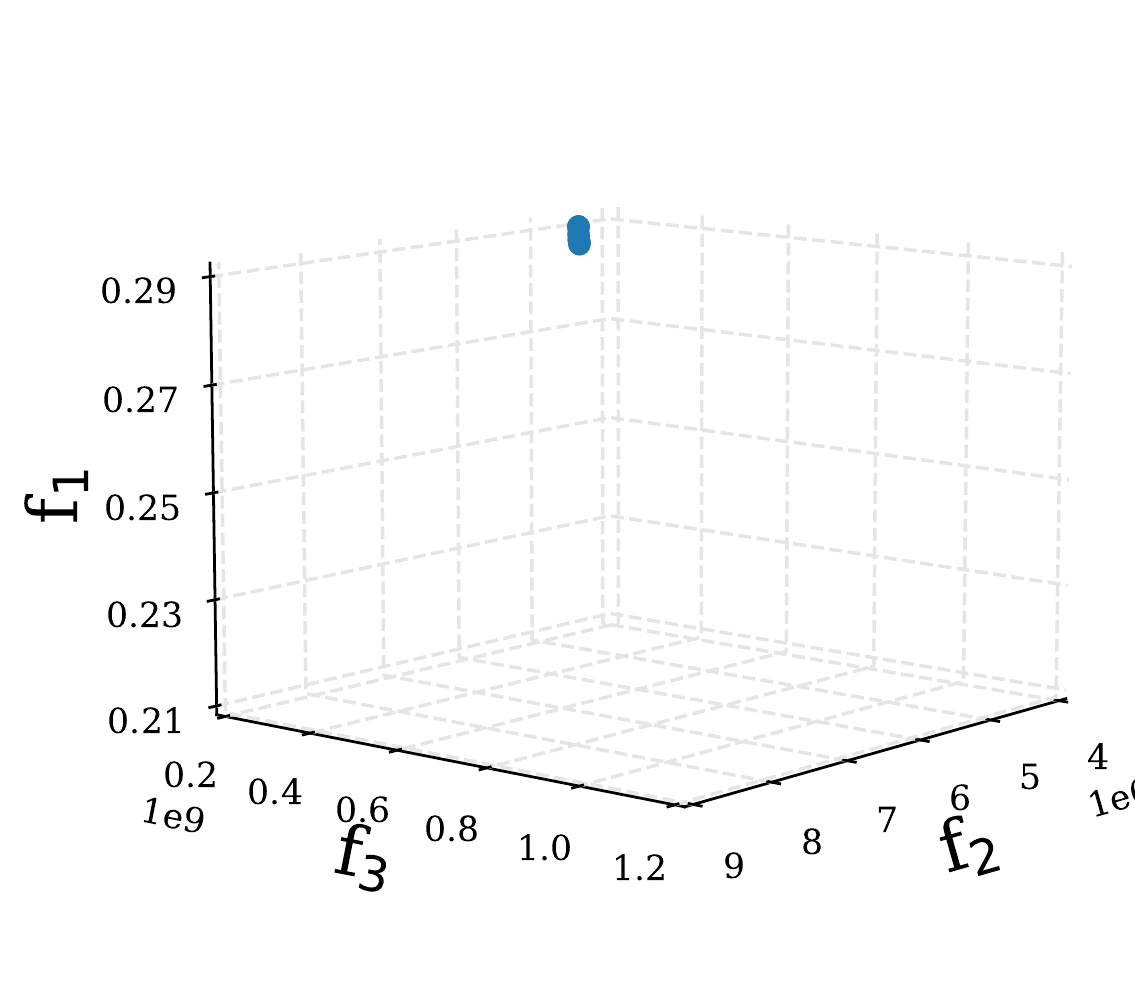}\hfill
        \includegraphics[trim={0 0 0 0}, clip, width=.16\textwidth]{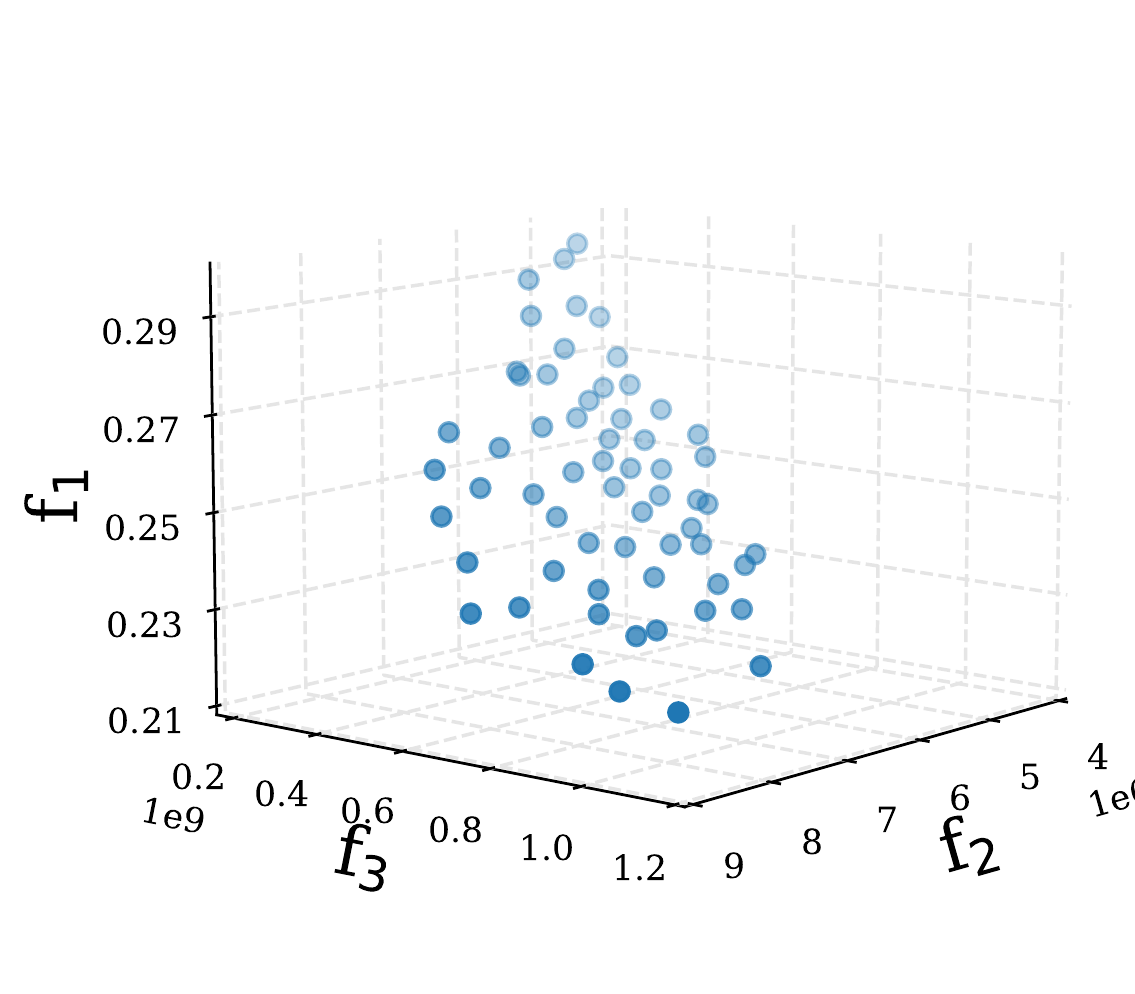}\hfill
        \includegraphics[trim={0 0 0 0}, clip, width=.16\textwidth]{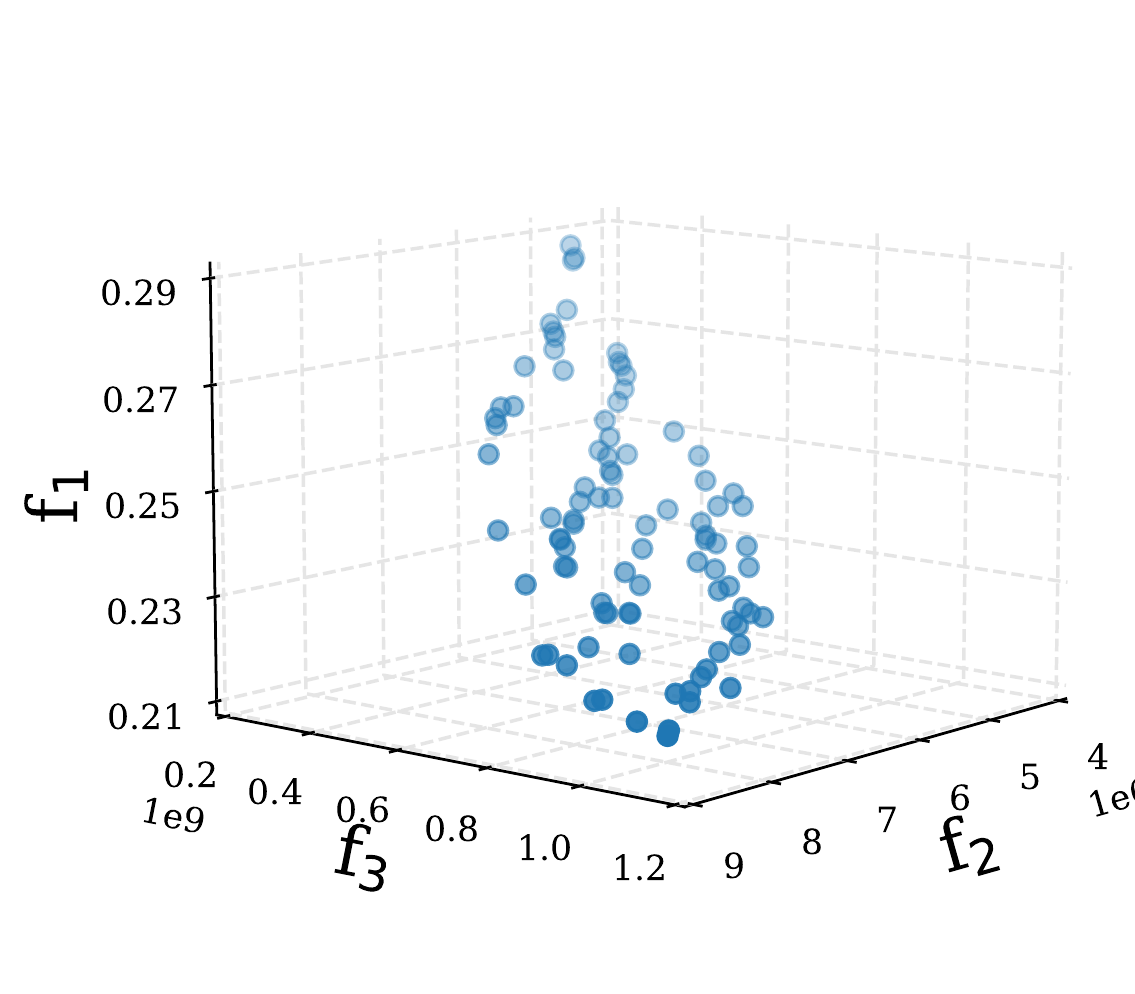}\hfill
        \includegraphics[trim={0 0 0 0}, clip, width=.16\textwidth]{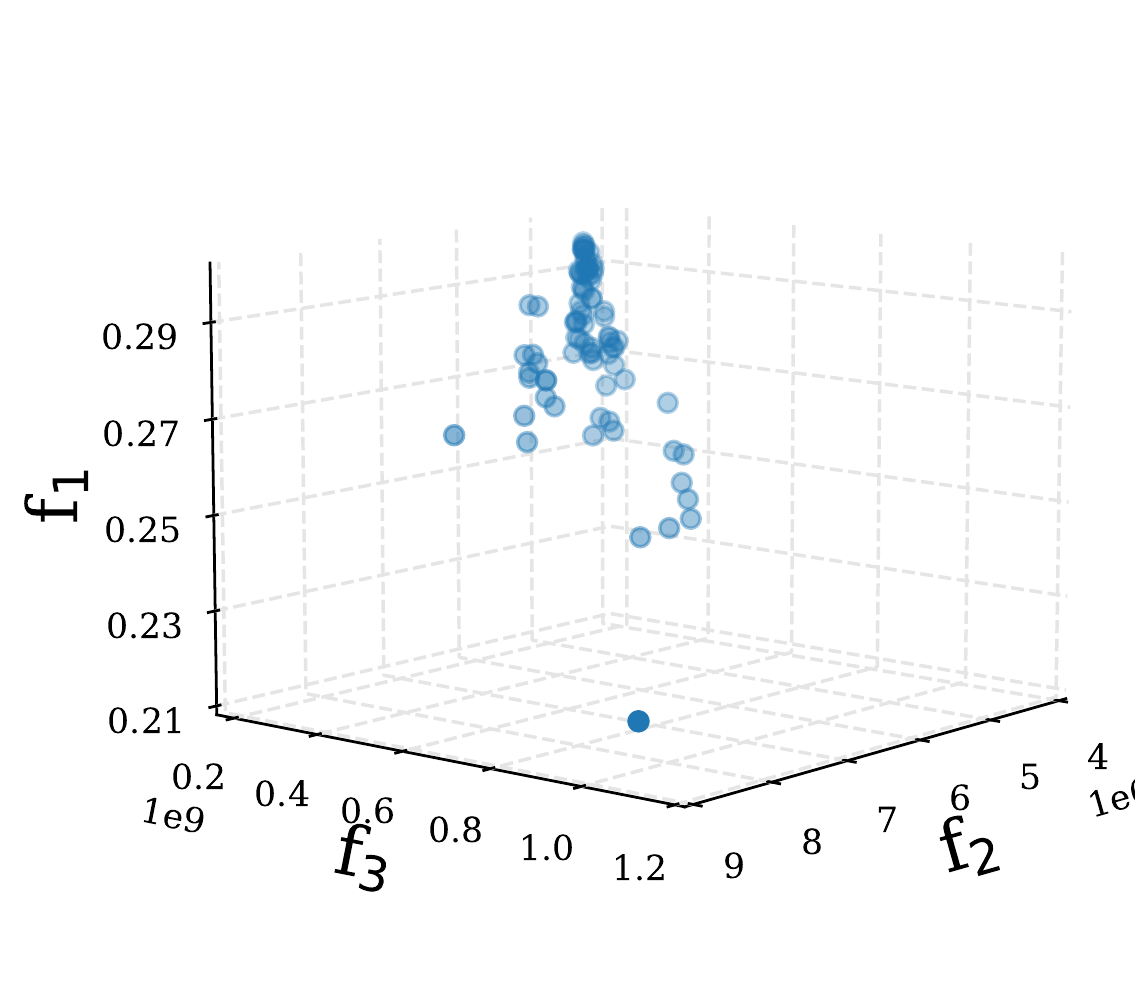}\hfill
        \vspace{-5pt}
        \caption{\secondtestsuite{}8 \label{fig:in1kmop8_pf}}
    \end{subfigure} 
    \caption{Nondominated solutions obtained by each algorithm on (a) \secondtestsuite{}1 and (b) \secondtestsuite{}8. We select the run associated with the median HV value. For each row, the subfigures correspond to NSGA-II, IBEA, MOEA/D, NSGA-III, HypE, and RVEA, respectively. \label{fig:in1kmop_visualization}}
\end{figure*}

This test suite comprises primarily bi-/three-objective MOPs except for the last one -- \secondtestsuite{}9 with four objectives.
The statistical values of the HV metric achieved by the six algorithms are summarized in Table~\ref{tab:in1kmop_hv}.
In general, similar to the results of the previous test suite, we can observe that none of the six algorithms can effectively solve all problems in \secondtestsuite{}.
In particular, on bi-objective problems (i.e., \secondtestsuite{}1, \secondtestsuite{}2, \secondtestsuite{}4, \secondtestsuite{}5, and \secondtestsuite{}7), NSGA-II performs consistently better than other algorithms;
by contrast, on problems with more than two objectives (i.e., \secondtestsuite{}3, \secondtestsuite{}8, and \secondtestsuite{}9), IBEA performs the best among six algorithms.
Additionally, without an explicit normalization mechanism, the decomposition-based algorithms (i.e., MOEA/D and RVEA) perform substantially worse than the others.
For further analysis, we plot the nondominated solutions achieved by each algorithm in the median run on two typical test instances in Fig.~\ref{fig:in1kmop1_pf} and Fig.~\ref{fig:in1kmop8_pf} for \secondtestsuite{}1 and \secondtestsuite{}8 respectively. 
In the following, we provide some discussions based on the observations made from Fig.~\ref{fig:in1kmop_visualization}.

\begin{table*}[ht]
\centering
\caption{Statistical results (median and standard deviation) of the HV values on \secondtestsuite{} test suite. The best results of each instance are in bold. \label{tab:in1kmop_hv}}
\resizebox{.85\textwidth}{!}{%
\begin{threeparttable}
    \begin{tabular}{@{\hspace{2mm}}ccccccc@{\hspace{2mm}}}
    \toprule
    Problem & NSGA-II & IBEA & MOEA/D & NSGA-III & HypE & RVEA \\ \midrule
    \secondtestsuite{}1 & \textbf{0.9299 (0.0045)}$\bm{^+}$ & 0.9246 (0.0054)$^-$               & 0.7447 (0.0494)$^-$  & 0.9196 (0.0124)$^-$  & 0.9020 (0.0075)$^-$ & 0.7662 (0.0372)$^-$ \\
    \secondtestsuite{}2 & \textbf{0.8849 (0.0011)}$\bm{^+}$ & 0.8843 (0.0019)$^-$               & 0.5239 (0.0883)$^-$  & 0.8799 (0.0017)$^-$  & 0.8592 (0.0047)$^-$ & 0.2038 (0.1123)$^-$ \\
    \secondtestsuite{}3 & 0.7945 (0.0051)$^-$               & \textbf{0.8187 (0.0046)}$\bm{^+}$ & 0.0603 (0.0579)$^-$  & 0.7881 (0.0624)$^-$  & 0.8079 (0.0046)$^-$ & 0.7382 (0.0178)$^-$ \\
    \secondtestsuite{}4 & \textbf{0.9930 (0.0048)}$\bm{^+}$ & 0.9876 (0.0058)$^-$               & 0.6234 (0.0700)$^-$  & 0.9882 (0.0052)$^-$  & 0.9311 (0.0264)$^-$ & 0.3872 (0.1290)$^-$ \\
    \secondtestsuite{}5 & \textbf{0.9975 (0.0040)}$\bm{^+}$ & 0.9913 (0.0060)$^-$               & 0.6143 (0.0833)$^-$  & 0.9908 (0.0049)$^-$  & 0.9311 (0.0301)$^-$ & 0.3416 (0.1351)$^-$ \\
    \secondtestsuite{}6 & \textbf{0.9832 (0.0041)}$\bm{^+}$ & 0.9761 (0.0061)$^-$               & 0.1570 (0.0323)$^-$  & 0.9541 (0.0040)$^-$  & 0.8971 (0.0307)$^-$ & 0.8495 (0.0288)$^-$ \\
    \secondtestsuite{}7 & \textbf{0.9049 (0.0134)}$\bm{^+}$ & 0.8895 (0.0197)$^-$               & 0.6324 (0.0707)$^-$  & 0.8843 (0.0164)$^-$  & 0.8687 (0.0156)$^-$ & 0.8119 (0.0280)$^-$ \\
    \secondtestsuite{}8 & 0.6884 (0.0067)$^-$               & \textbf{0.7267 (0.0054)}$\bm{^+}$ & 0.0533 (0.1158)$^-$  & 0.6989 (0.0462)$^-$  & 0.7135 (0.0057)$^-$ & 0.5918 (0.0432)$^-$ \\
    \secondtestsuite{}9 & 0.5783 (0.0000)$^-$               & \textbf{0.6514 (0.0072)}$\bm{^+}$ & 0.1416 (0.1002)$^-$  & 0.5956 (0.0060)$^-$  & 0.6269 (0.0122)$^-$ & 0.4946 (0.0394)$^-$ \\ \bottomrule
    \end{tabular}%
\begin{tablenotes}
\setlength\labelsep{0pt}
\small
\item \hspace{1em}$^{+}$ indicates a method achieving significantly better performance.
\item \hspace{1em}$^{-}$ indicates a method achieving significantly worse performance.
\end{tablenotes}
\end{threeparttable}
}
\end{table*}

\secondtestsuite{}1 is a bi-objective problem of simultaneous minimization of prediction error ($f^e$) and No. of parameters ($f^c_1$). 
In addition to the multi-modal and noisy fitness landscape, another primary challenge of this problem is that the two objectives are of different scales: the objective of $f^e$ falls into the range of $[0, 1]$, while the objective of $f^c_1$ is in the range of millions. 
Consequently, as depicted in Fig.~\ref{fig:in1kmop1_pf}, we can observe that both MOEA/D and RVEA fail to converge to the PF. 
By contrast, NSGA-III achieves comparable performance to indicator-based algorithms (i.e., IBEA and HypE) owing to its internal normalization mechanism. 
Similar observations can be made on \secondtestsuite{}8 test instance from Fig.~\ref{fig:in1kmop8_pf}.

\subsection{Further Discussion}
In addition to the primary goal of facilitating an empirical comparison among EMO algorithms, the results of these test suites also provide valuable insights toward understanding the effectiveness of DNNs from an architectural perspective. 
For instance, let us consider the \firsttestsuite{}5, a five-objective problem for simultaneously optimizing prediction error ($f^e$), model complexity ($\bm{f}^{c}$) and hardware efficiency ($\bm{f}^{h_1}$) on GPUs. 
We visualize the Pareto architectures obtained by NSGA-II in the median run measured by the HV metric. 
More specifically, we consider architectures from three different trade-off subsets of the final nondominated solutions: (i) a subset  preferred for $f^e$ and $\bm{f}^{c}$; (ii) a subset preferred for $f^e$ and $\bm{f}^{h_1}$; and (iii) a subset striking a balance among $f^e$, $\bm{f}^{c}$, and $\bm{f}^{h_1}$. 

\begin{figure*}[t]
    \centering
    \begin{subfigure}[b]{0.49\textwidth}
        \centering
        \includegraphics[trim={0 0 0 0}, clip, width=\textwidth]{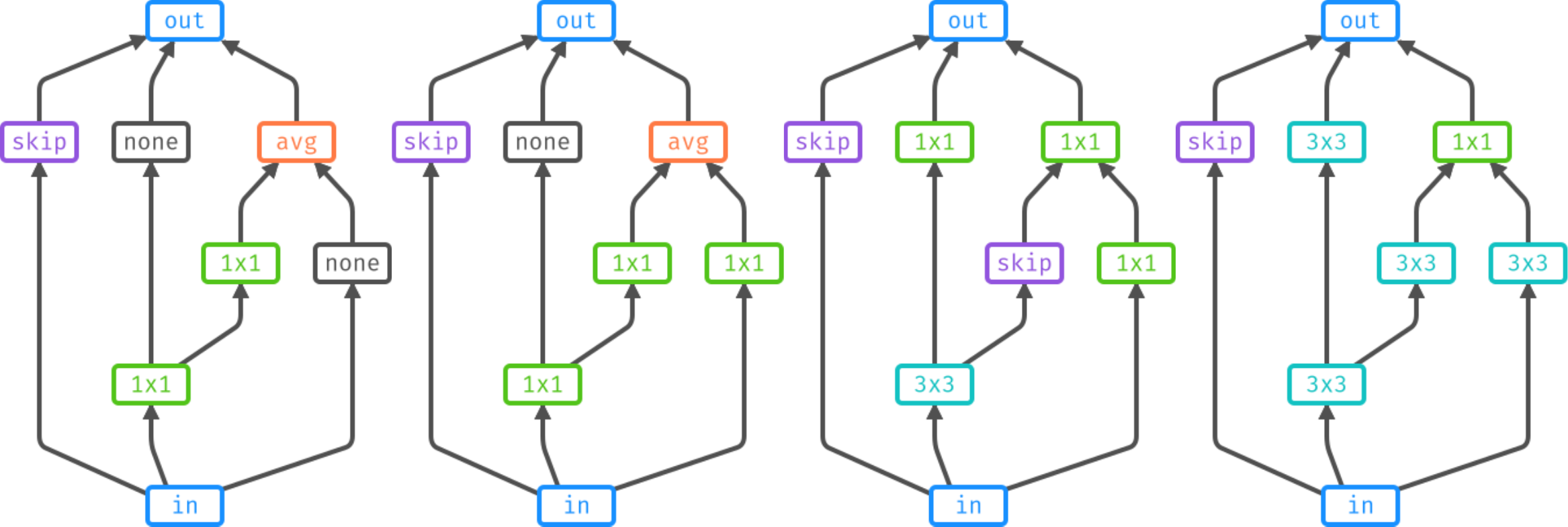}
        \caption{\label{fig:archs_nb201_params}}
    \end{subfigure} \hfill
    \begin{subfigure}[b]{0.49\textwidth}
        \centering
        \includegraphics[trim={0 0 0 0}, clip, width=\textwidth]{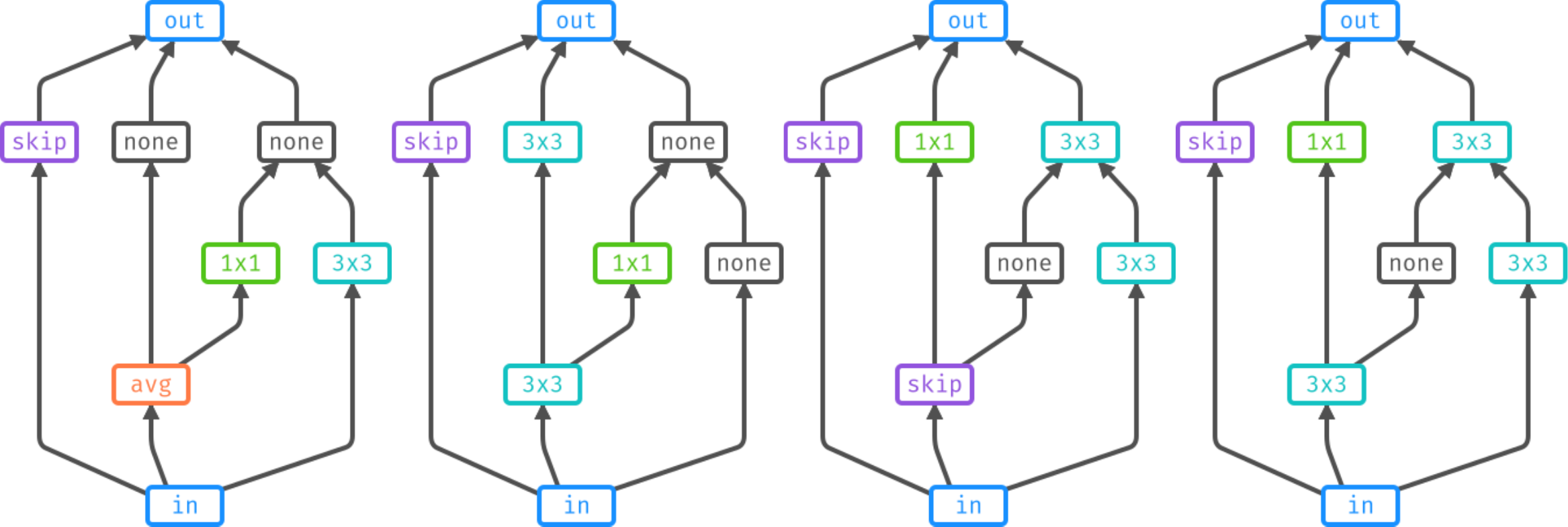}
        \caption{\label{fig:archs_nb201_latency}}
    \end{subfigure}\\
    \begin{subfigure}[b]{0.49\textwidth}
        \centering
        \includegraphics[trim={0 0 0 0}, clip, width=\textwidth]{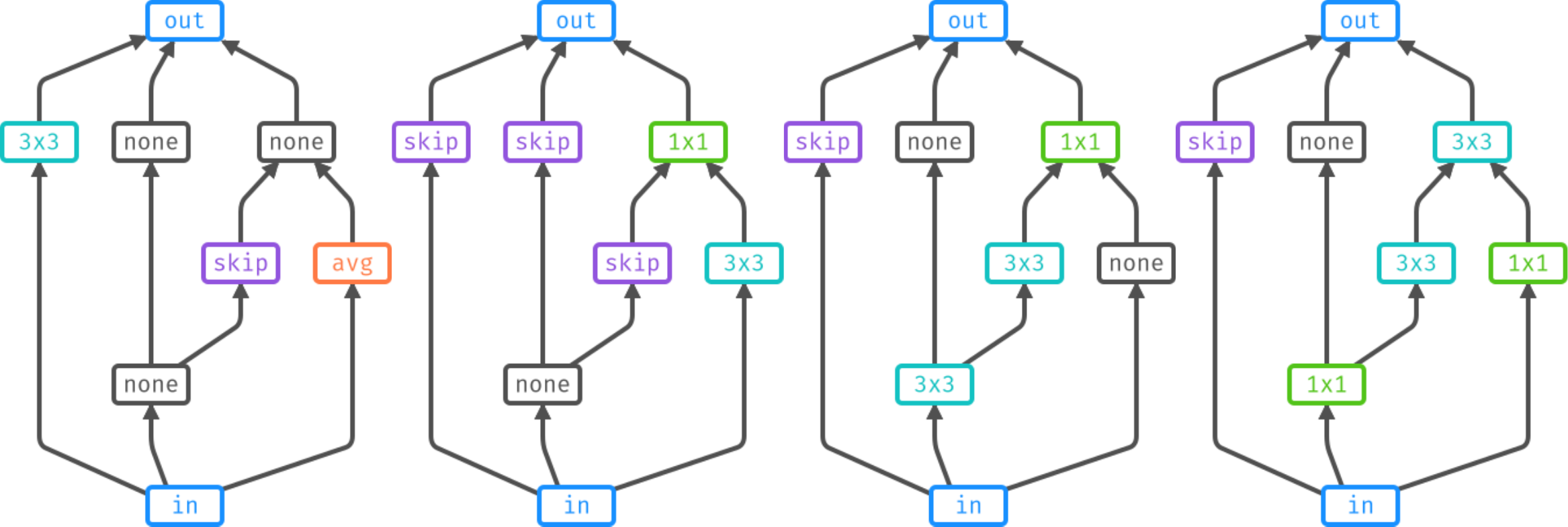}
        \caption{\label{fig:archs_nb201_union}}
    \end{subfigure} \hfill
    \begin{subfigure}[b]{0.49\textwidth}
        \centering
        \includegraphics[trim={0 0 0 0}, clip, width=\textwidth]{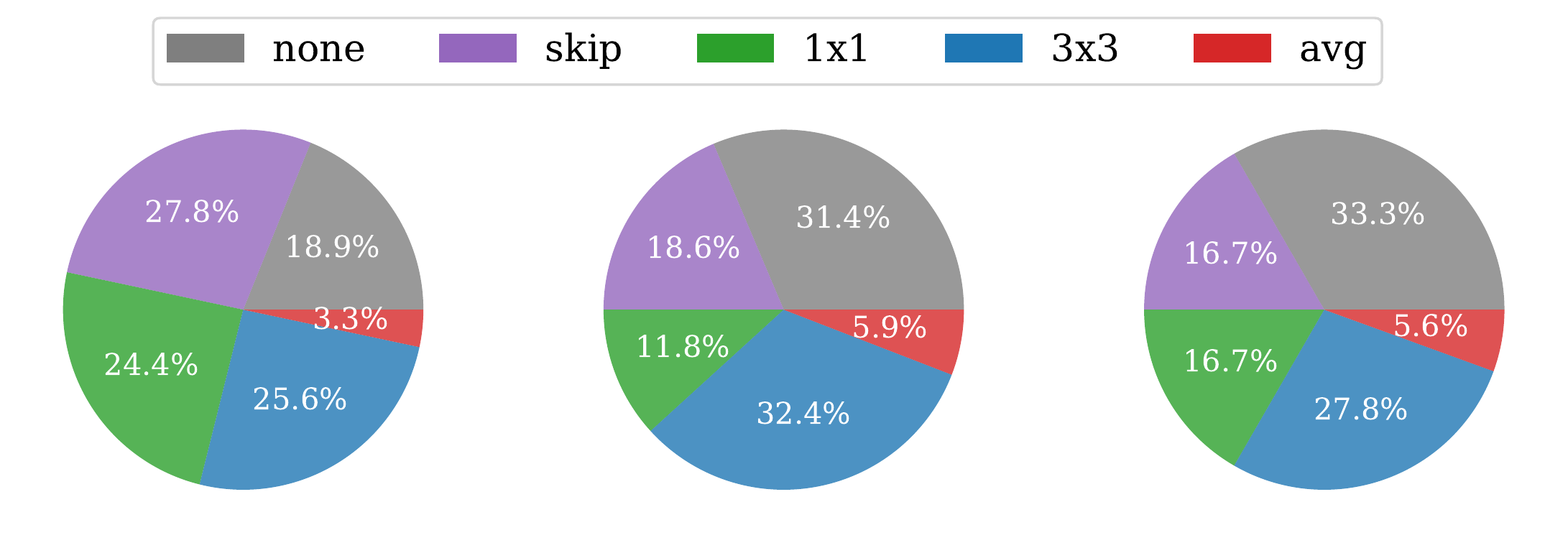}
        \caption{\label{fig:operator_distribution_nb201}}
    \end{subfigure}
    \caption{The architectures obtained by NGSA-II on \firsttestsuite{}5 in the run associated with the median HV value.
    We visualize architectures from three different trade-off subsets: (a) the subset preferred for prediction error and model complexity, i.e., nondominated considering prediction error ($f^e$), No. of parameters ($f^c_1$) and FLOPs ($f^c_2$); (b) the subset preferred for prediction error and hardware efficiency, i.e., nondominated considering prediction error ($f^e$), GPU latency ($f^{h_1}_1$) and energy ($f^{h_1}_2$); (c) the subset striking a balance between prediction error, model complexity, and hardware efficiency. (d) We also provide the frequency of the operators (i.e., decision variables of \firsttestsuite{}5) used by these three subsets respectively (from left to right). \label{fig:archs_nb201} 
    }
\end{figure*}

As depicted in Fig.~\ref{fig:operator_distribution_nb201}, there are consistent trends on the architectural choices (i.e., decision variables) across three different scenarios. 
In particular, we can observe that the $3\times3$ average pooling (i.e., ``avg'' in Fig.~\ref{fig:archs_nb201}) is the least frequently used operator, while the $3\times3$ convolution (i.e., ``$3\times3$'' in Fig.~\ref{fig:archs_nb201}) is preferred consistently. 
Given that $f^e$ is the common objective in all three scenarios, we argue that the $3\times3$ convolution is important for achieving a better prediction error, while the $3\times3$ average pooling adversely affects the prediction error. 
Moreover, we can observe that a transition from the $f^e$-$\bm{f}^{c}$ preferred subset to the $f^e$-$\bm{f}^{h_1}$ preferred subset can be achieved by gradually dropping the $1\times1$ convolution (i.e., replace ``$1\times1$'' with ``none'' in Fig.~\ref{fig:archs_nb201_params} and \ref{fig:archs_nb201_latency}). 

In the meantime, the results of these test suites also provide valuable feedback on hardware devices. 
For instance, as depicted in Fig.~\ref{fig:c10mop6_pf}, the inference latency, on one hand, is correlated (objective 4 in Fig.~\ref{fig:c10mop6_pf}) with the energy consumption (objective 5 in Fig.~\ref{fig:c10mop6_pf}) on Eyeriss\footnote{The lines between objective 4 and objective 5 are in parallel.}, i.e., a DNN with a faster inference speed tends to draw less energy as well;
on the other hand, however, trade-offs exist between the inference speed/energy consumption and the arithmetic intensity (objective 6 in Fig.~\ref{fig:c10mop6_pf}; defined as operations per unit memory traffic)\footnote{The lines between objectives 5 and 6 are in a crisscross pattern.}.

As demonstrate above, running EMO algorithms on \ourbenchmark{} provides rich results worthy of deep investigations.
Beyond the main target of NAS for automating hardware-related design/deployment of DNNs, the empirical observations can be also useful in hardware designs.

%% file: 7-conclusion.tex
\section{Conclusion\label{sec:conclusion}}
This paper was devoted to bridging the gap between EMO and NAS. 
First, we provided a general multiobjective formulation of NAS together with the analyses of complex characteristics from the optimization point of view. 
Then, we presented an end-to-end pipeline -- \ourbenchmark{}, to streamline the generation of benchmark test problems for EMO algorithms to run efficiently without the requirements of GPUs or sophisticated software such as Pytorch/TensorFlow.
Next, we initiated two test suites comprising two image classification datasets, seven search spaces, three types of hardware, and up to eight objectives.
Finally, an experimental study was conducted using six representative algorithms implemented in MATLAB, having demonstrated the effectiveness of \ourbenchmark{} and provided insightful analyses of the results. 

On top of the generic problem formulations of NAS problems, \ourbenchmark{} provides a generic framework which can be extended to include additional search spaces, hardware devices, and tasks by following the pipeline as described in the paper. 
Meanwhile, more benchmark test instances can be also generated via \ourbenchmark{}.

%% file: 8-appendix.tex



    


\section{Background Continued\label{sec:app_background}}

\subsection{Extended Description of Evolutionary Multiobjective Optimization}\label{sec:app_background_emo}

Dating back to the 1990s, the \emph{dominance-based} EMO algorithms were originally motivated to tailor the selection operators of EAs on the basis of \emph{Pareto dominance}~\cite{fonseca1993genetic, srinivas1994muiltiobjective}.
Later, it was proposed that the use of \emph{elitism mechanism} would substantially improve the convergence of the algorithms.
Afterward, the emergence of several milestone algorithms such as  the elitist non-dominated sorting genetic algorithm (NSGA-II)~\cite{deb2002fast}, the strength Pareto evolutionary algorithms (SPEA2)~\cite{zitzler2001spea2} and the Pareto envelope-based selection algorithm (PESA)~\cite{corne2001pesa} has laid a solid foundation for future research of the field.

Following the classic divide-and-conquer strategies as widely adopted in algorithm designs, more recently, some \emph{decomposition-based} approaches have been proposed to divide a complex MOP into a number of sub-problems (either multiobjective or single objective ones) and conquer them in a collaborative manner.
As the milestone EMO algorithm of this category, the multiobjective evolutionary algorithm based on decomposition (MOEA/D)~\cite{moead} was proposed to divide an MOP into a number of scalar optimization sub-problems via weighted aggregation of the multiple objectives.
Another representative algorithm of this category is the MOEA/D-M2M~\cite{liu2013decomposition}, which was proposed to divide an MOP into a number of simple multiobjective sub-problems via objective space division.

Since the performance of an EMO algorithm is measured by the performance indicators, intuitively, various \emph{indicator-based} EMO algorithms have been proposed.
An early milestone is the binary-indicator-($I_{\epsilon +}$ )-based evolutionary algorithm (IBEA) \cite{zitzler2004indicator}, where $I_{\epsilon +}$ is considered to be compliant with the Pareto dominance.
Following this pathway, some works proposed to integrate EMO algorithms with a more powerful performance indicator -- the hypervolume indicator (or the S metric), e.g., the s-metric selection evolutionary multiobjective algorithm (SMS-EMOA)~\cite{emmerich2005emo}.

\subsection{Literature Review of NAS Continued}

\subsubsection{A Brief Review of Existing NAS Search Algorithms}

In general, existing NAS approaches can be categorized into reinforcement learning (RL) based ones, differentiable/gradient-based ones (DARTS), and evolutionary algorithm (EA) based ones. 
RL treats the design of network architecture as a sequential decision process, where an agent is trained to optimally choose pieces to assemble the network \cite{zoph2016}. 
Despite that modeling NAS into an RL task is intuitive, RL-based approaches often require a large number of iterations (a.k.a \emph{episodes}) to converge, resulting in weeks of wall clock time on hundreds of GPUs. 
DARTS seeks to improve search efficiency by problem reformulation via continuous relaxation, allowing the architectural parameters (i.e., the decision variables of a NAS problem) to be jointly optimized with the weights by gradient descent through back-propagation \cite{liu2018darts}. 
However, such a problem reformulation suffers from excessive GPU memory requirements during the search, resulting in overly confined search spaces such as reduced layer operation choices.
EA treats NAS as a black-box discrete optimization problem, where a population of candidate solutions is iterated to be gradually better via genetic operations or heuristics \cite{xie2017genetic}. 
Due to its modular framework, flexible encoding, and population-based nature, EA has attracted increasing attention, leading to a plethora of emerging EA-based NAS approaches \cite{real2017large,suganuma2017genetic,lu2019nsga,8712430}. 

Despite that, there are advanced methods such as weight sharing \cite{pham2018efficient} to improve RL's search efficiency, or the binary gating \cite{cai2018proxylessnas} to reduce the GPU memory footprint of DARTS, most existing non-EA-based methods are not directly applicable to multiobjective NAS. 
To this end, this work is dedicated to standardizing the problem of NAS in terms of both formulation and benchmarking, such that previous algorithmic efforts from the EMO community can be inherited and future algorithm designs can be properly and fairly assessed. 

\subsubsection{A Brief Review of Existing NAS Fitness Evaluators}

A thorough training from scratch to assess the prediction error (i.e., fitness) of an architecture is computationally prohibitive for realistically-sized search spaces \cite{nasnet2018,liu2018darts,tan2019mnasnet}. 
Consequently, architectures are almost always evaluated \emph{imperfectly} during a NAS process. 
Specifically, architectures are typically evaluated with the following methods: 
(i) proxy tasks (e.g., small-scale architectures \cite{nsganet_tevc} and fewer training epochs \cite{baker2018accelerating}); 
(ii) surrogate models (e.g., inherited weights from a supernet \cite{9672175,9488309} or a ``neighbor'' architecture (network morphism) \cite{elsken2018efficient}, performance predictors \cite{e2epp,liu2022survey}); 
(iii) a fixed set of training hyperparameters, neglecting the fact that different architectures are likely to enjoy different sets of hyperparameters to be optimal in training \cite{dai2021fbnetv3}. 

\section{Design of Search Spaces Continued\label{sec:app_search_space}}

All search spaces in \ourbenchmark{} follow the canonical schema outlined in Algorithm~\ref{algo:search_space}. Specifically, a search space object facilitates the following four functions:
\begin{enumerate}
    \item \texttt{is\_valid}: a method to verify the validity of an architecture phenotype (\texttt{arch}) before an architecture can be sent for fitness evaluation. 
    
    \item \texttt{sample}: a method to randomly create a \emph{valid} architecture genotype \texttt{x}. Note that, for some search spaces such as NB101, the default sampling methods (e.g., uniform or Latin hypercube) in existing EMO algorithms do not always produce a valid solution. 
    
    \item \texttt{encode}: a method to convert an architecture phenotype (\texttt{arch}) to its corresponding genotype (\texttt{x}). Note that this is \emph{NOT} a one-to-one mapping, i.e., there exist many genotypes that correspond to one (the same) phenotype. 
    
    \item \texttt{decode}: a method to convert an architecture genotype (\texttt{x}) to its corresponding phenotype (\texttt{arch}). 
\end{enumerate}

\begin{algorithm}[t]
\caption{Pseudocode of a search space class\label{algo:search_space}}
\definecolor{codeblue}{rgb}{0.25,0.5,0.5}
\definecolor{codekw}{rgb}{0.85, 0.18, 0.50}
\begin{lstlisting}[language=python, mathescape]
class SearchSpace():
    def __init__(self, **kwargs):
        arch: phenotype  # used for querying fitness
        x: genotype  # used by genetic operators
    
    def is_valid(arch):
        # a method to verify the validity of a solution
        return True/False
        
    def sample():
        # a method to randomly create a solution
        return x
        
    def encode(arch):
        # a method to convert phenotype to genotype
        return x
    
    def decode(x):
        # a method to convert genotype to phenotype 
        return arch
\end{lstlisting}
\end{algorithm}

The default \texttt{encode} method in \ourbenchmark{} is a simple concatenation of integer values to indicate the choices of architectural elements. We acknowledge that the \texttt{encode} method itself is an important design choice for NAS and can (or rather should) be studied/explored further. However, such studies are beyond the scope of this paper.   

\subsection{Fitness Landscape}
In order to visualize the fitness landscapes of the seven search spaces shown in Table II, we need to project the high-dimensional decision variables to a two-dimensional space. We consider two projection methods, i.e., principal component analysis (PCA) and t-distributed stochastic neighbor embedding (t-SNE) \cite{tsne}. And we use 10,000 solutions uniformly sampled from each search space. The visualizations are provided in Fig.~A1 and Fig.~A2 respectively for t-SNE and PCA.

\begin{figure}[ht]
    \centering
    \begin{subfigure}[b]{0.24\textwidth}
        \centering
        \includegraphics[trim={0 0 0 0}, clip, width=.98\textwidth]{figures/nb101_landscape_tsne.pdf}
        \caption{$\Omega$ = NB101 \label{fig:app_nb101_landscape_tsne}}
    \end{subfigure}\hfill
    \begin{subfigure}[b]{0.24\textwidth}
        \centering
        \includegraphics[trim={0 0 0 0}, clip, width=.98\textwidth]{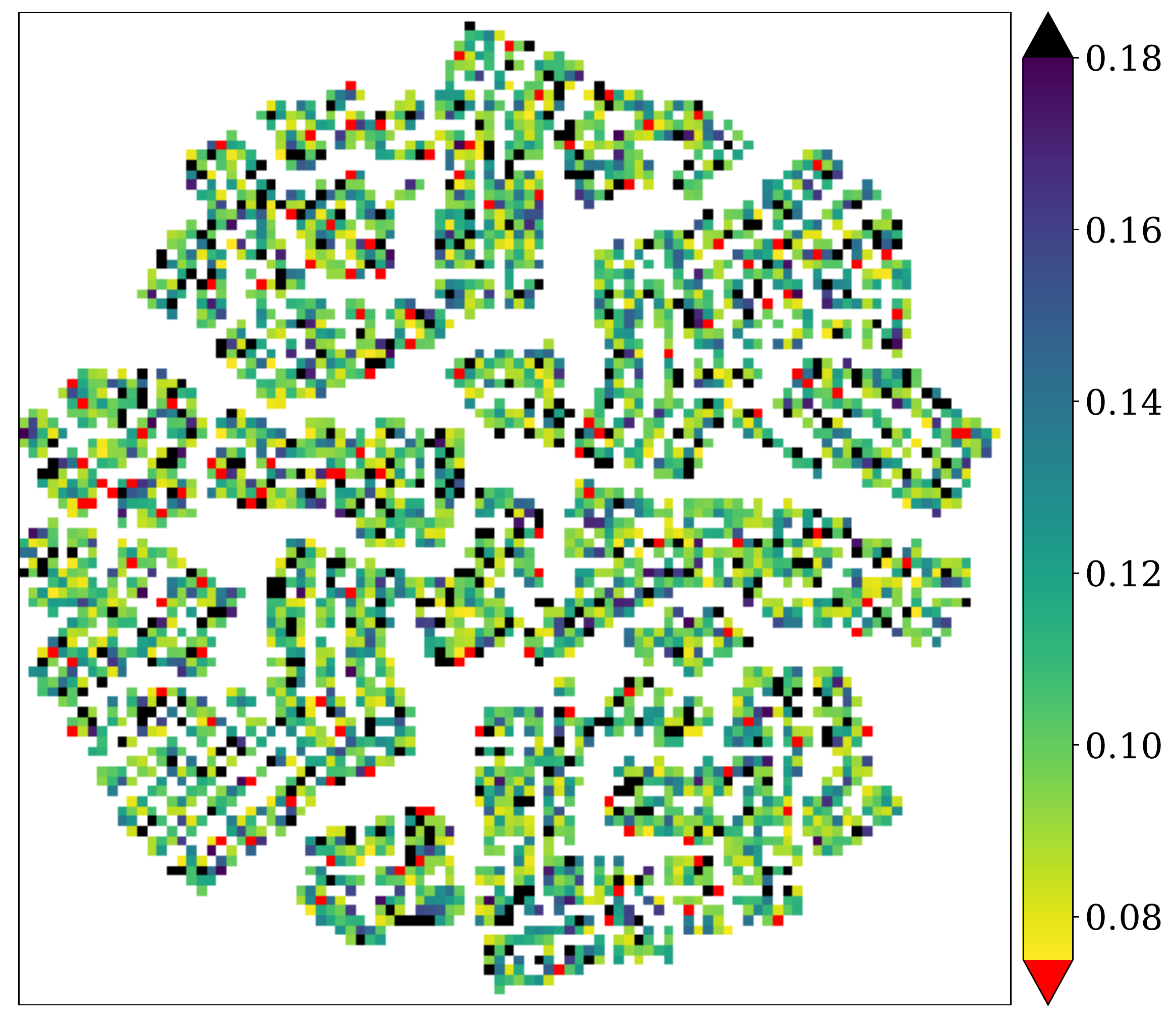}
        \caption{$\Omega$ = NATS \label{fig:nb201_landscape_tsne}}
    \end{subfigure}\\
    \begin{subfigure}[b]{0.24\textwidth}
        \centering
        \includegraphics[trim={0 0 0 0}, clip, width=.98\textwidth]{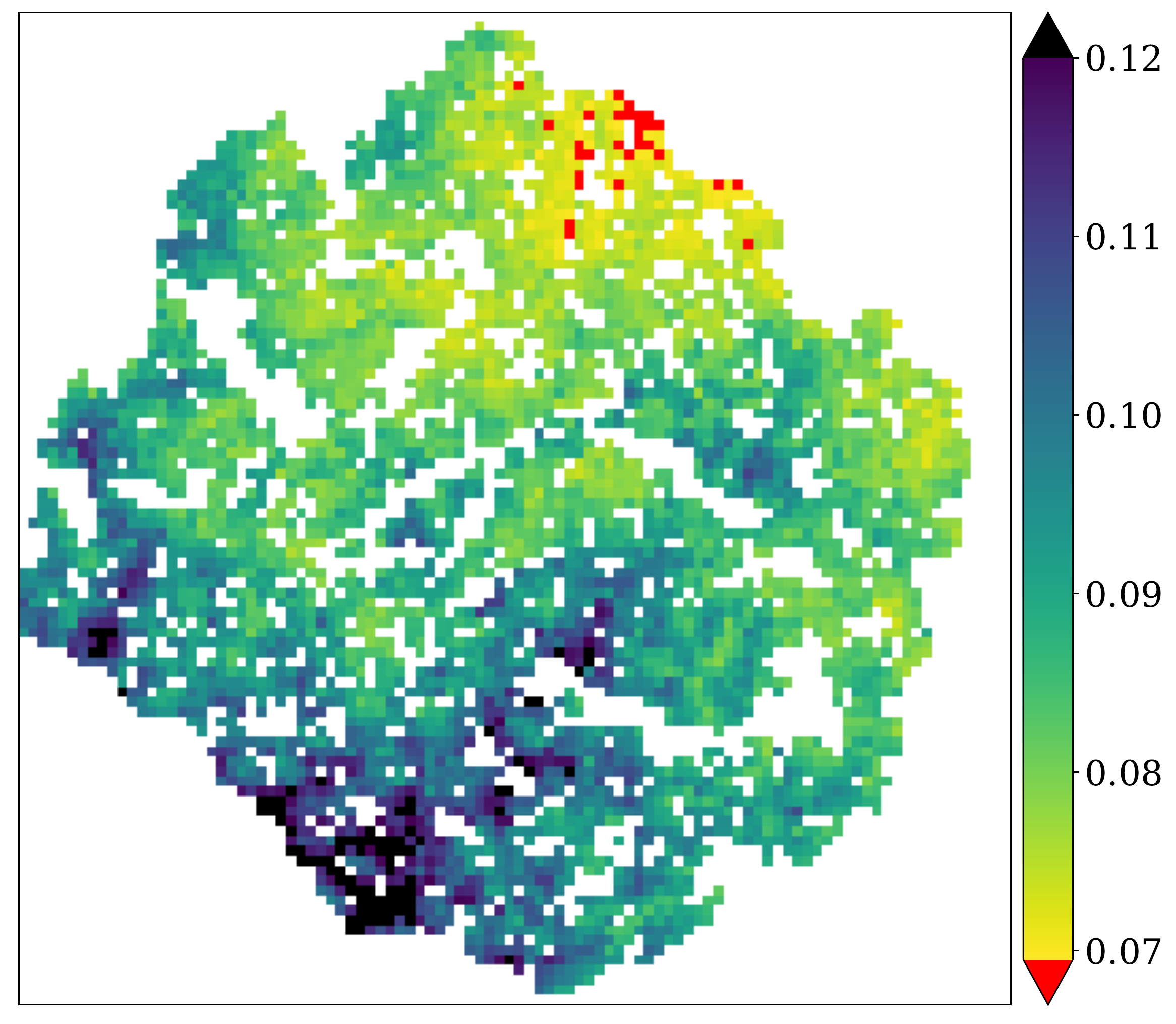}
        \caption{$\Omega$ = NATS \label{fig:nats_landscape_tsne}}
    \end{subfigure}\hfill
    \begin{subfigure}[b]{0.24\textwidth}
        \centering
        \includegraphics[trim={0 0 0 0}, clip, width=.98\textwidth]{figures/darts_landscape_tsne.pdf}
        \caption{$\Omega$ = DARTS \label{fig:app_darts_landscape_tsne}}
    \end{subfigure}\\
    \begin{subfigure}[b]{0.24\textwidth}
        \centering
        \includegraphics[trim={0 0 0 0}, clip, width=.98\textwidth]{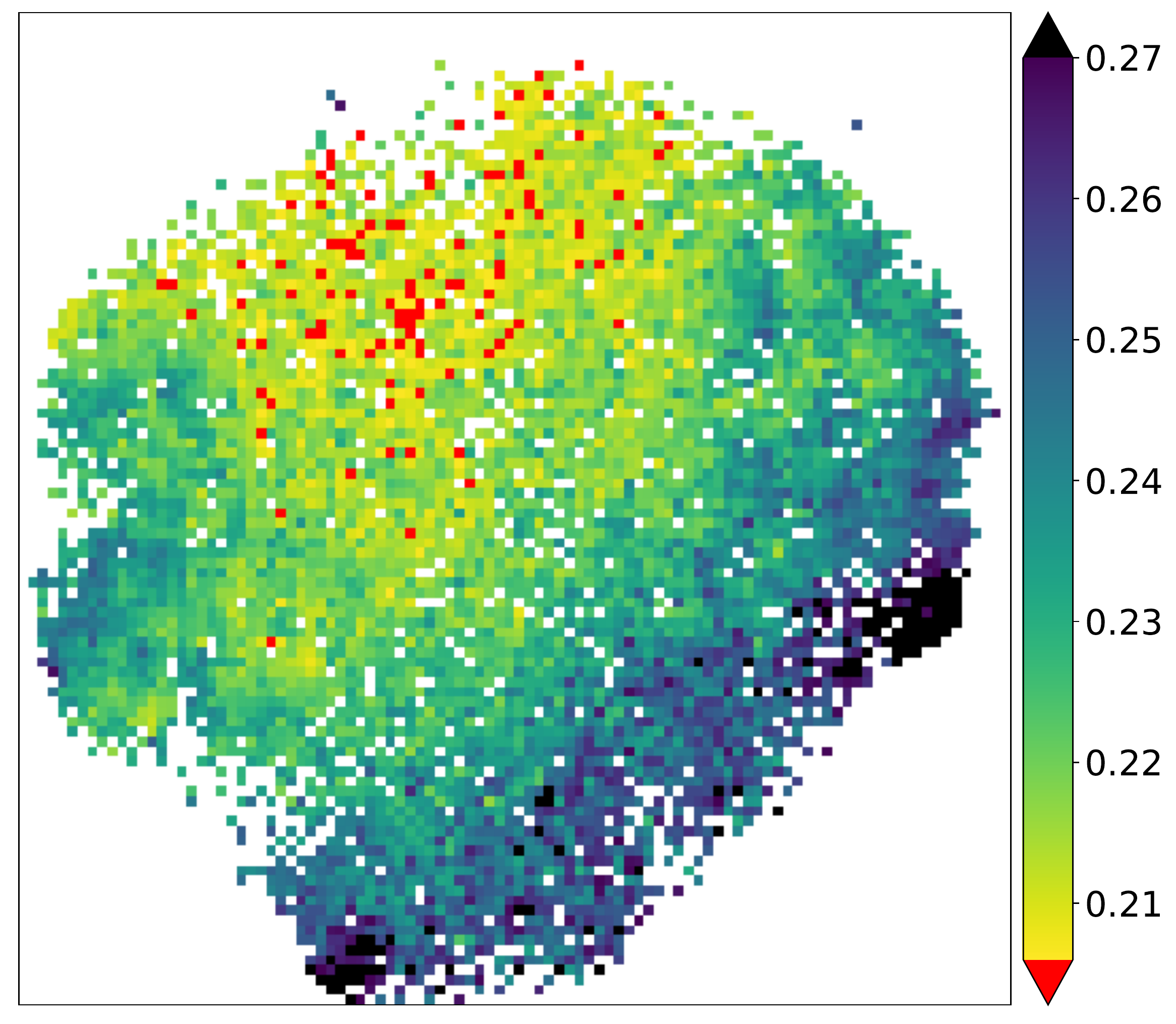}
        \caption{$\Omega$ = ResNet50 \label{fig:r50_landscape_tsne}}
    \end{subfigure}\hfill
    \begin{subfigure}[b]{0.24\textwidth}
        \centering
        \includegraphics[trim={0 0 0 0}, clip, width=.98\textwidth]{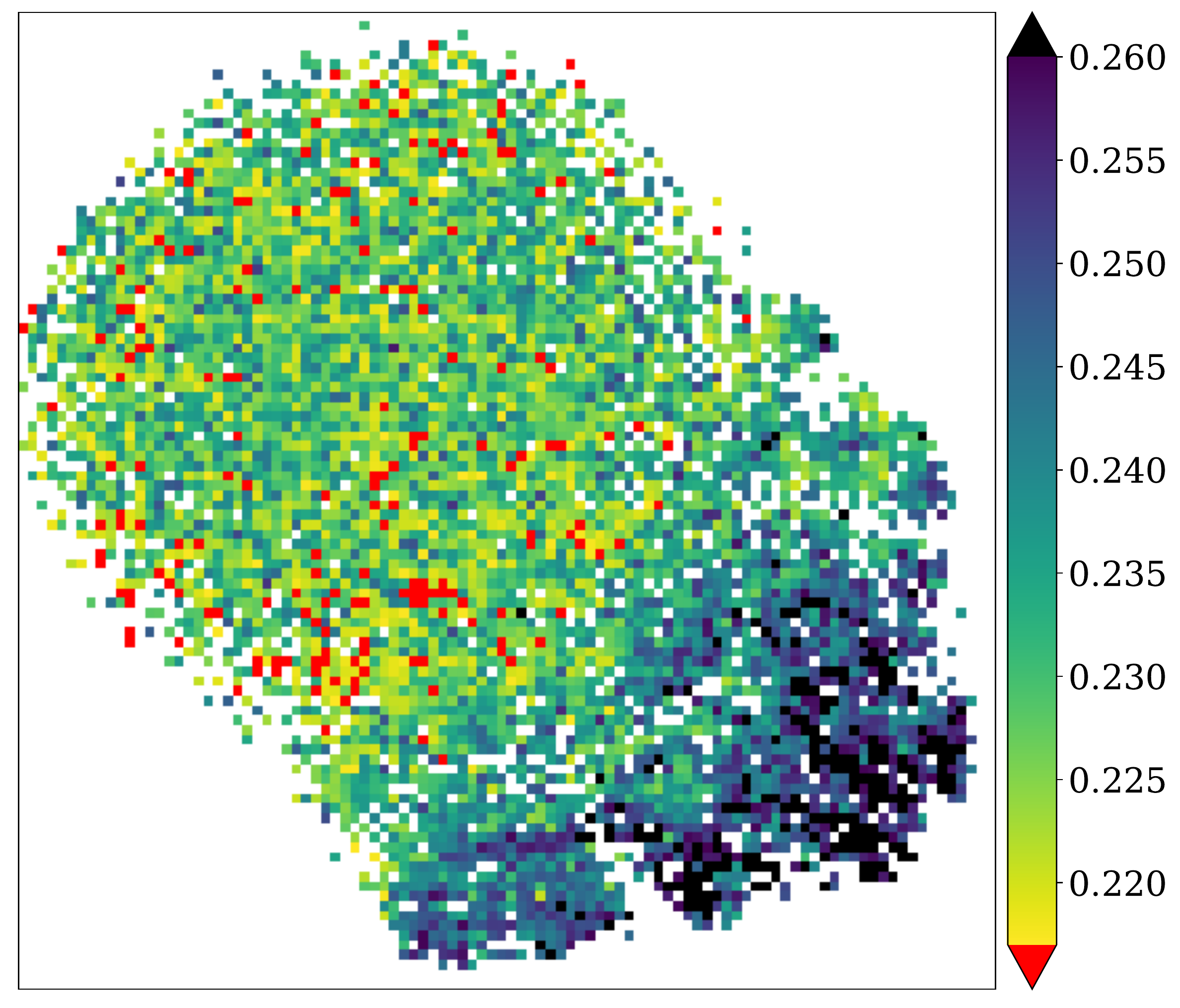}
        \caption{$\Omega$ = MNV3 \label{fig:mnv3_landscape_tsne}}
    \end{subfigure}
    \caption*{Fig.~A1: Fitness landscapes of $f^e$ (i.e. prediction error) on (a) NB201, (b) NATS, (c) ResNet50, and (d) MNV3. We project the original high-dimensional decision space to 2D latent space via t-SNE \cite{tsne}. We random sample 10K solutions from each search space and average ${f}^e$ within each small area. The top-performing solutions are highlighted in red.
    \label{fig:multimodal_lanscape_tsne}}
\end{figure}

\begin{figure}[ht]
    \centering
    \begin{subfigure}[b]{0.24\textwidth}
        \centering
        \includegraphics[trim={0 0 0 0}, clip, width=.98\textwidth]{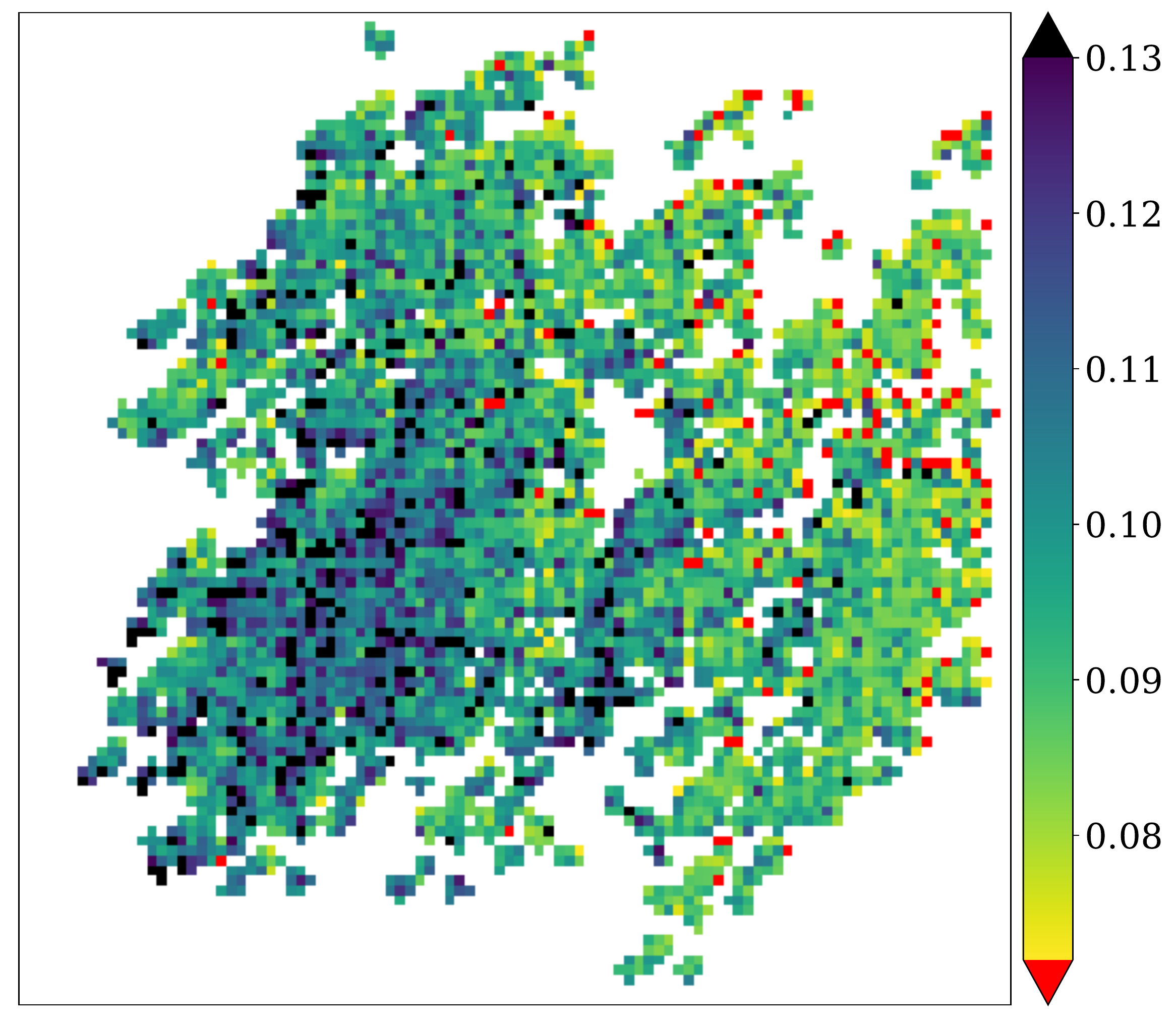}
        \caption{$\Omega$ = NB101 \label{fig:nb101_landscape_pca}}
    \end{subfigure}\hfill
    \begin{subfigure}[b]{0.24\textwidth}
        \centering
        \includegraphics[trim={0 0 0 0}, clip, width=.98\textwidth]{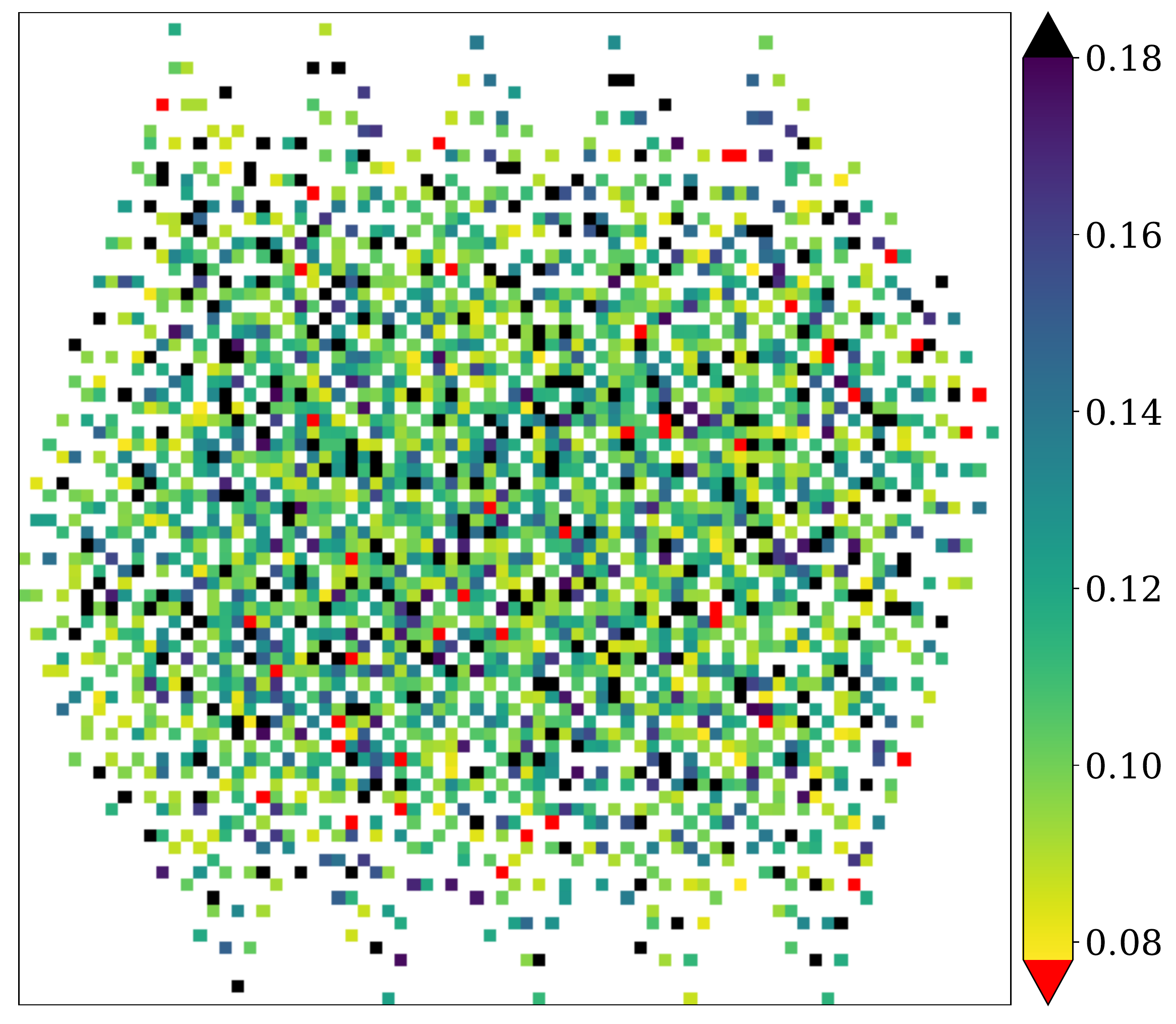}
        \caption{$\Omega$ = NATS \label{fig:nb201_landscape_pca}}
    \end{subfigure}\\
    \begin{subfigure}[b]{0.24\textwidth}
        \centering
        \includegraphics[trim={0 0 0 0}, clip, width=.98\textwidth]{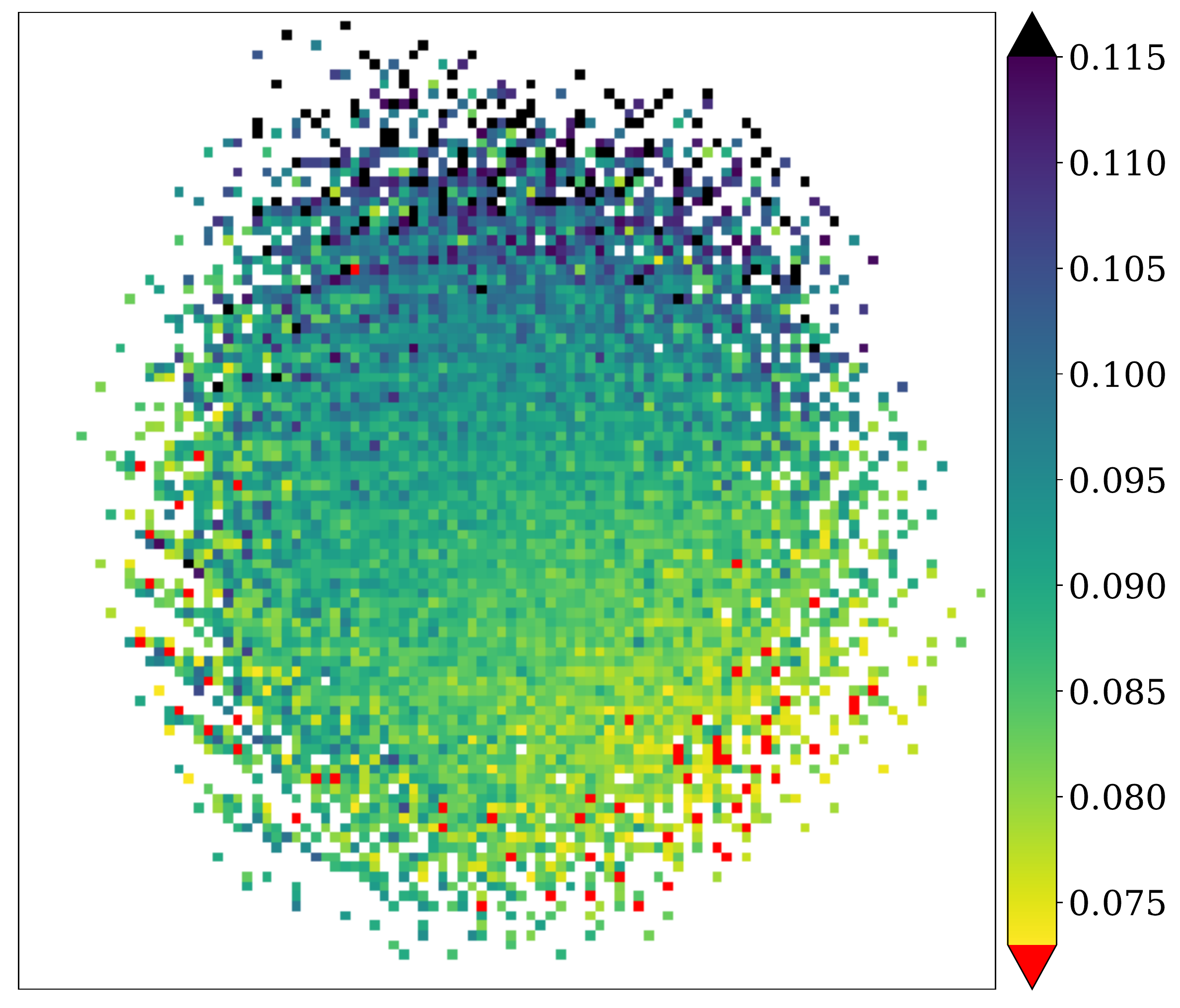}
        \caption{$\Omega$ = NATS \label{fig:nats_landscape_pca}}
    \end{subfigure}\hfill
    \begin{subfigure}[b]{0.24\textwidth}
        \centering
        \includegraphics[trim={0 0 0 0}, clip, width=.98\textwidth]{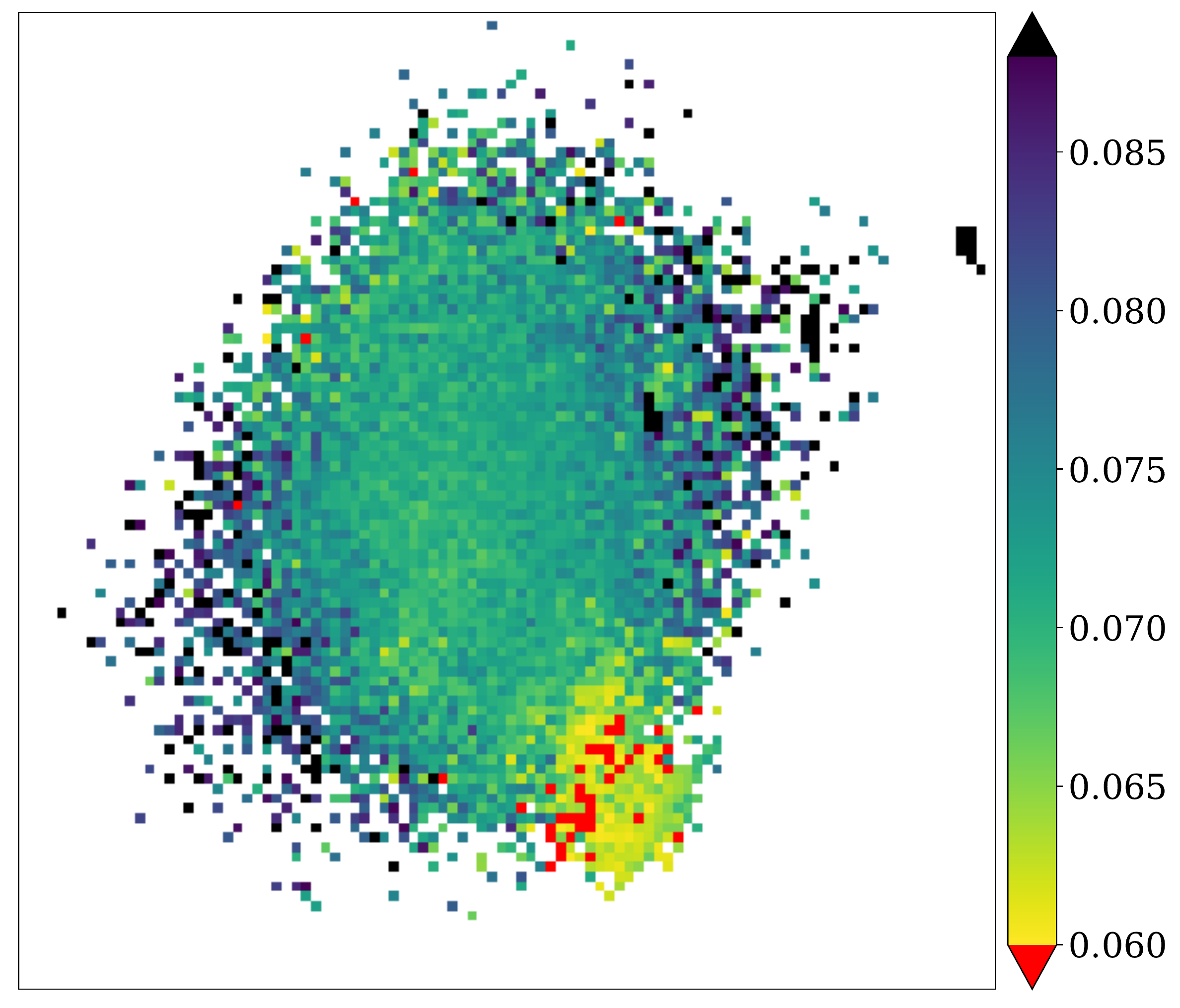}
        \caption{$\Omega$ = DARTS \label{fig:darts_landscape_pca}}
    \end{subfigure}\\
    \begin{subfigure}[b]{0.24\textwidth}
        \centering
        \includegraphics[trim={0 0 0 0}, clip, width=.98\textwidth]{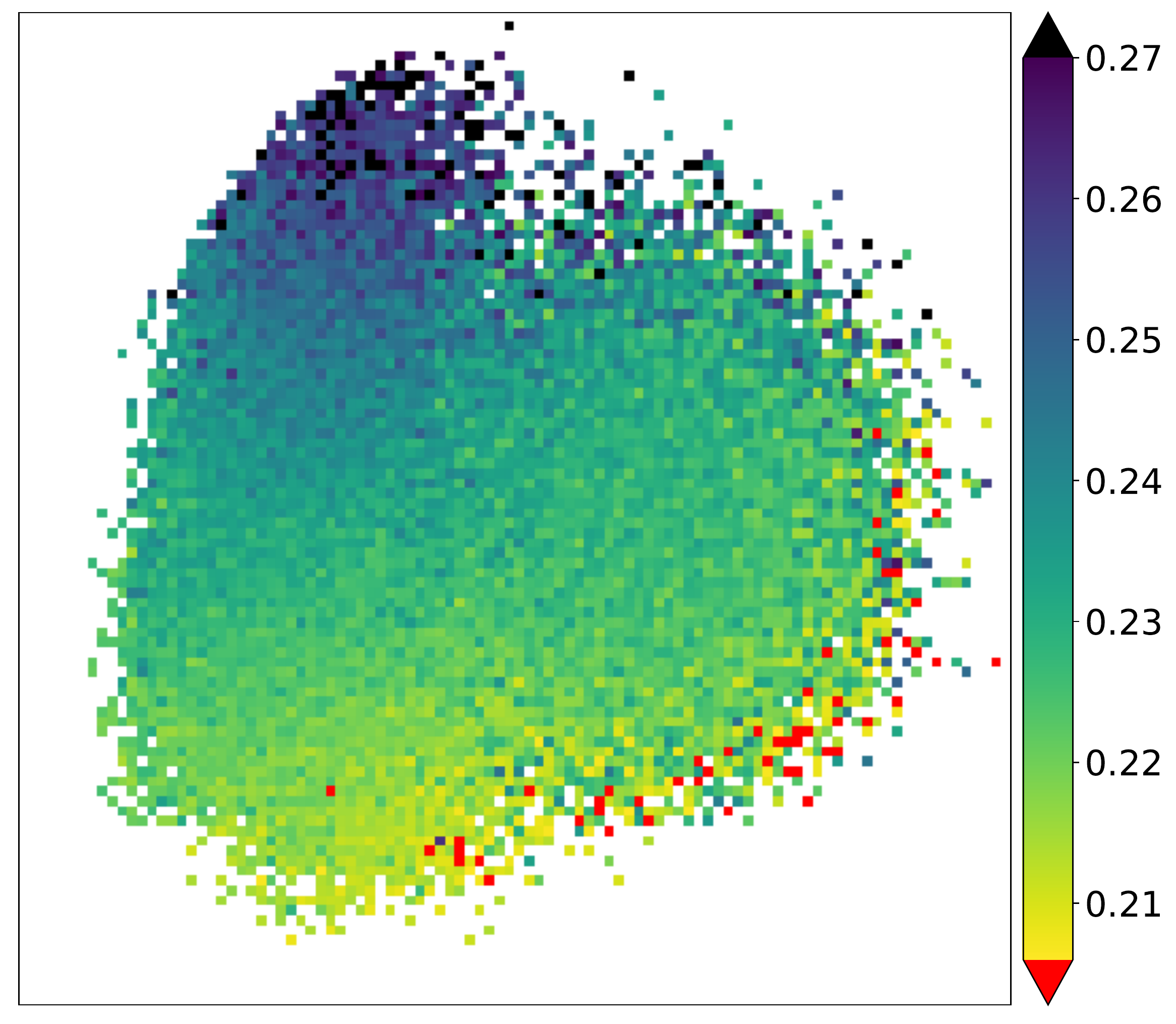}
        \caption{$\Omega$ = ResNet50 \label{fig:r50_landscape_pca}}
    \end{subfigure}\hfill
    \begin{subfigure}[b]{0.24\textwidth}
        \centering
        \includegraphics[trim={0 0 0 0}, clip, width=.98\textwidth]{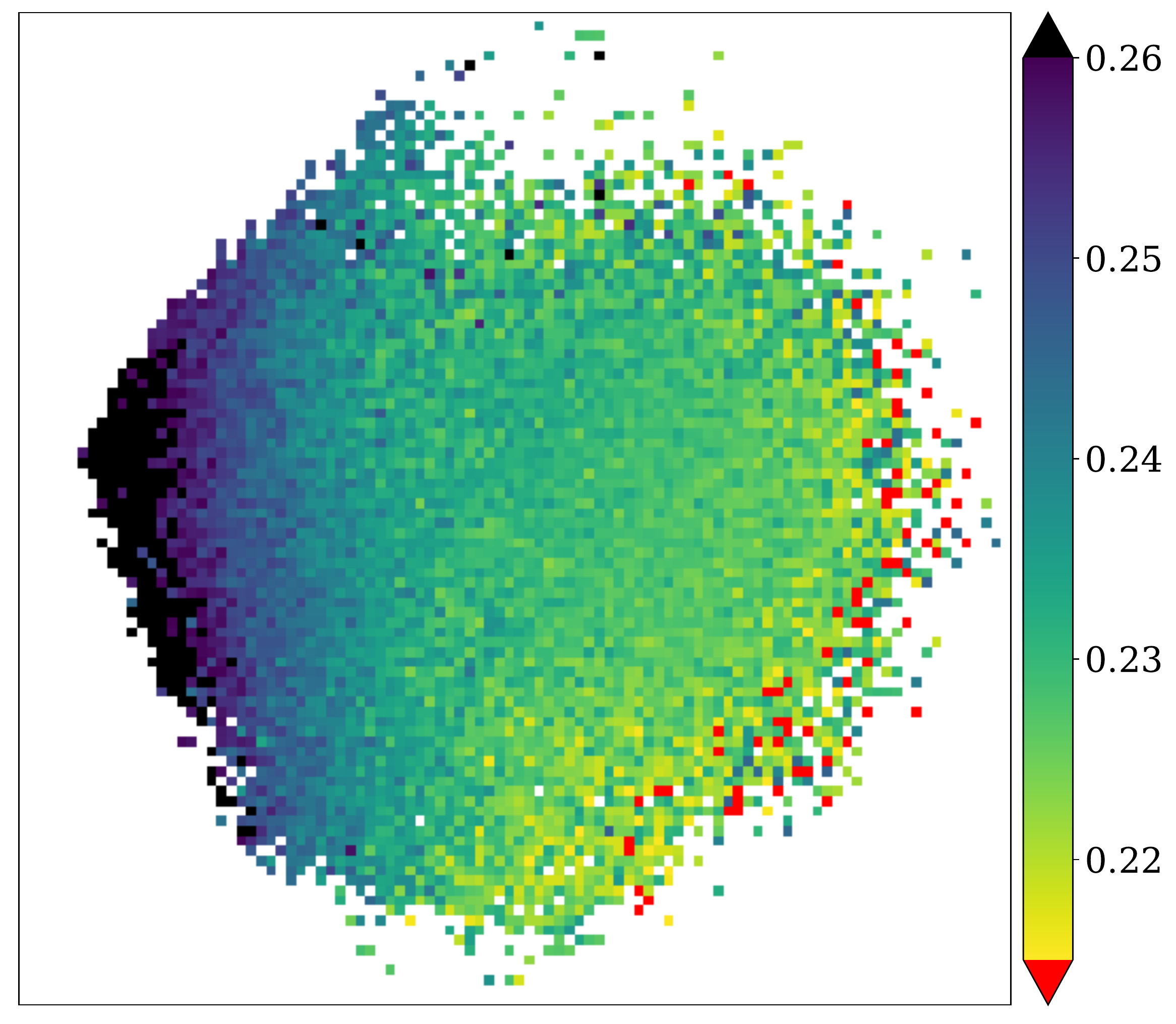}
        \caption{$\Omega$ = MNV3 \label{fig:mnv3_landscape_pca}}
    \end{subfigure}
    \caption*{Fig.~A2: Fitness landscapes of $f^e$ (i.e. prediction error) on (a) NB201, (b) NATS, (c) ResNet50, and (d) MNV3. We project the original high-dimensional decision space to 2D latent space via PCA. We random sample 10K solutions from each search space and average ${f}^e$ within each small area. The top-performing solutions are highlighted in red.
    \label{fig:multimodal_lanscape_pca}}
\end{figure}

\section{Relation between Number of Parameters and FLOPs\label{sec:app_relation}}

In general, the nature of these two objectives (i.e., number of parameters and FLOPs) depends on the search space and the architecture. 
Hence, whether these two objectives are correlated or in conflict varies from case to case. 
We have visualized the correlation between these two objectives (i.e., number of parameters and FLOPs) for architectures from the search spaces supported in \ourbenchmark{} in Fig. A3. 
Evidently, we observe that two clusters of search spaces emerge:

\begin{itemize}
    \item For one cluster of search spaces (i.e., NB101, NB201, and DARTS), their architecture decision variables mainly represent topological choices and connections, i.e., which operator (e.g., convolution or pooling) to use and how these operators are connected (see Figure 3 in the main paper for visualizations). We observe that these two objectives (i.e., number of parameters and FLOPs) are in harmony for architectures sampled from this cluster of search spaces. 

    \item For the other cluster of search spaces (i.e., NATS, MobileNetV3, and ResNet50), their architecture decision variables mainly represent the size and hyperparameters of the network (i.e., number of layers, channels, and kernel sizes, etc.). We observe that these two objectives (i.e., number of parameters and FLOPs) are in conflict for architectures sampled from this cluster of search spaces. 
\end{itemize}

\begin{figure*}[t]
    \begin{subfigure}[b]{0.325\textwidth}
    \centering
    \includegraphics[width=0.95\textwidth]{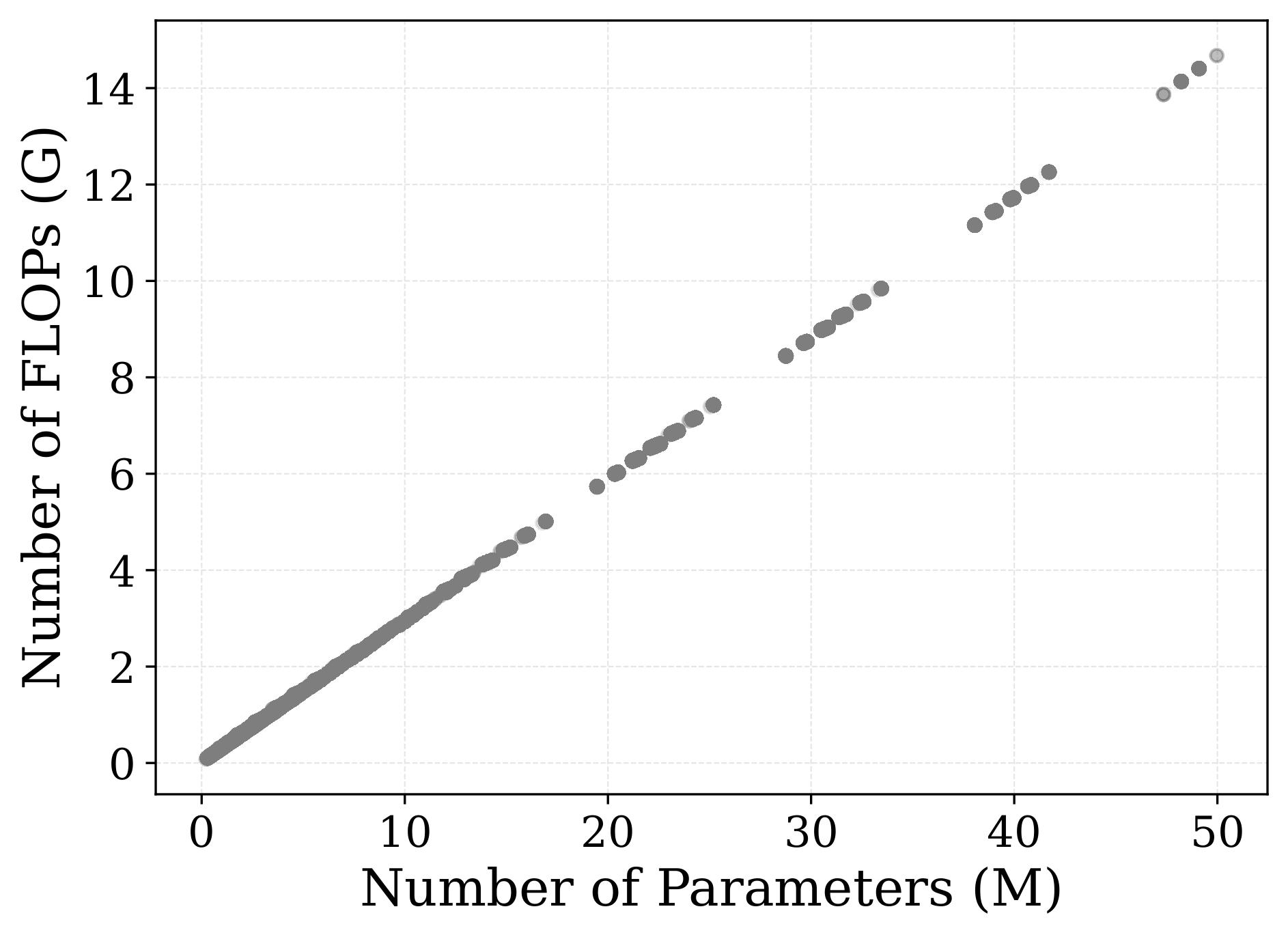}
    \caption{NB101 \label{fig:nb101_params_vs_flops}}
    \end{subfigure}\hfill
    \begin{subfigure}[b]{0.325\textwidth}
    \centering
    \includegraphics[width=0.95\textwidth]{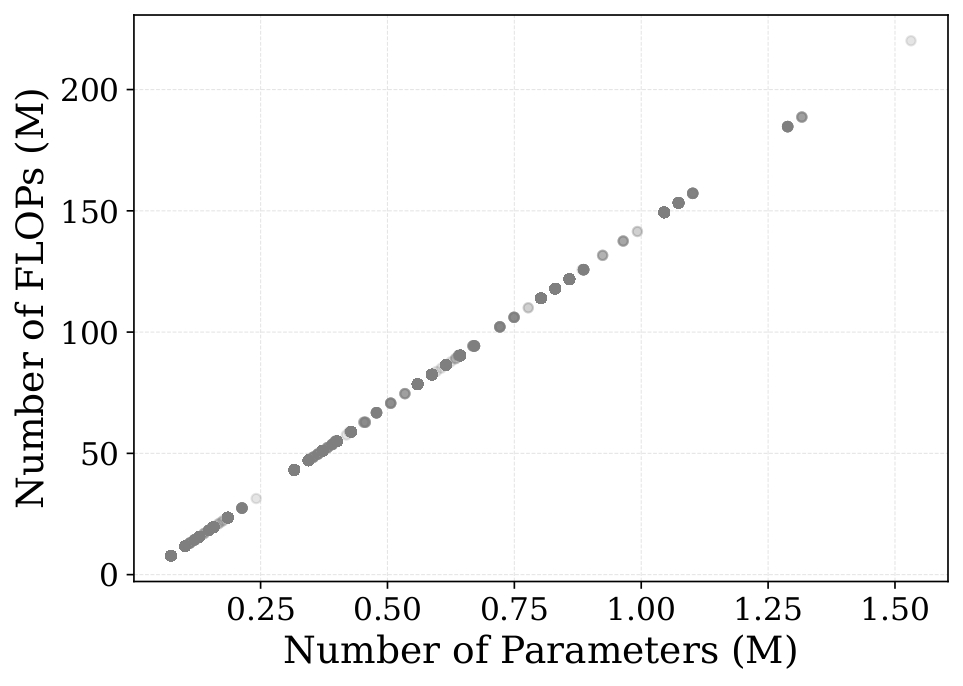}
    \caption{NB201 \label{fig:nb201_params_vs_flops}}
    \end{subfigure}\hfill
    \begin{subfigure}[b]{0.325\textwidth}
    \centering
    \includegraphics[width=0.95\textwidth]{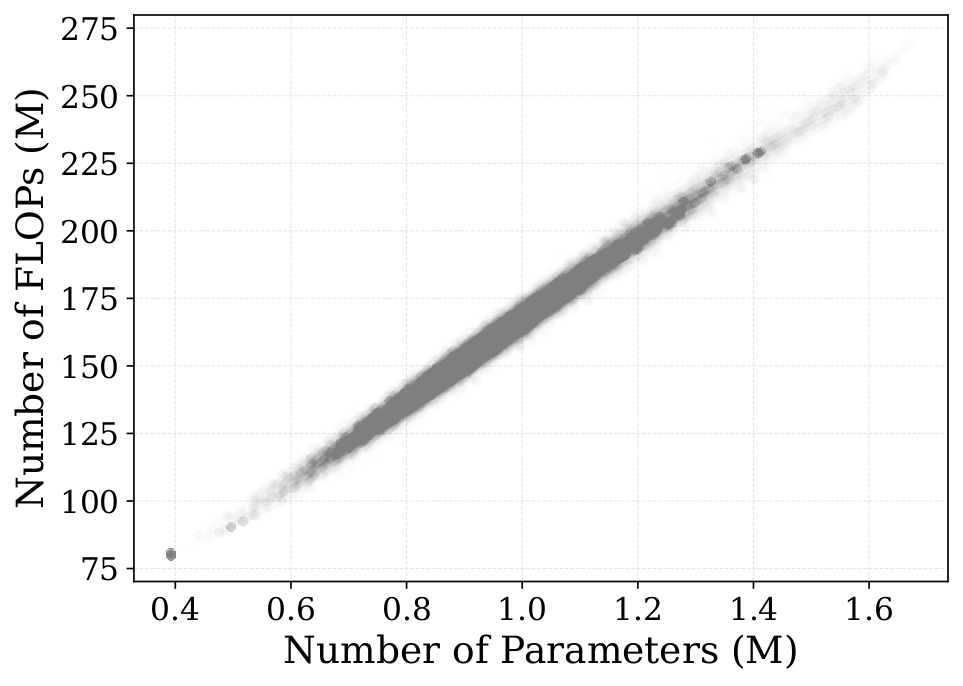}
    \caption{DARTS \label{fig:darts_params_vs_flops}}
    \end{subfigure}\\
    \begin{subfigure}[b]{0.325\textwidth}
    \centering
    \includegraphics[width=0.95\textwidth]{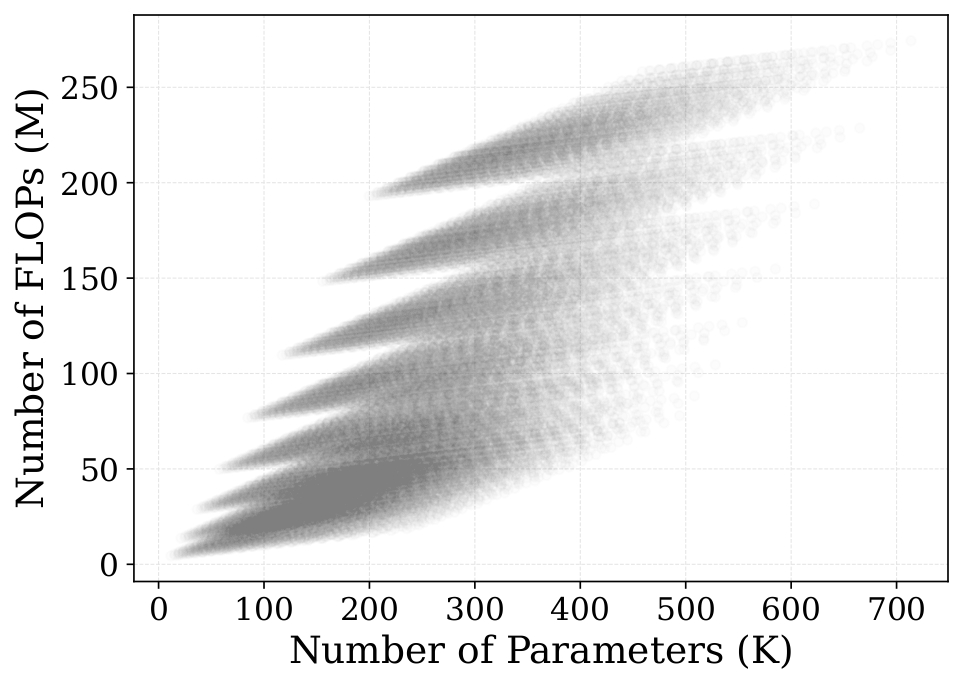}
    \caption{NATS \label{fig:nats_params_vs_flops}}
    \end{subfigure}
    \hfill
    \centering
    \begin{subfigure}[b]{0.325\textwidth}
    \centering
    \includegraphics[width=0.95\textwidth]{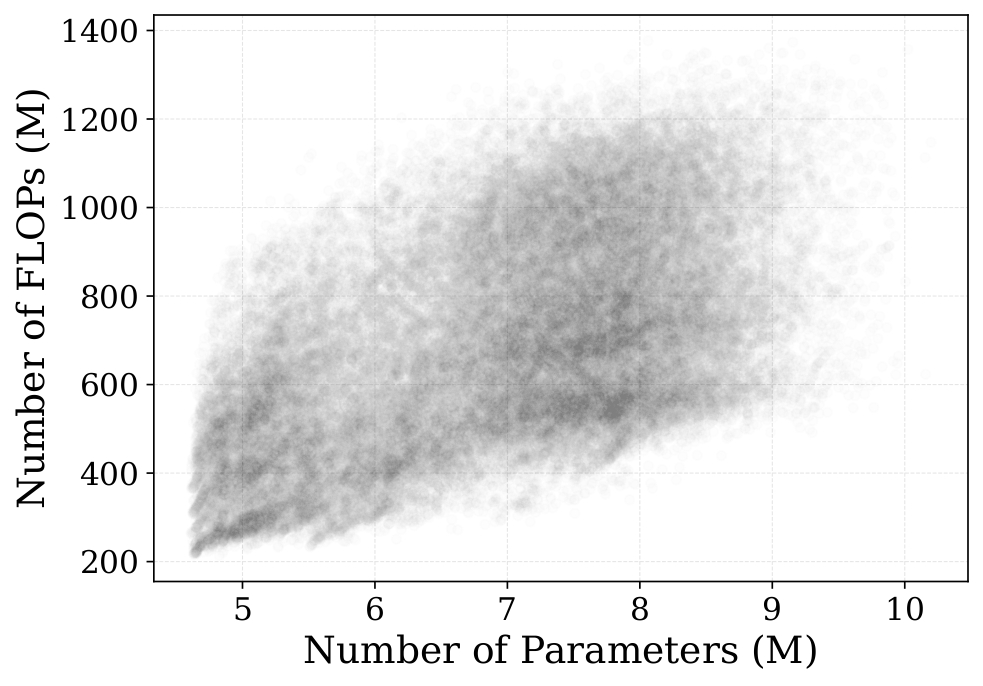}
    \caption{MobileNetV3 \label{fig:mnv3_params_vs_flops}}
    \end{subfigure}\hfill
    \begin{subfigure}[b]{0.325\textwidth}
    \centering
    \includegraphics[width=0.95\textwidth]{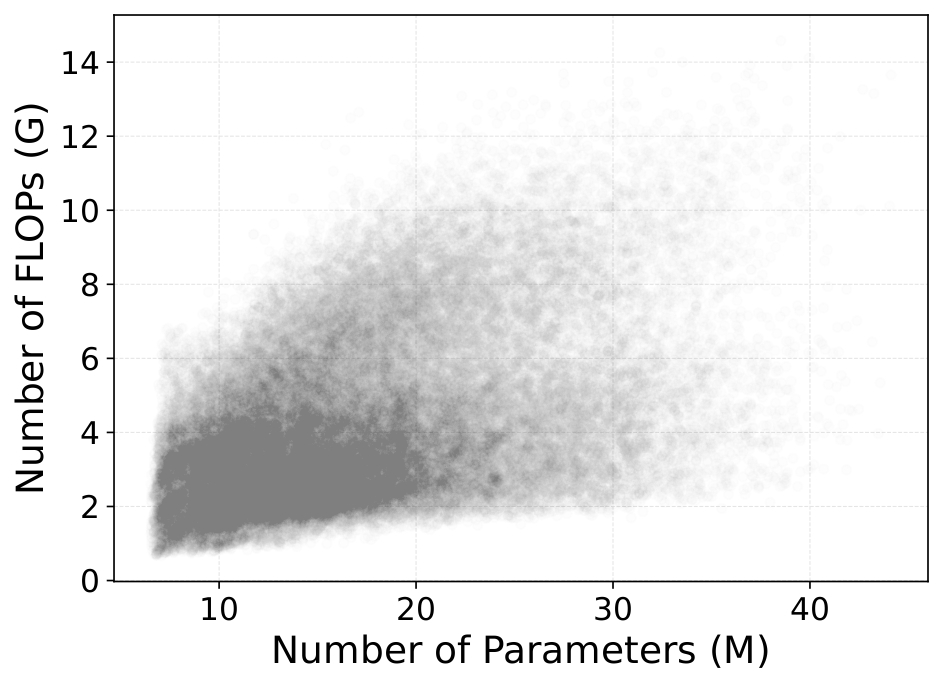}
    \caption{ResNet50 \label{fig:r50_params_vs_flops}}
    \end{subfigure}
    \caption*{Fig. A3: Visualization of correlation between number of parameters and FLOPs of the architectures from the search spaces supported in \ourbenchmark{}. 
    \label{fig:params_vs_flops}}
\end{figure*}

Further investigations are required to dissect the fundamental cause of these diverging behaviors from the two clusters of search spaces. However, such investigations are beyond the scope of this work. Nevertheless, we believe that these interesting observations can be viewed as additional evidences towards supporting the claim that our \ourbenchmark{} include a diverse set of search spaces of NAS. 

\section{Experimental Study Continued \label{sec:app_results}}




\subsection{Implementation Details} 
\subsubsection{Settings of population size}
We follow the Das-and-Dennis' approach to generate reference directions~\cite{das1998normal} for MOEA/D, NSGA-III, and RVEA. The population size $N$ is then determined by the simplex-lattice factor $H$ and the number of objectives $M$. For test instances with more than five objectives, we follow the bi-layer reference direction generation method as recommended in~\cite{nsga3}. The detailed settings are summarized in Table~V.III in the main paper. And to be consistent, we use the same population size for NSGA-II, IBEA, and HypE. Note that we also round $N$ to be a multiple of four and two for NSGA-II and NSGA-III, respectively.

\subsubsection{Settings of termination} A maximum of 10,000 function evaluations is used as the termination criterion for all algorithms. We perform 31 repetitions to each algorithm for a statistically meaningful comparison. 

\subsubsection{Other settings} we follow the default settings provided by the original papers to configure the genetic operators and the hyperparameters of the algorithms. And the settings of reference points for calculating the HV metric are provided in Table~A.I. Additionally, we visualize the true PFs in Fig.~A3 for \firsttestsuite{}1 - \firsttestsuite{}7, where all attainable solutions are exhaustively evaluated.

\begin{table*}[ht]
\centering
\caption*{Table A.I: The reference points used for calculating the HV metric on (a) \firsttestsuite{} and (b) \secondtestsuite{} test suites. \label{tab:hv_ref_point}}
    \begin{minipage}[b]{.61\textwidth}
    \caption*{(a) \firsttestsuite{}\label{tab:hv_ref_point_c10}}
    \resizebox{\textwidth}{!}{%
    \centering
    \begin{tabular}{@{\hspace{2mm}}lc@{\hspace{2mm}}}
    \toprule
    Problem & Reference point \\ \midrule
    \firsttestsuite{}1 & [0.1534, 3.2427e7] \\
    \firsttestsuite{}2 & [0.1577, 3.2427e7, 9.5450e9] \\
    \firsttestsuite{}3 & [0.2021, 5.7995e5, 2.5706e8] \\
    \firsttestsuite{}4 & [0.2021, 5.7995e5, 2.5706e8, 2.0064e-2] \\
    \firsttestsuite{}5 & [0.9000, 1.0735e6, 1.5327e8, 6.8889e-3, 3.2651e-2]\\
    \firsttestsuite{}6 & [0.5098, 1.0735e6, 1.5327e8, 1.0527e-2, 2.0139e-3, 26.596] \\
    \firsttestsuite{}7 & [0.9000, 1.0735e6, 1.5327e8, 8.1821e-3, 3.4711e-2, 1.0527e-2, 2.0139e-3,   27.078] \\
    \firsttestsuite{}8 & [0.2750, 1.6724e6] \\
    \firsttestsuite{}9 & [0.2750, 1.6724e6, 2.7034e8] \\ \bottomrule
    \end{tabular}}
    \end{minipage} \hfill
    \begin{minipage}[b]{.37\textwidth}
    \caption*{(b) \secondtestsuite{}\label{tab:hv_ref_point_in1k}}
    \resizebox{\textwidth}{!}{%
    \centering
    \begin{tabular}{@{\hspace{2mm}}lc@{\hspace{2mm}}}
    \toprule
    Problem & Reference point \\ \midrule
    \secondtestsuite{}1 & [0.3124, 4.4114e7] \\
    \secondtestsuite{}2 & [0.3124, 1.4577e10] \\
    \secondtestsuite{}3 & [0.3124, 4.4114e7, 1.4577e10] \\
    \secondtestsuite{}4 & [0.1832, 7.4134e7] \\
    \secondtestsuite{}5 & [0.1832, 1.5403e10] \\
    \secondtestsuite{}6 & [0.1832, 7.4134e7, 1.5403e10] \\
    \secondtestsuite{}7 & [0.2980, 1.0198e7] \\
    \secondtestsuite{}8 & [0.2980, 1.0198e7, 1.3768e9] \\
    \secondtestsuite{}9 & [0.2980, 1.0198e7, 1.3768e9, 7.0386e-2] \\ \bottomrule
    \end{tabular}}%
    \end{minipage}
\end{table*}

\begin{figure*}[ht]
    \centering
    \begin{subfigure}[b]{0.24\textwidth}
        \centering
        \includegraphics[trim={0 0 0 0}, clip, width=\textwidth]{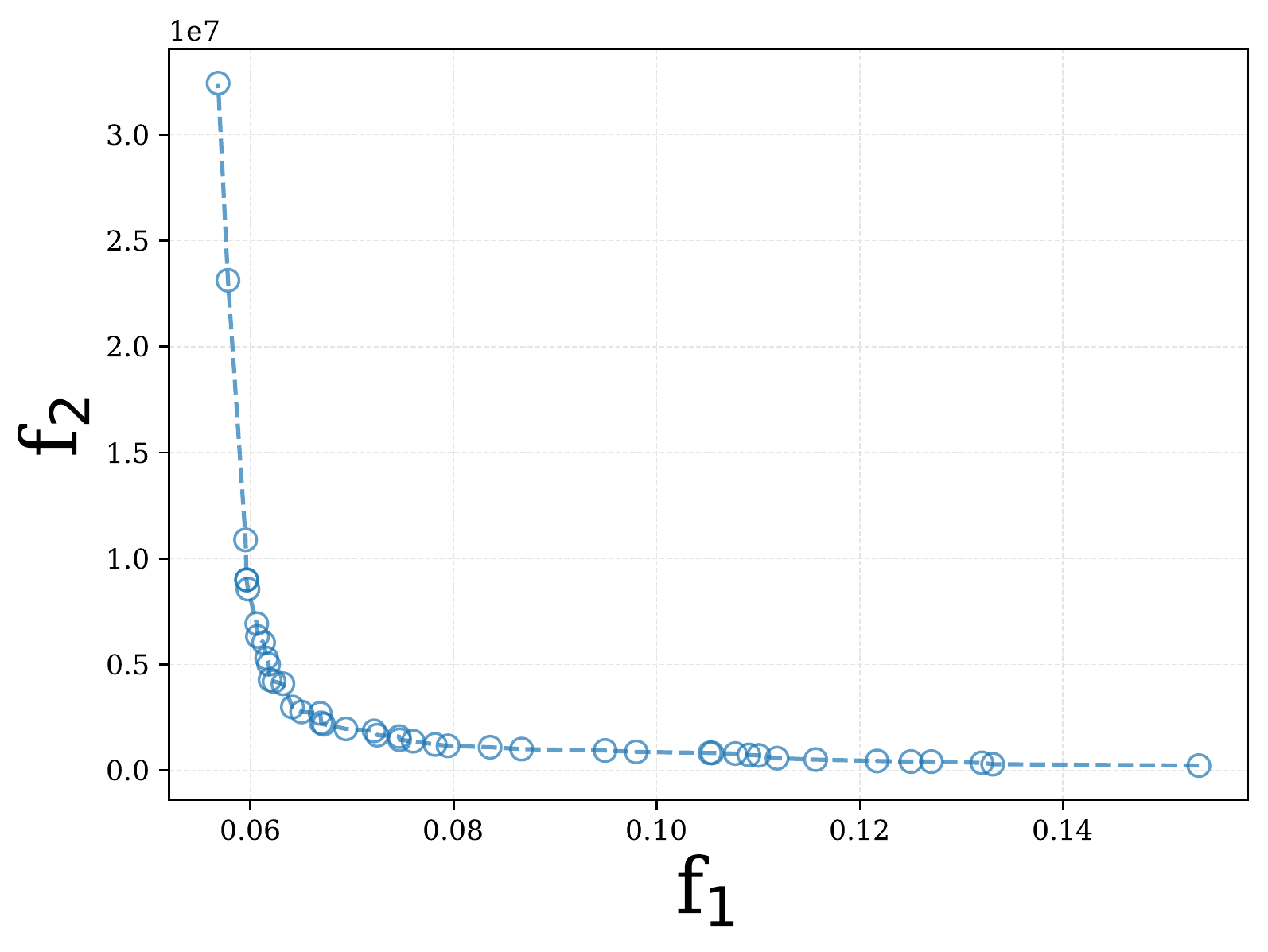}
        \caption{\firsttestsuite{}1\label{fig:app_c10mop1_pf}}
    \end{subfigure}\hfill
    \begin{subfigure}[b]{0.24\textwidth}
        \centering
        \includegraphics[trim={0 0 0 0}, clip, width=\textwidth]{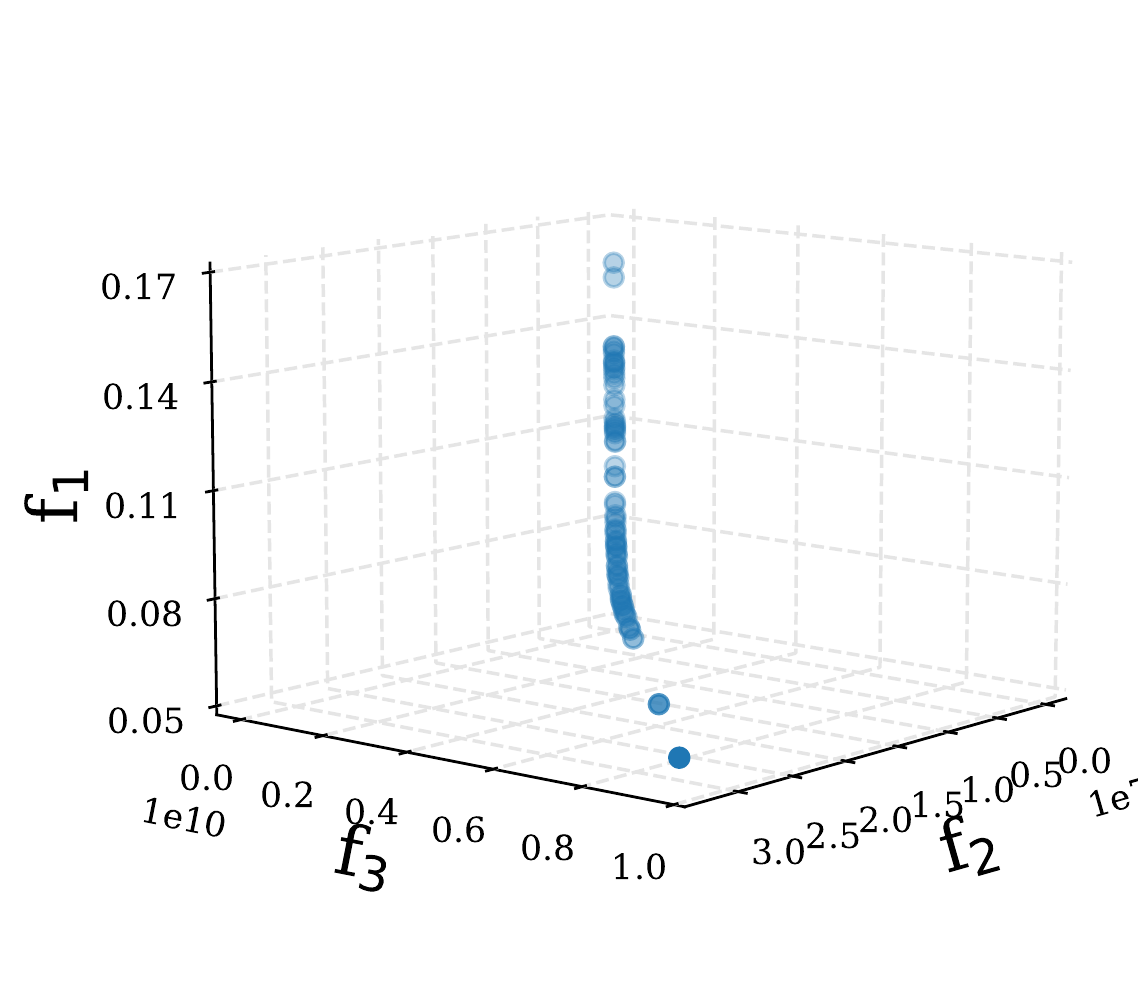}
        \caption{\firsttestsuite{}2\label{fig:c10mop2_pf}}
    \end{subfigure}\hfill
    \begin{subfigure}[b]{0.24\textwidth}
        \centering
        \includegraphics[trim={0 0 0 0}, clip, width=\textwidth]{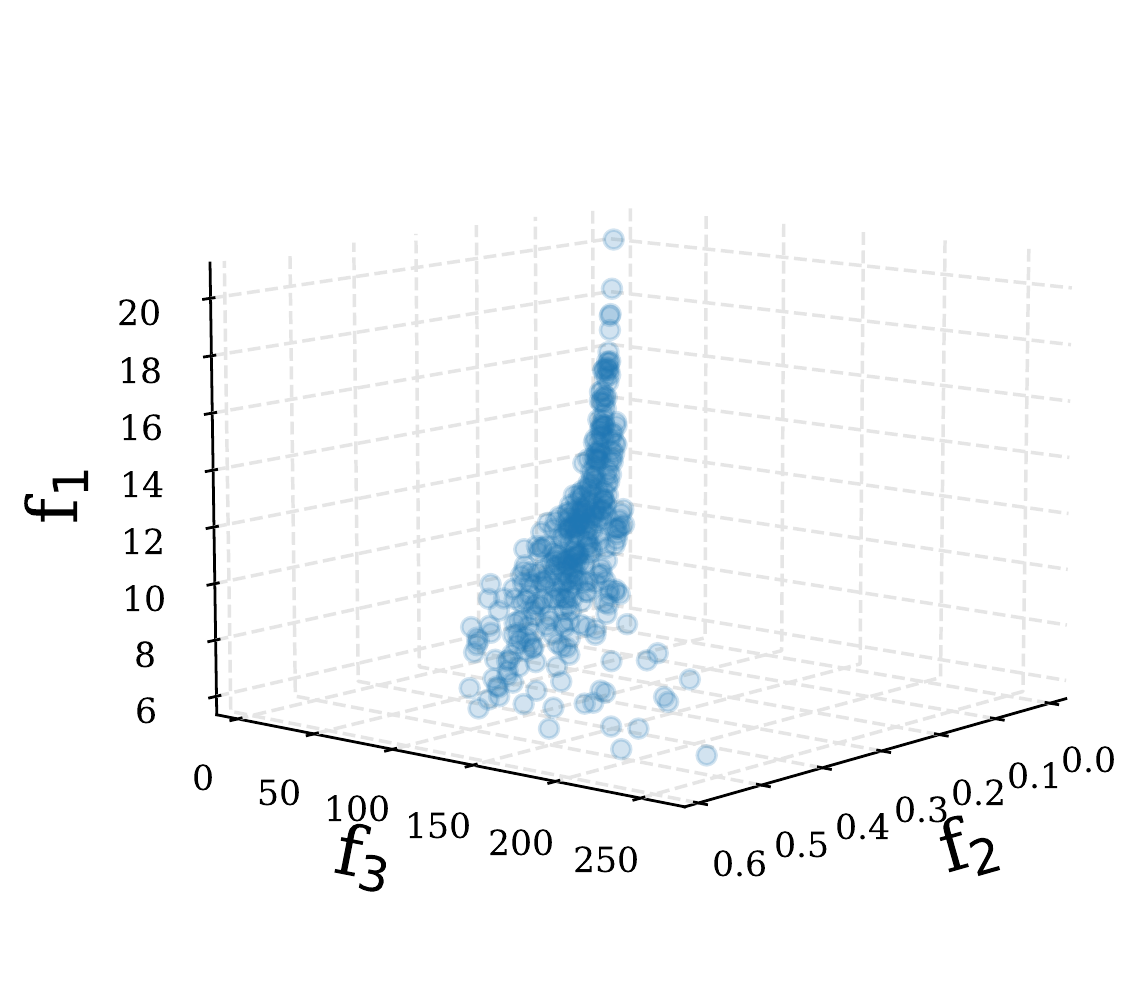}
        \caption{\firsttestsuite{}3\label{fig:c10mop3_pf}}
    \end{subfigure}\hfill
    \begin{subfigure}[b]{0.24\textwidth}
        \centering
        \includegraphics[trim={0 0 0 0}, clip, width=\textwidth]{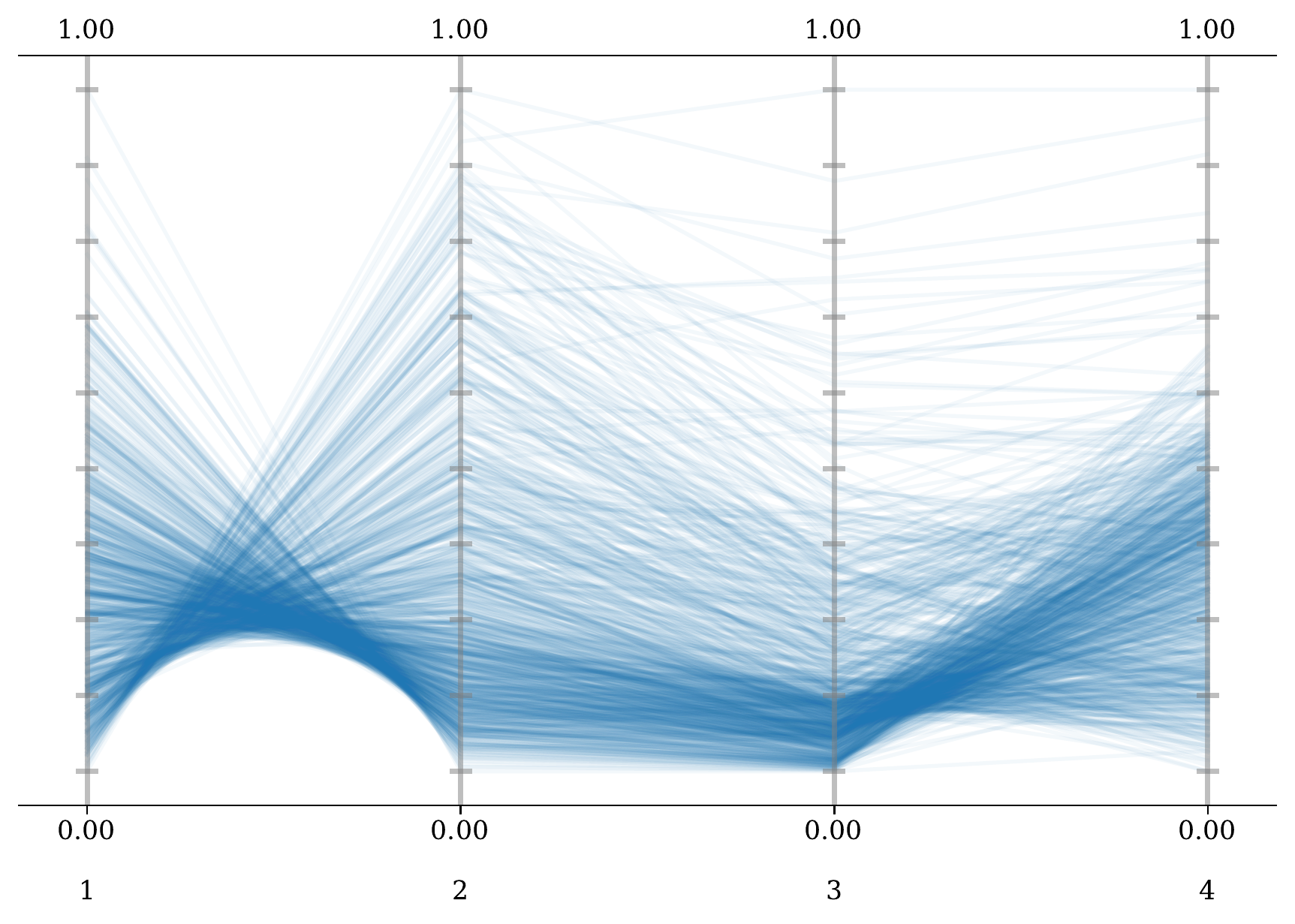}
        \caption{\firsttestsuite{}4\label{fig:c10mop4_pf}}
    \end{subfigure}\\
    \centering
    \begin{subfigure}[b]{0.24\textwidth}
        \centering
        \includegraphics[trim={0 0 0 0}, clip, width=\textwidth]{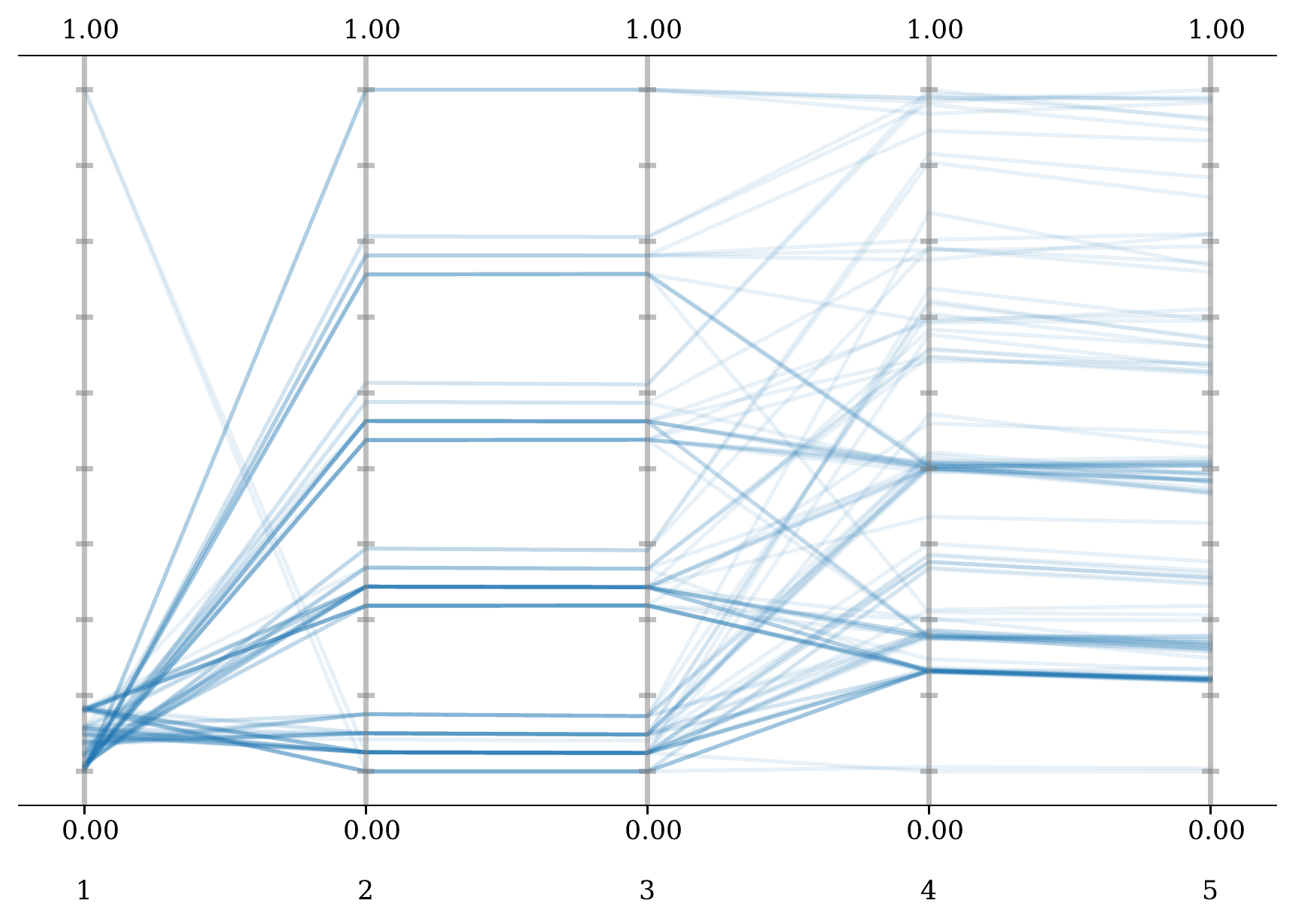}
        \caption{\firsttestsuite{}5\label{fig:c10mop5_pf}}
    \end{subfigure}
    \begin{subfigure}[b]{0.24\textwidth}
        \centering
        \includegraphics[trim={0 0 0 0}, clip, width=\textwidth]{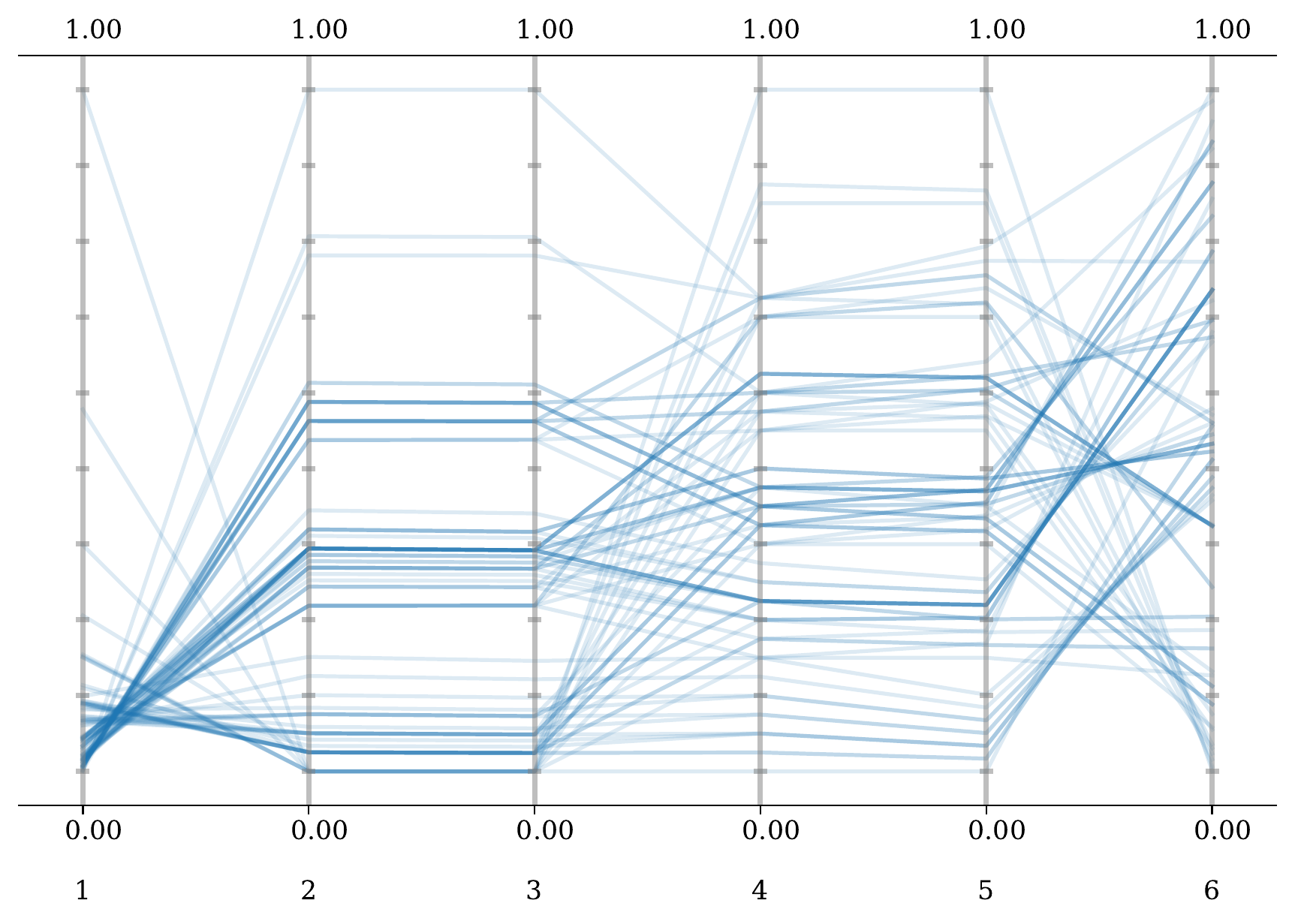}
        \caption{\firsttestsuite{}6\label{fig:app_c10mop6_pf}}
    \end{subfigure}
    \begin{subfigure}[b]{0.24\textwidth}
        \centering
        \includegraphics[trim={0 0 0 0}, clip, width=\textwidth]{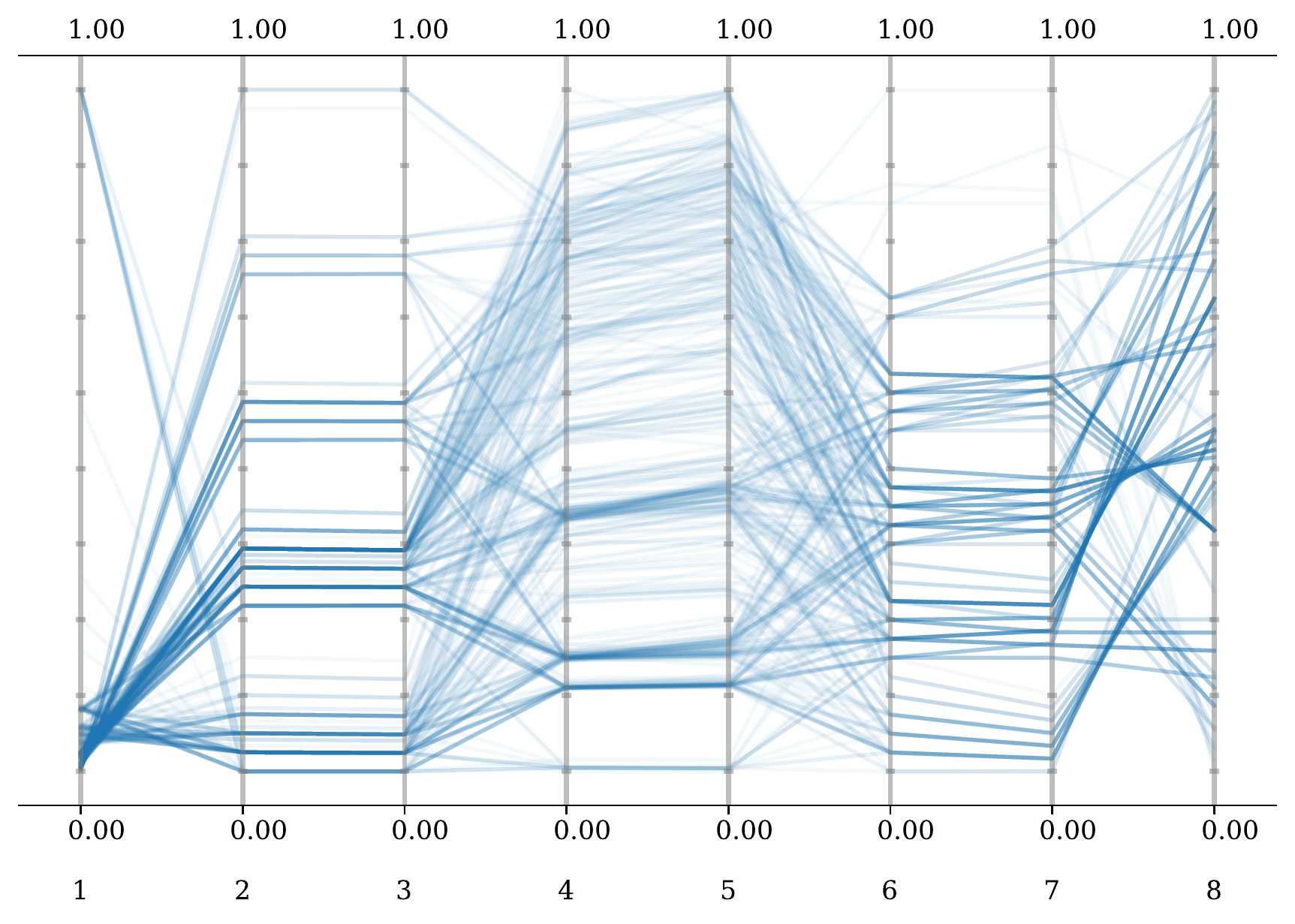}
        \caption{\firsttestsuite{}7\label{fig:c10mop7_pf}}
    \end{subfigure}\hfill
    \caption*{Fig.~A4: The true PFs of \firsttestsuite{}1 - \firsttestsuite{}7. 
    \label{fig:true_pf}}
\end{figure*}

\subsection{IGD results}
We present the statistical values of the IGD metric achieved by various EMO algorithms in Table A.II. Note that we only consider the first seven instances from the \firsttestsuite{} test suite, whose true PFs are known from exhaustive evaluations. 

\begin{table*}[ht]
\centering
\caption*{TABLE A.II: Statistical results (median and standard deviation) of the IGD values on the first seven instances of \firsttestsuite{} test suite. The best results of each instance are in bold. \label{tab:app-c10-igd}}
\resizebox{.9\textwidth}{!}{%
\begin{tabular}{@{\hspace{2mm}}lccccccc@{\hspace{2mm}}}
\toprule
EMO Algorithms & \firsttestsuite{}1 & \firsttestsuite{}2 & \firsttestsuite{}3 & \firsttestsuite{}4 & \firsttestsuite{}5 & \firsttestsuite{}6 & \firsttestsuite{}7 \\ \midrule
NSGA-II & \textbf{0.0309 (0.011)} & \textbf{0.0278 (0.009)} & \textbf{0.0235 (0.001)} & \textbf{0.0591 (0.002)} & \textbf{0.0054 (0.001)} & \textbf{0.0003 (0.000)} & \textbf{0.0444 (0.003)} \\
IBEA & 0.0917 (0.014) & 0.0927 (0.020) & 0.0248 (0.002) & 0.0736 (0.005) & 0.2030 (0.044) & 0.4025 (0.025) & 0.5005 (0.017) \\
MOEA/D & 0.0764 (0.015) & 0.2239 (0.014) & 0.0802 (0.010) & 0.1309 (0.009) & 0.5590 (0.016) & 0.5607 (0.048) & 0.7520 (0.078) \\
NSGA-III & 0.0346 (0.009) & 0.3437 (0.294) & 0.0471 (0.003) & 0.0933 (0.005) & 0.3901 (0.006) & 0.4569 (0.014) & 0.6061 (0.038) \\
HypE & 0.1194 (0.047) & 0.1317 (0.054) & 0.0486 (0.004) & 0.0765 (0.004) & 0.1021 (0.022) & 0.0699 (0.016) & 0.1713 (0.017) \\
RVEA$^\ast$ & 0.0505 (0.011) & 0.0585 (0.007) & 0.0443 (0.005) & 0.0849 (0.008) & 0.1937 (0.030) & 0.1421 (0.025) & 0.2736 (0.037) \\
\bottomrule
\end{tabular}%
}
\end{table*}

\section{Extending \ourbenchmark{} \label{sec:app_extending}}

\noindent\textbf{-- By a new search space: } depending on the volume, there are two ways to incorporate a new search space into \ourbenchmark{}. The specific procedures are outlined below with a pictorial illustration shown in Figure~A4 and an overview comparison provided in Table~A.III.

\begin{enumerate}
    \item If the volume of the new search space is confined such that an enumeration is possible (e.g., typically the total number of unique solutions/architectures should be $< 10^{5}$), we will simply exhaustively evaluate them all. Specifically, we train all solutions thoroughly from scratch on the training set and then evaluate them on the validation set. We repeat this process three times and store the results in a database. See the top path shaded in {\color{blue}blue} in Figure~A4.

    \item If the volume of the new search space is beyond enumeration, we resort to surrogate modeling. First, we train a supernet -- a composite network such that every attainable architecture (solution) becomes a sub-part of it. Second, we validate the effectiveness of the trained supernet on a subset of architectures uniformly sampled from the search space (we sample 100 architectures in this work). Depending on if the performance (i.e., prediction error, $f^{e}$) calculated based on the inherited weights (from the supernet) is sufficiently correlated with the performance calculated based on the thoroughly trained from-scratch weights, there are two ways to proceed:

    \begin{itemize}
        \item If the Kendall $\tau$ rank correlation is greater than the threshold ($\theta$ which is set to 0.7 in this work), the surrogate models are learned based on the objective values derived from the supernet. See the middle path shaded in \textcolor{green}{green} in Figure~A4. 

        \item Otherwise, we discard the supernet and revert to training from scratch to obtain the objective values for learning the surrogate models. See the bottom path shaded in \textcolor{red}{red} in Figure~A4. 
    \end{itemize}
    
\end{enumerate}

\begin{figure*}[ht]
    \centering
    \includegraphics[trim={0 0 0 0}, clip, width=.95\textwidth]{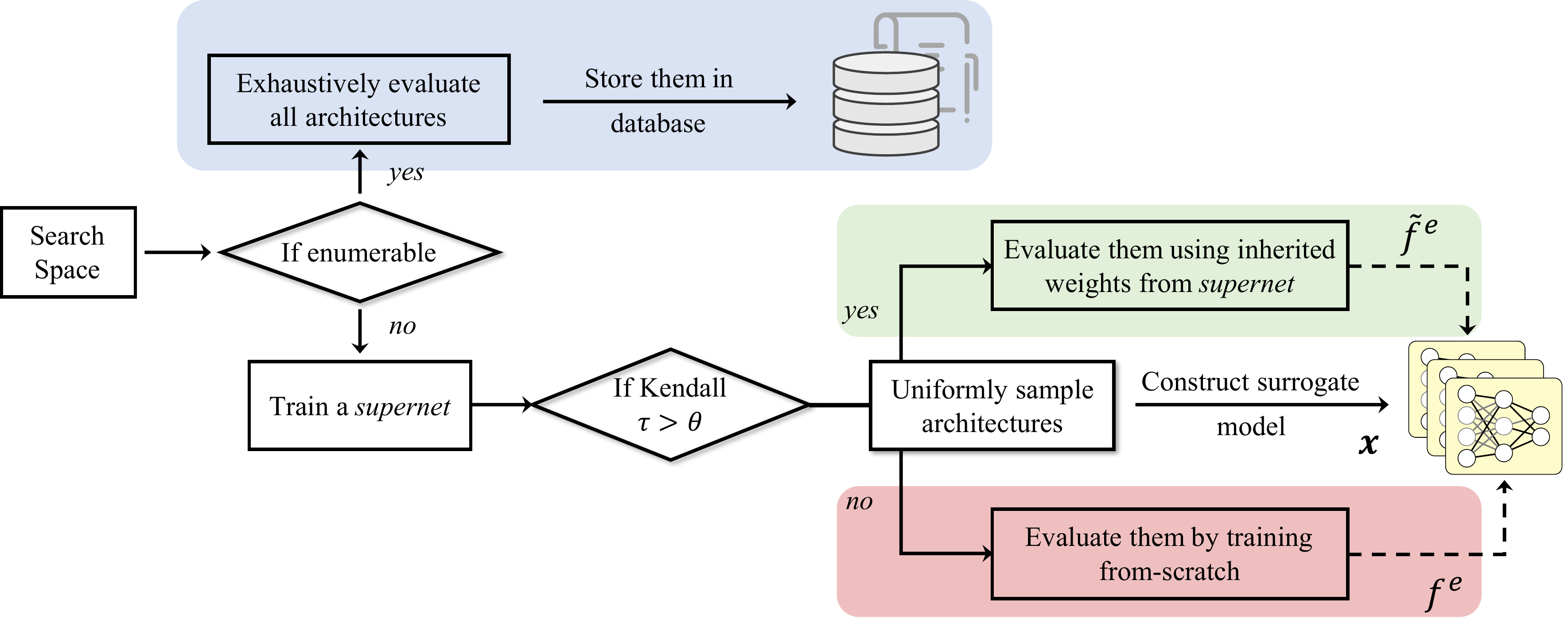}
\caption*{Fig.~A4: The procedure flowchart for adding a new search space to \ourbenchmark{}.}
\end{figure*}

\begin{table*}[ht]
\centering
\caption*{TABLE A.III: An overview comparison among the three ways of incorporating a search space to \ourbenchmark{} shown in Figure~A4. \label{tab:search_space_table}}
\resizebox{.95\textwidth}{!}{%
\begin{tabular}{@{\hspace{2mm}}p{0.22\textwidth}|p{0.315\textwidth}|p{0.315\textwidth}|p{0.15\textwidth}@{\hspace{2mm}}}
\toprule
\begin{tabular}[c]{@{}l@{}}Search spaces\\derived from\end{tabular} & Pros & Cons & Examples\\ \midrule
\cellcolor{blue!15}
\multirow{4}{*}{\begin{tabular}[c]{@{}l@{}}Exhaustive\\evaluation\end{tabular}} & 
\begin{itemize}
    \item Actual objective landscapes are completely known. 
\end{itemize} & 
\begin{itemize}
    \item Typically confined in size.
    \item Expensive to construct.
\end{itemize} & \multirow{4}{*}{\begin{tabular}[c]{@{}l@{}}NB101, NB201,\\NATS\end{tabular}} \\ \midrule
\cellcolor{green!15}
\multirow{4}{*}{\begin{tabular}[c]{@{}l@{}}Surrogate modeling\\ with supernet\end{tabular}} &  
\begin{itemize}
    \item Efficient to construct.
    \item Mild restriction on search space size.
\end{itemize} &
\begin{itemize}
    \item The supernet training is not trivial.
    \item Objective landscapes are approximated.
\end{itemize} & \multirow{4}{*}{\begin{tabular}[c]{@{}l@{}}ResNet50, \\ Transformer, \\ MNV3\end{tabular}}\\ \midrule
\cellcolor{red!15}
\multirow{5}{*}{\begin{tabular}[c]{@{}l@{}}Surrogate modeling\\ without supernet\end{tabular}} & 
\begin{itemize}
    \item Support almost any search space.
\end{itemize} & 
\begin{itemize}
    \item Expensive to construct.
    \item Objective landscapes are approximated.
    \item Final obtained architecture (solution) needs to be re-trained. 
\end{itemize} & \multirow{5}{*}{\begin{tabular}[c]{@{}l@{}}DARTS \end{tabular}} \\ \bottomrule
\end{tabular}%
}
\end{table*}

\noindent\textbf{-- By a new task: } more complex vision tasks (e.g., semantic segmentation) can be naturally incorporated into \ourbenchmark{} following the same steps of adding a new search space, as shown above.